\documentclass[10pt]{article}

\usepackage[preprint]{tmlr}

\usepackage{siunitx}
\usepackage[utf8]{inputenc}

\usepackage{graphicx}
\usepackage{booktabs}
\usepackage{multirow}
\usepackage{microtype}
\usepackage{colortbl}
\usepackage[colorlinks=true, citecolor=blue, linkcolor=blue, urlcolor=blue]{hyperref}
\usepackage{xcolor}
\usepackage{amsmath}
\usepackage{amssymb}
\usepackage{subcaption}
\usepackage{enumitem}

\usepackage[capitalise,noabbrev]{cleveref}

\def\month{MM}
\def\year{YYYY}
\def\openreview{\url{https://openreview.net/forum?id=XXXX}}

\begin{document}

\newcommand{\todo}[1]{\textcolor{red}{(#1)}}
\providecommand{\doi}{}
\renewcommand{\doi}[1]{doi: \href{https://doi.org/#1}{\detokenize{#1}}}
\makeatletter
\newcommand{\codeavailability}{\if@accepted Our code is publicly available at \url{https://github.com/ecker-lab/FoAViS}.\else Our code will be made publicly available upon acceptance.\fi}
\makeatother


\title{FoAViS: Analyzing Learned Fixation Strategies in a Foveated Active Vision System}

\author{\name Valentin~Hassler$^{1,*}$\quad Jonathan~Henrich$^{1}$\quad Yoha\"i-Eliel~Berreby$^{2,3,4}$\quad Haoze~Zhu$^{1}$ \\
      Alexander~Ecker$^{1,5,*}$ \\
      \addr $^{1}$University of G\"ottingen, Germany\quad
      $^{2}$McGill University, Canada\quad
      $^{3}$Mila - Quebec AI Institute, Canada \\
      $^{4}$Lamarck Labs, USA\quad
      $^{5}$Max Planck Institute for Dynamics and Self-Organization, G\"ottingen, Germany \\
      $^{*}$Correspondence: \texttt{valentin.hassler@uni-goettingen.de}, \texttt{ecker@cs.uni-goettingen.de}}

\maketitle

\begin{abstract}
Human vision is foveated and active: resolution is highest at the fovea and falls off towards the periphery, while fixations are actively directed to locations of interest. Building a machine system that processes information in the same way lets us study gaze behavior in a controlled setting where the system's components and objectives are known. The emergence of phenomena such as inhibition of return or task-specific fixation patterns can therefore be examined systematically, which may in turn inform our understanding of biological vision. This requires a model that can sample like the fovea, handle various tasks from classification to dense prediction, and be trained concurrently with a fixation policy. While prior work on active vision typically addresses only a subset of these aspects, a unified framework would be desirable to simplify model training and investigate gaze behavior under different task constraints. In our study, we present such a system and study its capabilities and the fixation policies that emerge from it. Our \emph{Foveated Active Vision System (FoAViS)} samples visual input by adopting the established log-polar foveation and accumulates information in a 2D spatial memory. Due to its 2D structure, the same memory can be read out globally for classification or densely for semantic segmentation, without any architectural change. Fixations are drawn from a 2D map by a policy trained concurrently with the task. We find that learned fixation policies select more informative, task-specific locations than non-learned heuristics. These policies also shape the task representations in a beneficial way: for both classification and segmentation, models trained concurrently with a policy outperform models trained with random locations, even when the latter are provided with learned scanpaths. Moreover, these scanpaths exhibit similarities to human gaze behavior by capturing aspects of empirical scanpath statistics and suppressing re-fixation of recently attended locations. Our work marks an important step away from standard passive, single-pass architectures, towards biologically grounded models of vision. It opens a path towards more comprehensive, biologically plausible models for studying goal-directed gaze behavior and information representation in the visual system. \codeavailability
\end{abstract}

\section{Introduction}
\label{sec:intro}

\begin{figure}[t!]
  \centering
  \includegraphics[width=1\textwidth]{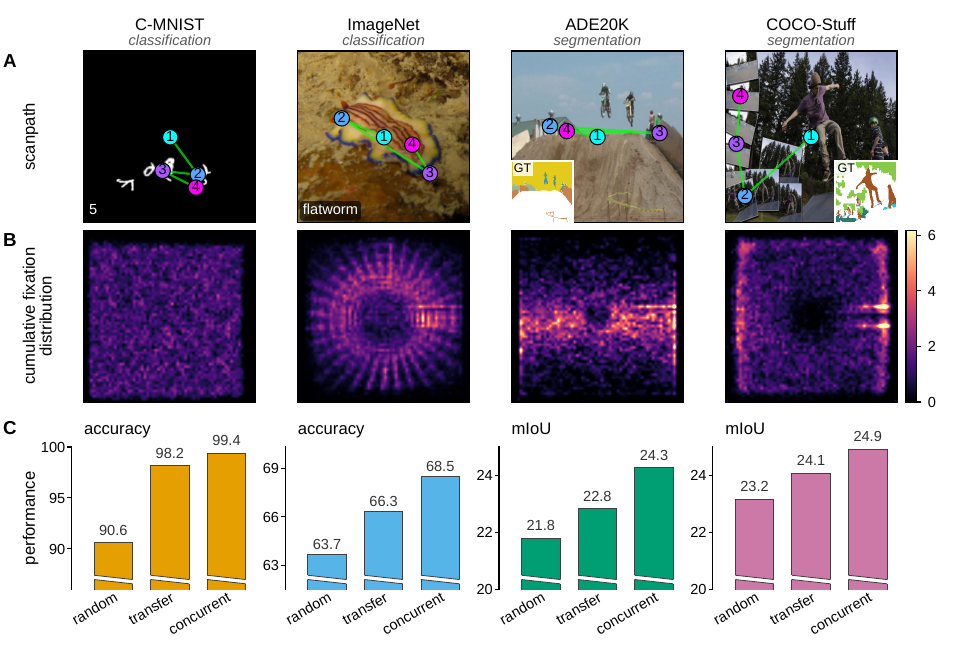}
\caption{\textbf{Emergent fixation strategies and model performance under concurrent policy and task training.} 
  \textbf{(A)}~Example scanpaths directing fixations towards informative image areas. 
  \textbf{(B)}~Cumulative distribution over fixations two to four (excluding the initial center fixation) across 10k (C-MNIST and ImageNet), 5k (COCO-Stuff) and 2k (ADE20K) images, showing task- and dataset-dependent spatial priors. The colorbar indicates likelihood relative to a uniform baseline, with values reflecting the factor by which probability exceeds chance. The distributions are consistent across models trained with different random seeds (\cref{app:shape_of_prior}). \textbf{(C)}~Downstream model performance comparing models trained concurrently with their policy and evaluated with it against models trained on random locations, evaluated either with random locations or via transfer of the learned policy.}
  \label{fig:intro}
\end{figure}
Traditionally, artificial neural networks that receive a static image as input have been used to gain insights into biological vision: networks trained on tasks such as object recognition develop representations that align with computations along the ventral stream~\citep{yamins2014, nayebi2018, cadena2024}, and such correspondences emerge even in self-supervised models never optimized for biological fidelity~\citep{raugel2025disentangling}. While these models have proven valuable, they remain structurally mismatched to the system they aim to explain: they process the whole image at once, passively, often in a single feedforward pass \citep[but see][]{kubilius2019brain}, typically with large monolithic architectures that neglect the hierarchical and modular organization of the visual system \citep[but see][]{bakhtiari2021functional,choi2023dual} as well as top-down feedback, and cross-area coordination~\citep{simeoni2025, ravi2025}. In contrast, biological vision is not a passive camera, but an active exploratory process. Because the visual field pairs a high-resolution foveal bottleneck with a heavily undersampled periphery~\citep{schwartz_SpatialMappingPrimate_1977}, the visual system must deploy rapid saccades sequentially towards informative regions. It is therefore forced to solve two coupled problems in tandem: identifying what is present at the current fixation to update an ongoing spatiotopic representation of the scene~\citep{golomb2021visual}, and using low-resolution peripheral cues to decide where to look next. This way of perceiving the world gives rise to characteristic fixation patterns~\citep{yarbus1967eye, kummerer2025modeling, itti1998model} and neural representations ~\citep{kowler2011eye,friston2012perceptions,pourrahimi_EmergentBrainlikeRepresentations_2025}. For an artificial model, learning behavioral and representational strategies that resemble those of biological vision is likely facilitated by analogous inductive biases: an eccentricity-dependent sampling strategy, a spatiotopic memory capable of integrating sequential glimpses over time, and an eye movement policy trained across diverse visual tasks. 

While active vision has been explored in computational research, existing models diverge considerably across several core design axes (\Cref{tab:feature_matrix}). At the sensory level, many architectures prioritize computational efficiency over biological plausibility: they typically rely on rectangular crops ~\citep{pardyl_AdaGlimpseActiveVisual_2025, canvit2026activevision, wang_EmulatingHumanlikeAdaptive_2025} or multi-scale patch pyramids~\citep{seifi2021glimpse, seifi2020attend, pourrahimi_EmergentBrainlikeRepresentations_2025} rather than preserving the eccentricity-dependent resolution falloff of human vision~\citep{lukanov_BiologicallyInspiredDeep_2021, gizdov2025seeing}. Furthermore, the memory component of previous works often flattens or averages away spatial structure~\citep{mnihRecurrentModelsVisual2014, baMultipleObjectRecognition2015,Olszewski_2025_WACV,wang_EmulatingHumanlikeAdaptive_2025}. While some models incorporate an explicit spatial memory, they remain constrained by low spatial resolution~\citep{wang_EmulatingHumanlikeAdaptive_2025} or limited representational capacity~\citep{liu_ImprovedEfficiencyBased_2023}, restricting their scope to spatially coarse tasks like image classification. A high-capacity memory that stays coordinate-aligned across glimpses would preserve scene topography and lend itself naturally to a broad range of tasks such as classification, segmentation or visual search. Recent concurrent work moves in this direction by integrating a high-resolution spatial memory into an active vision transformer~\citep{canvit2026activevision}. However, this model is trained using only random locations rather than a learned, task-driven policy, and has no support for eccentricity-dependent acuity. In general, we are not aware of any active vision model combining biologically motivated foveation, an explicit spatiotopic memory, and the ability to learn both image-level and dense tasks concurrently with a fixation policy. Such a system would provide a natural testbed for studying how characteristic fixation strategies and representations emerge under different task constraints.

In our work, we introduce FoAViS, a Foveated Active Vision System designed around these core principles. Visual input is sampled in a log-polar pattern, adopting the radial undersampling of biological vision. Information from multiple such glimpses is integrated over time using an allocentric, spatially aligned 2D memory, which can be seen as a computational analogue to spatiotopic updating~\citep{golomb2021visual}. The 2D memory is compatible with standard task heads, so it supports both global, image-level tasks like classification and dense, pixel-level tasks like semantic segmentation without any architectural change. Furthermore, our architecture incorporates a fixation policy that is trained concurrently with the task.

We find that FoAViS learns to direct fixations towards informative image regions (\Cref{fig:intro}A) and develops distinct fixation patterns across different tasks and datasets (\Cref{fig:intro}B). In addition, the policy exhibits inhibition of return (IoR)~\citep{itti1998model}, suppressing re-fixation of recently visited locations. These specialized gaze behaviors emerge naturally from downstream task optimization as solutions under biologically inspired active vision constraints. They also translate directly into performance gains: concurrently optimizing the visual representation and the fixation policy consistently outperforms architectures trained on random locations, even when the latter are evaluated using the learned policy at test time (\Cref{fig:intro}C). This underscores the importance of co-adapting the visual representation and fixation strategy.

\begin{table}[t]
  \centering
  \caption{\textbf{Comparison of active vision architectures across core design axes.} Entries are ordered chronologically by publication year. Columns indicate whether a requirement is fully supported (\checkmark), partially supported ($\sim$), or not supported ($\times$). Entries reflect each model as described and evaluated in its original publication. Some models could plausibly be extended to meet further requirements, but this was not explored in the respective work. FoAViS is discussed in more detail in \Cref{sec:method}.}
  \label{tab:feature_matrix}
  \resizebox{\textwidth}{!}{%
  \begin{tabular}{lcccccc}
    \toprule
    \textbf{Model} & \textbf{\shortstack{Biologically\\Motivated\\Foveation}} & \textbf{\shortstack{Explicit 2D\\Spatial\\Memory}} & \textbf{\shortstack{Trainable\\Fixation\\Policy}} & \textbf{\shortstack{Global Tasks\\(e.\,g., Cls.)}} & \textbf{\shortstack{Dense Tasks\\(e.\,g., Seg.)}} & \textbf{\shortstack{Concurrent Policy\\and Task\\Training}} \\
    \midrule
    RAM \citep{mnihRecurrentModelsVisual2014}                     & $\sim$ (Patch Pyramid)           & $\times$ (1D Recurrent)         & \checkmark                 & \checkmark & $\times$                & \checkmark \\
    DRAM \citep{baMultipleObjectRecognition2015}                  & $\sim$ (Patch Pyramid)           & $\times$ (1D Recurrent)         & \checkmark                 & \checkmark & $\times$                & \checkmark \\
    \citet{cheung2017emergence}                     & \checkmark (Learned Sampling)     & $\times$ (1D Recurrent)         & \checkmark                 & \checkmark & $\times$                & \checkmark \\
    \citet{renEndtoEndInstanceSegmentation2017}      & $\times$ (Rect. Crops)         & $\sim$ (Prediction-Masks Memory)    & \checkmark                 & \checkmark & \checkmark              & \checkmark \\
    Saccader \citep{elsayed_SaccaderImprovingAccuracy_2019}       & $\times$ (Feature Sampling)          & $\times$ (Mean Pooling)               & \checkmark                 & \checkmark & $\times$                & \checkmark \\
    Attend-and-Segment \citep{seifi2020attend}                    & $\sim$ (Patch Pyramid)           & \checkmark (Multi-Level Memory) & $\times$ (Uncertainty Map) & $\times$   & \checkmark              & $\times$ (No Policy) \\
    GFNet \citep{wangGlanceFocusDynamic2020}                      & $\times$ (Rect. Crops)           & $\times$ (1D Recurrent)         & \checkmark                 & \checkmark & $\times$                & $\times$ (Multi-Stage) \\
    GAE \citep{seifi2021glimpse}                                  & $\sim$ (Patch Pyramid)           & \checkmark (Multi-Level Memory) & $\times$ (Attention Map)   & \checkmark & \checkmark              & $\times$ (No Policy) \\
    TNet \citep{papadopoulosHardAttentionScalable2021}            & $\times$ (Rect. Crops)           & $\times$ (Mean Pooling)               & \checkmark                 & \checkmark & $\times$                & \checkmark \\
    \citet{lukanov_BiologicallyInspiredDeep_2021}  & \checkmark (Log-Polar Warp)    & $\times$ (None)                 & $\times$ (Feature Saliency)      & \checkmark & $\times$                & $\times$ (No Policy) \\
    FoveaTer \citep{jonnalagadda2023foveater}                     & $\times$ (Feature Sampling)       & $\sim$ (Token Bank)             & $\times$ (Attention Map)       & \checkmark & $\times$                & $\times$ (No Policy) \\
    STAM \citep{rangrej2022stam}                                  & $\times$ (Rect. Crops)           & $\sim$ (Token Bank)             & \checkmark                 & \checkmark & $\times$                & \checkmark \\
    NeVA \citep{schwinn2022behind}                                & $\sim$ (Gaussian Blending)       & $\sim$ (Pixel Memory)     & \checkmark                 & $\sim$     & $\times$                & $\times$ (Frozen Task Model) \\
    \citet{li2023modeling}                              & $\sim$ (Exponential Mask)        & \checkmark (Conv. Recurrent)    & \checkmark                 & \checkmark & $\times$                & \checkmark \\
    SimGlim \citep{jha2023simglim}                                & $\times$ (Rect. Crops)           & $\sim$ (Token Bank)             & $\times$ (Recon. Error Map)      & $\times$   & $\sim$ (Reconstruction) & $\times$ (No Policy) \\
    AME \citep{pardyl2023active}                                  & $\sim$ (Patch Pyramid)           & $\sim$ (Token Bank)             & $\times$ (Entropy)         & \checkmark & \checkmark              & $\times$ (No Policy) \\
    \citet{killick2023foveation}                   & \checkmark (Retina-Like Sampling) & $\times$ (None)                 & \checkmark                 & \checkmark & $\times$                & \checkmark \\
    \citet{liu_ImprovedEfficiencyBased_2023}           & $\sim$ (Two-Level Sampling)       & $\sim$ (Low Capacity) & \checkmark                 & \checkmark & $\times$                & $\times$ (Multi-Stage) \\
    AdaGlimpse \citep{pardyl_AdaGlimpseActiveVisual_2025}         & $\times$ (Rect. Crops)           & $\sim$ (Token Bank)             & \checkmark                 & \checkmark & \checkmark              & $\sim$ (Alternating) \\
    TORE \citep{Olszewski_2025_WACV}                              & $\times$ (Rect. Crops)           & $\sim$ (Token Bank)             & $\times$ (Entropy)         & \checkmark & $\sim$ (Reconstruction) & $\times$ (No Policy) \\
    GVSM \citep{pourrahimi_EmergentBrainlikeRepresentations_2025} & $\sim$ (Patch Pyramid)           & $\times$ (1D Recurrent)       & \checkmark                 & \checkmark & $\times$                & $\times$ (Multi-Stage) \\
    AdaptiveNN \citep{wang_EmulatingHumanlikeAdaptive_2025}       & $\times$ (Rect. Crops)           & $\sim$ (Low Res.)               & \checkmark                 & \checkmark & $\times$                & \checkmark \\
    CanViT \citep{canvit2026activevision}                         & $\times$ (Rect. Crops)           & \checkmark (High-Res Latent)    & $\times$ (Heuristic, Entropy)       & \checkmark & \checkmark              & $\times$ (No Policy) \\
    \midrule
    \textbf{FoAViS (ours)} & \textbf{\checkmark (Log-Polar Sampling)} & \textbf{\checkmark (High-Res LSTM)} & \textbf{\checkmark (2D map + RL)} & \textbf{\checkmark} & \textbf{\checkmark} & \textbf{\checkmark} \\
    \bottomrule
  \end{tabular}%
  }
\end{table}

\section{Related work}
\label{sec:related_work}
Active vision is about controlling a sensor to gather relevant information about a scene in an efficient way~\citep{bajcsy1988active}. While the scene of interest could in general be a video or a real-world setting, we focus on the simplest case in our work: a single fixed image that is viewed through a sequence of fixations, a paradigm long used to study human vision \citep{buswell1935people, yarbus1967eye}. By introducing Recurrent Models of Visual Attention (RAM), \citet{mnihRecurrentModelsVisual2014} were the first to provide a systematic deep learning framework for this paradigm, drawing on earlier works that processed visual scenes sequentially~\citep{schmidhuber1991fovea, itti1998model, felzenszwalb2010cascade, larochelle2010learning, denil2012learning}. Notably, this first deep learning framework already laid out the full set of components that more modern active vision models are still composed of:
\begin{itemize}[nosep]
  \item a \emph{foveation module} that turns a chosen location into a glimpse,
  \item a \emph{backbone} that accepts that glimpse as input,
  \item a \emph{memory module} that accumulates information across glimpses,
  \item a \emph{fixation policy} for choosing where to look next, and
  \item a \emph{task module} that turns the memory into a task-specific output, e.\,g. a classification.
\end{itemize}

Subsequent work has explored different choices for these components, for better task performance, lower compute, or closer correspondence to biological vision. Not all of these components carry equal weight as design decisions. The backbone and task module are usually kept simple and reflect whatever was standard at the time rather than being tailored to the active vision setting. The choices specific to active vision, and the ones that most determine what a model is capable of doing, are the foveation module, the memory module, and the fixation policy. In \Cref{sec:foveation_strategies,sec:rw_memory,sec:sampling}, we examine prior work along these three axes. In \Cref{sec:models_of_gaze}, we position FoAViS within the broader landscape of gaze models and highlight its value for providing a normative perspective.

\subsection{Foveation module}
\label{sec:foveation_strategies}

Models differ considerably in how a fixation is turned into network input. While some formulations stay close to the geometry of biological foveation, others depart from it substantially. Among the latter, one line of active vision models feeds rectangular crops directly to the backbone, with no foveation applied within a glimpse. To be able to explore larger scenes with this formulation, crops cover different extents of the image but are always resized to the same number of pixels: the first glimpse is typically the whole image at low resolution, giving a coarse overview, with later crops smaller and finer~\citep{wangGlanceFocusDynamic2020,papadopoulosHardAttentionScalable2021,pardyl_AdaGlimpseActiveVisual_2025,wang_EmulatingHumanlikeAdaptive_2025,canvit2026activevision}. Any foveal character thus comes from the changing crop size across glimpses, not the sampling geometry of a single glimpse.

Another widespread choice is the multi-scale patch pyramid, following RAM~\citep{mnihRecurrentModelsVisual2014}: several crops of different real extents are taken around the fixation point, resized to a common size, and passed to the backbone jointly~\citep{mnihRecurrentModelsVisual2014, seifi2020attend, seifi2021glimpse, pardyl2023active, pourrahimi_EmergentBrainlikeRepresentations_2025}. Each glimpse then carries both a fine view of the fixated point and a coarser view around it, but the falloff is discrete rather than the smooth space-variant compression of the retina.

Another strategy avoids sampling the input image altogether: the backbone first processes the full, non-foveated image, and the foveation is applied to the resulting feature map~\citep{jonnalagadda2023foveater, elsayed_SaccaderImprovingAccuracy_2019}.
For example, FoveaTer~\citep{jonnalagadda2023foveater} passes the whole image through a shallow convolutional backbone, giving a $14 \times 14$ grid of feature vectors, each summarizing a local neighborhood of the image. For each fixation, this map is then pooled around the corresponding grid position: the vectors immediately around it are kept as they are, while those further away are averaged together in increasingly large blocks. Averaging discards the detail, so resolution falls off with distance from the fixation without the input image itself ever being foveated. This is in contrast to biological vision, where foveation acts on the input itself.

Closer to biological foveation are methods that sample the image on a space-variant grid. In primates, retinal eccentricity maps approximately logarithmically onto primary visual cortex, so that central acuity is preserved while the periphery is progressively compressed~\citep{schwartz_SpatialMappingPrimate_1977,schwartz1980computational}. The log-polar mapping formalizes this as an image representation~\citep{weiman1979logarithmic,sandini_AnthropomorphicRetinaLikeStructure_1980,braccini1982model}. The image is sampled along concentric rings whose spacing grows with eccentricity, yielding a regular grid whose axes are angle and log-eccentricity. \citet{lukanov_BiologicallyInspiredDeep_2021} adopt such a mapping in an active vision model. However, two caveats come with this mapping: (1)~It is not translation-equivariant, so convolutions applied afterwards are not translation-equivariant in image space either (instead, they are equivariant to rotation and scaling around the center of fixation). (2)~Because the samples are mapped onto a rectangular grid, their radial and angular spacing differ. The sampling is therefore locally anisotropic, and the degree of anisotropy varies with eccentricity.

Two recent methods aim to resolve both caveats while keeping the eccentricity-dependent sampling. \citet{killick2023foveation} sample on an irregular point set, denser at the center and sparser towards the periphery but locally close to isotropic throughout. The samples are not remapped onto a radial-angular grid but processed directly with a graph convolution, so translation equivariance is preserved. \citet{blauch2026fovi} use a sampling pattern that directly implements the retino-cortical mapping of \citet{schwartz1980computational}, which is locally isotropic by construction. Here too the samples form no grid, and features are computed with $k$-nearest-neighbor convolutions. They provide a complete framework for combining this foveation with convolutional and transformer backbones, though it has not yet been demonstrated in an active vision system.

All of the sampling patterns discussed so far are hand-crafted beforehand. In contrast, \citet{cheung2017emergence} let the model learn how to sample the image under a constrained budget. They discover that classic biological foveation, with high resolution in the center and lower resolution towards the periphery, emerges entirely on its own, provided the model can only shift its gaze across the image rather than zoom in and out.

Despite the mentioned drawbacks, we adopt the log-polar sampling scheme (see \Cref{sec:fov}), following \citet{lukanov_BiologicallyInspiredDeep_2021}. It is space-variant like the retina, partially invertible, and produces a rectangular grid that can be processed by off-the-shelf convolutional backbones without modification. The framework of \citet{blauch2026fovi} offers a promising alternative, but emerged concurrently with our work.

\subsection{Memory module}
\label{sec:rw_memory}

Active vision models differ in how their memory accumulates information across glimpses and how much spatial structure it preserves. In the simplest case, per-glimpse features or predictions are pooled into a single vector with no spatial layout ~\citep{papadopoulosHardAttentionScalable2021, elsayed_SaccaderImprovingAccuracy_2019}. The pooling operation does not depend on what has already been seen, so earlier glimpses cannot change how later ones accumulate information. The memory therefore only has limited capacity to capture relationships between different glimpses.

In contrast, there is a line of models that replace the pooling step with a learned recurrent update, while still using a state vector without spatial layout. This is the setting of early active vision models~\citep{mnihRecurrentModelsVisual2014, baMultipleObjectRecognition2015} and of sequential routing architectures~\citep{wangGlanceFocusDynamic2020}. The recent GVSM~\citep{pourrahimi_EmergentBrainlikeRepresentations_2025} also follows this design, integrating visual information across sequential fixations through a dynamic internal state, implemented either as an LSTM or a causal transformer.

Another line of research keeps no fixed-size memory at all but instead accumulates a growing collection of per-glimpse entries. STAM~\citep{rangrej2022stam}, AME~\citep{pardyl2023active}, SimGlim~\citep{jha2023simglim} and AdaGlimpse~\citep{pardyl_AdaGlimpseActiveVisual_2025} accumulate raw image patches and re-encode the entire growing sequence through the full transformer from scratch at each step. In contrast, TORE~\citep{Olszewski_2025_WACV} splits the transformer into an extractor and an aggregator, processing each glimpse through the extractor once to cache its intermediate midway tokens, which the aggregator then reuses as new glimpses arrive.

A fourth group keeps a fixed-size 2D memory, preserving spatial layout but storing output or image content rather than semantic features. For example,~\citet{renEndtoEndInstanceSegmentation2017} accumulate predicted instance masks in a 2D ConvLSTM memory and pass them as an additional input channel in subsequent timesteps to provide top-down feedback. Other models iteratively uncover the input image itself~\citep{schwinn2022behind}, so the spatial memory holds raw pixel values that serve as input to a downstream task such as classification.

Closest to our setting are models whose memory stores semantic features. Attend-and-Segment~\citep{seifi2020attend} and its successor~\citep{seifi2021glimpse} extract multi-scale feature maps from each glimpse and write them deterministically into a persistent allocentric 2D spatial memory at coordinates corresponding to the glimpse's position in the image. Instead of deterministic coordinate-based writes, \citet{liu_ImprovedEfficiencyBased_2023} perform dynamic memory updates using a shallow ConvLSTM as a joint encoder and recurrent state. Yet, its constrained channel capacity restricts the expressive power of the stored representations. Similarly, AdaptiveNN~\citep{wang_EmulatingHumanlikeAdaptive_2025} maintains a dynamically updated, spatially organized state, but operates only at a low spatial resolution tailored for non-spatial, image-level tasks. CanViT~\citep{canvit2026activevision} instead maintains a high-resolution latent memory (termed a canvas in their work) that is updated by cross-attention with the current glimpse. In combination with distilling DINOv3 \citep{simeoni2025} features, this memory formulation has been shown to successfully integrate fine-grained semantic information over multiple glimpses to perform tasks like semantic segmentation.

Our memory (see \Cref{sec:memory}) also employs a global 2D spatial layout that accumulates semantically rich features at high resolution. Before updating the memory, incoming foveal features are mapped to the image reference frame to maintain precise topographic alignment across glimpses. Memory updates are then implemented as a per-position LSTM with shared weights, enabling step-by-step visualization and analysis of the gating dynamics. Because the memory preserves deep features at a fixed spatial resolution, it can be read out directly for both image-level classification and dense semantic segmentation without architectural modifications.

\subsection{Fixation policy}
\label{sec:sampling}

Existing active vision models generally determine where to look next through one of three paradigms: non-learned selection rules derived from network activations, end-to-end differentiable coordinate prediction, or stochastic policies that choose the next fixation. Both learned paradigms are forms of reinforcement learning, but they differ in how they are trained.

In non-learned selection, fixation targets are derived directly from internal activation maps or spatial heuristics calculated during the primary task's forward pass. By simply reusing existing feature representations derived from previous visual input, these approaches bypass the need for a dedicated policy network. Prior work has extracted non-learned fixation maps from averaged backbone feature activations~\citep{lukanov_BiologicallyInspiredDeep_2021}, accumulated self-attention weights~\citep{seifi2021glimpse, jonnalagadda2023foveater}, spatial uncertainty maps~\citep{seifi2020attend}, or the entropy of an internal representation~\citep{pardyl2023active, Olszewski_2025_WACV, canvit2026activevision}. Because selection is decoupled from training, models can even alter policies between training and inference: For example, CanViT~\citep{canvit2026activevision} trains on random locations but switches to an entropy-guided policy at inference to increase task performance. Because such heuristics rely on representation byproducts rather than task optimization, they may result in suboptimal fixation sequences.

This limitation can be addressed by formulating fixation selection as an end-to-end differentiable pipeline. In this framework, the model predicts continuous coordinates and extracts visual features through soft interpolation kernels~\citep{jaderberg2015spatial} centered at the predicted locations~\citep{renEndtoEndInstanceSegmentation2017,schwinn2022behind,li2023modeling,killick2023foveation}. For instance, \citet{schwinn2022behind} regress the next fixation location and center a Gaussian mask around it that smoothly blends the high-resolution image with a low-resolution version, assigning decreasing weight towards the periphery based on the mask's values. Because small shifts in predicted location continuously alter the extracted visual features, error gradients flow directly back from the downstream task loss to optimize the location predictor. In recent literature, however, these continuous formulations are less commonly adopted, as fixation selection via stochastic policies has been empirically found to be easier to optimize.

In these approaches, past observations are mapped to a spatial probability distribution from which the next fixation location is drawn. Although differentiable formulations exist for both the spatial cropping \citep{jaderberg2015spatial} and the stochastic selection of gaze targets~\citep{jang2016categorical, maddison2016concrete, cordonnier2021differentiable}, these models are typically optimized using policy gradient methods, which associate each selected fixation with a scalar reward based on downstream task performance. Early foundational architectures like RAM~\citep{mnihRecurrentModelsVisual2014} and DRAM~\citep{baMultipleObjectRecognition2015} established this formulation by predicting continuous Cartesian coordinates and drawing fixations from 2D Gaussian distributions centered on them, using policy gradients with a baseline network to reduce variance. More recent continuous approaches have found that active vision models benefit from receiving coarse global information before attending to details:
AdaGlimpse~\citep{pardyl_AdaGlimpseActiveVisual_2025} uses Soft Actor-Critic to dynamically adjust both the location and the scale of each glimpse, enabling the agent to learn an emergent strategy of starting with an overview before zooming in on details. AdaptiveNN~\citep{wang_EmulatingHumanlikeAdaptive_2025} builds this in explicitly by supplying an initial low-resolution version of the full image. In contrast to continuous coordinate approaches, discretized formulations simplify spatial search by evaluating candidate locations over fixed grids: Saccader~\citep{elsayed_SaccaderImprovingAccuracy_2019} restricts fixations to a pre-segmented grid of candidate patches, and similarly, GVSM~\citep{pourrahimi_EmergentBrainlikeRepresentations_2025} discretizes the scene into a $10 \times 10$ grid to output a 100-dimensional categorical distribution via spatial softmax. 

Following this discretized approach, we similarly draw fixation locations from a spatial probability map rather than regressing continuous coordinates (see \Cref{sec:sampling_mechanism}), which gives the model a natural mechanism for expressing uncertainty about where to look next. To optimize this fixation policy, we adopt the standard policy gradient formulation.

\subsection{Mechanistic, data-driven, and normative accounts of gaze behavior}
\label{sec:models_of_gaze}

Human saccadic trajectories are highly task-dependent, a phenomenon established by \citet{yarbus1967eye} and corroborated quantitatively across subsequent studies~\citep{borji2014defending}. Existing computational models of this behavior largely fall into three paradigms. First, mechanistic models formulate interpretable saccade-selection dynamics, progressing from early architectures based on saliency, winner-take-all, and inhibition of return~\citep{itti1998model} to activation-inhibition dynamical systems like SceneWalk~\citep{engbert2015spatial,schwetlick2023dynamical}. 

Second, data-driven frameworks directly fit image-to-gaze mappings~\citep{kummererbethge2021sota}, moving from free-viewing models like DeepGaze III~\citep{kummerer2022deepgaze} to task-conditioned successors such as ScanDiff~\citep{cartella2025modeling}. Bridging these two paradigms, recent hybrid approaches in free viewing use deep networks as discovery tools to identify what mechanistic models are missing~\citep{d2026moves}.

Third, a normative tradition asks neither how gaze is produced mechanistically nor how it can be empirically predicted, but what an observer \emph{should} look at under a given task and visual constraints~\citep{najemnik2005optimal, butko2010infomax, friston2012perceptions, sprague2007modeling, tatler2011eye, hayhoe2014modeling, hoppe2019multistep}. In this framework, fixations follow directly from an explicit objective, such as maximizing information gain or task reward. For example, the task of the Bayesian ideal searcher~\citep{najemnik2005optimal} is to locate a known target in a noisy background through successive views with a foveated sensor. Each fixation is then derived mathematically as the one that most reduces uncertainty about the target's location, and humans were found to search nearly as efficiently as this optimal strategy. 

While these prior works provide normative accounts in simplified settings, such as using synthetic visual stimuli or searching for a single known target, they do not scale to natural images and tasks. Deep active vision models instead learn fixation policies directly from raw pixels and a task loss, letting us study emergent visual behavior in complex scenes under controlled computational constraints. In our work, the underlying architecture remains constant across tasks, such that any divergence in the resulting fixation policy directly reflects the objective. We investigate this specialization (\Cref{sec:results_transfer}) and also show how established gaze characteristics emerge within this framing (\cref{sec:results_properties}). While traditional scanpath models must explicitly engineer mechanisms like IoR to suppress re-fixation~\citep{itti1998model,wloka2018starfc,chen_ScanpathPredictionVisual_2018,sun_VisualScanpathPrediction_2021}, our learned policies exhibit emergent IoR-like re-fixation avoidance~\citep{posner1984components,kleinmacinnes1999} without explicit reward or penalty, a behavior also exhibited by the recent GVSM~\citep{pourrahimi_EmergentBrainlikeRepresentations_2025}.

\section{Method}
\label{sec:method}

\begin{figure}[t!]
    \centering
    \includegraphics[width=\textwidth]{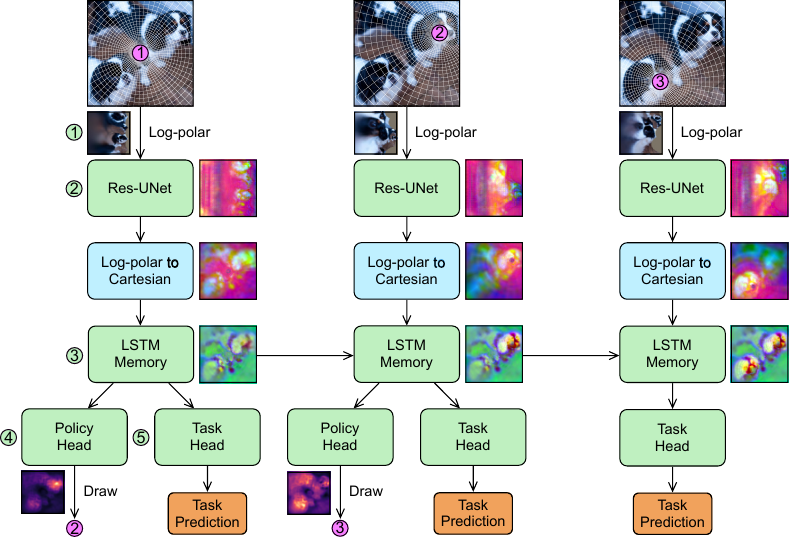}
    \caption{
            \textbf{Overview of FoAViS unrolled over three glimpses.} At each step, the current fixation serves as the center for a foveation module that extracts a log-polar glimpse~\textcircled{\raisebox{0.5pt}{\scriptsize 1}} (sampling grid shown in white). 
            The glimpse is encoded by the backbone~\textcircled{\raisebox{0.5pt}{\scriptsize 2}} (feature map visualized via PCA), backprojected to Cartesian image space, and accumulated into a 2D spatial memory using a shared per-position LSTM~\textcircled{\raisebox{0.5pt}{\scriptsize 3}} (memory state visualized via PCA). 
            The memory state is then read out by the policy head~\textcircled{\raisebox{0.5pt}{\scriptsize 4}} to produce a spatial probability map (logits visualized as a heatmap) from which the next fixation is drawn (purple number), while the task head~\textcircled{\raisebox{0.5pt}{\scriptsize 5}} uses it to generate the task prediction. \Cref{app:visualizations} shows these stages over full example scanpaths as visualizations.}
    \label{fig:architecture}
\end{figure}

FoAViS operates as an autoregressive loop over $T$ steps, where a step consists of the following five components:
\textbf{(1)}~a log-polar foveation module that samples the image around a fixation point with resolution decreasing towards the periphery, producing a foveated view we refer to as a \emph{glimpse},
\textbf{(2)}~a Res-UNet backbone that extracts a feature map from this glimpse,
\textbf{(3)}~an LSTM-cell memory module applied per position that accumulates these features across glimpses in a unified image reference frame,
\textbf{(4)}~a policy head that selects the next fixation point based on the current memory features (omitted for the last step), and
\textbf{(5)}~a task head that produces a task prediction based on the memory features.
\Cref{fig:architecture} provides an overview of the full pipeline, where the circled numbers 1--5 in green in the figure correspond to the five components listed above. In \Cref{sec:fov,sec:backbone,sec:memory,sec:sampling_mechanism,sec:task_readout}, we describe these components in more detail, followed by the training procedure in \cref{sec:model_training}.

\subsection{Foveation module}\label{sec:fov}
The model observes the scene through a glimpse (\Cref{fig:architecture}, \textcircled{\raisebox{0.5pt}{\scriptsize 1}}) that is obtained by applying a log-polar foveation module. We first center the Cartesian image coordinates $(x, y)$ around the current fixation point $f_t = (f_t^x, f_t^y)$, writing $x_{c,t} = x - f_t^x$ and $y_{c,t} = y - f_t^y$. Here and throughout the paper, $t$ denotes the time index of the fixation. We then map the centered coordinates to log-polar coordinates:
\begin{equation}
    \rho_t = \log \sqrt{x_{c,t}^2 + y_{c,t}^2},
    \qquad
    \alpha_t = \operatorname{atan2}(y_{c,t},\, x_{c,t}) \bmod 2\pi,
    \label{eq:logpolar}
\end{equation}
where $\operatorname{atan2}$ is the two-argument arctangent, so $\alpha_t$ is the angle of $(x, y)$ around the fixation point. Finally, we sample the image on a fixed grid of $R$ radial $\times\, A$ angular
points, reading off pixel values by bilinear
interpolation. We use $R = A = 64$, resulting in a fixed number of $64 \times 64 = 4096$ sampled pixels for each glimpse. The grid is uniform in angle $\alpha$ over $[0, 2\pi]$ and
uniform in log-radius $\rho$, extending from the fixation out to a maximum radius equal to the image side length $S$ (all images are square), i.\,e.\ $\rho_{\max} = \log S$. As a result, even a fixation at the very edge of the image still yields samples from the opposite side. Because the grid is uniform in log-radius,
sample points are packed densely near the fixation point and spread out towards the
periphery, encoding high resolution at the center and increasingly lower resolution towards the periphery. For quantitative characteristics of the sampling pattern, we refer to \Cref{app:logpolar_density}.

\subsection{Backbone}\label{sec:backbone}
The input to the backbone (\Cref{fig:architecture}, \textcircled{\raisebox{0.5pt}{\scriptsize 2}}) is the $64\times64$ glimpse obtained as described in \Cref{sec:fov}, which is in log-polar coordinates. Therefore, the features it computes are also in log-polar coordinates. We adopt a Res-UNet encoder-decoder architecture (details in \cref{app:architecture}) whose encoder consists of four stages of two residual blocks \citep{he2016deep}. With each stage, the number of channels doubles (starting at 64) and the spatial resolution is halved. The decoder likewise consists of four stages, each fusing the skip connection from the corresponding encoder stage and applying one residual block, restoring the full input spatial resolution. Unlike standard U-Net decoders, which typically reduce the number of channels towards the output, our decoder maintains a channel number of 128 throughout, ensuring that the recovered high-resolution features remain semantically rich. While stronger backbones would likely yield a better performance, our focus in this work was on designing a versatile active vision method, so we kept the backbone simple.

\subsection{Memory module}\label{sec:memory}
The memory module (\Cref{fig:architecture}, \textcircled{\raisebox{0.5pt}{\scriptsize 3}}) aggregates information from multiple glimpses over time in a unified image reference frame. Since the backbone operates in log-polar coordinates, its $64\times64$ output obtained as described in \Cref{sec:backbone}, must first be mapped back to the image reference frame by inverting the log-polar transformation \citep{cao2021lpnet}. Since the memory is processed by further modules (\cref{sec:sampling_mechanism,sec:task_readout}), we want it to be compact and to have a fixed size across varying image resolutions. We therefore do not map the backbone features to every individual image pixel. Instead, we map them onto a fixed $64\times64$ grid as follows: For each grid cell, we take its center point in image coordinates, compute its log-polar coordinates relative to the current fixation (\cref{eq:logpolar}), and bilinearly interpolate the backbone features. Grid points that are not covered by the glimpse that the backbone receives as input (\cref{sec:fov}) are set to zero. The resulting feature map $\mathbf{F}_t$ is passed to the memory module: a standard LSTM cell applied independently at every spatial position,
\begin{equation}
    \mathbf{H}_t = \operatorname{LSTM}_\theta\!\left(\mathbf{H}_{t-1},\, \mathbf{F}_t\right),
    \qquad \mathbf{H}_0 = \mathbf{0},
    \label{eq:memory}
\end{equation}
where $\mathbf{H}_t$ is a $64\times64$ grid of 128-dimensional hidden vectors that we refer to as the \emph{memory features} (cell state omitted from notation for brevity). Here and throughout the paper, $\theta$ denotes all learnable parameters of the model, and we subscript each learned module with $\theta$ accordingly.

\subsection{Policy head}\label{sec:sampling_mechanism}
The memory features $\mathbf{H}_t$ (\Cref{sec:memory}) aggregate information in a unified image reference frame, where each part of the image is represented at a different resolution depending on how close it was to the fovea during past glimpses. They therefore encode both what is in the scene and how well each region has been resolved so far, making them the ideal basis for deciding where to look next. Specifically, we use a policy head (\Cref{fig:architecture}, \textcircled{\raisebox{0.5pt}{\scriptsize 4}}) that turns the memory features into a probability distribution over candidate fixation points. It consists of a lightweight convolutional projection $\psi_\theta$ (three 128-channel convolutions and a final $1\times1$ convolution to a single channel; details in \cref{app:architecture}), followed by a softmax over all spatial positions,
\begin{equation}
    \pi_\theta(\cdot \mid \mathbf{H}_t) = \operatorname{softmax}\!\left(\psi_\theta(\mathbf{H}_t)\right),
    \qquad
    f_{t+1} \sim \pi_\theta(\cdot \mid \mathbf{H}_t),
    \label{eq:policy}
\end{equation}
yielding a $64\times64$ probability map from which the next fixation $f_{t+1}$ is drawn. Predicting a full spatial distribution, rather than regressing a coordinate directly, gives the model a natural mechanism for expressing uncertainty about where to look next.

\subsection{Task head}\label{sec:task_readout}
Analogous to the policy head, a task head $\phi_\theta$ (\Cref{fig:architecture}, \textcircled{\raisebox{0.5pt}{\scriptsize 5}}) produces a task prediction by reading out from the memory features (\Cref{sec:memory}):
\begin{equation}
    \hat{\mathbf{y}}_t = \phi_\theta(\mathbf{H}_t),
    \label{eq:task_head}
\end{equation} 
where $\hat{\mathbf{y}}_t$ is a vector or spatial map of logits for classification or segmentation, respectively. Because the memory is a standard 2D spatial feature map, it can be processed by whatever head the task requires. For classification, a convolutional projection (two 256-channel convolutions) is followed by global average pooling and a two-layer MLP. For semantic segmentation, a convolutional projection (a 128-channel convolution and a $1\times1$ convolution to the number of classes) maps each memory position to class logits, which are bilinearly upsampled to the input resolution. Details for both heads are given in \cref{app:architecture}. Task heads for other tasks such as visual search, object detection or instance segmentation would also be supported by the framework, though this is not part of the present work.

\subsection{Model training}\label{sec:model_training}
The components of \Cref{sec:fov,sec:backbone,sec:memory,sec:sampling_mechanism,sec:task_readout} form a single computational loop that can be unrolled over multiple glimpses. We train the model using a combination of a supervised loss for the differentiable task prediction path (foveation module, backbone, memory, and task head) and a policy-gradient loss for the fixation policy. The remainder of this section defines each loss term and how they combine into the total objective.

\paragraph{Episode structure.}
An episode unrolls the model over $T$ steps. A step of the model consists of foveating at the current fixation (\Cref{sec:fov}), encoding the resulting glimpse with the backbone (\Cref{sec:backbone}), updating the memory (\Cref{sec:memory}), and reading out from the memory using the policy head (\Cref{sec:sampling_mechanism}) and task head (\Cref{sec:task_readout}). The first fixation $f_1$ is fixed as the image center. At each step $t = 1, \dots, T-1$ the model then selects the next fixation $f_{t+1} \sim \pi_\theta(\cdot \mid \mathbf{H}_t)$ (\Cref{eq:policy}) and produces a task prediction $\hat{\mathbf{y}}_t$, while at the final step $t = T$ it produces only the task prediction $\hat{\mathbf{y}}_T$ and no further fixation. The model therefore takes only $T-1$ actions, the last at $t = T-1$, in contrast to standard reinforcement learning notation.

\paragraph{Task loss.}
The task predictions $\hat{\mathbf{y}}_t$ (\Cref{eq:task_head}) are supervised using a cross-entropy loss:
\begin{equation}
    \mathcal{L}_{\text{task}}(\theta) = \frac{1}{T} \sum_{t=1}^{T} \mathcal{L}_{\text{CE}}(\hat{\mathbf{y}}_t,\, \mathbf{y}).
    \label{eq:task_loss}
\end{equation}

For classification, this is the standard cross-entropy loss. For semantic segmentation, the cross-entropy is calculated per pixel and averaged over all spatial positions. Supervising at every timestep $t=1,\ldots,T$ encourages the model to provide an accurate prediction after any number of glimpses. Each of the loss terms is used to train the entire differentiable path, unrolled over multiple glimpses, via backpropagation through time. Supervising only the final step ($t = T$) yields comparable performance at the training horizon, but performance degrades when the model is unrolled for fewer or more steps (\Cref{app:task_loss_choices}).

\paragraph{Reward.}
A fixation should be rewarded for the improvement it brings. We therefore define the reward of the fixation selected at step $t$ as the resulting reduction in the task loss:
\begin{equation}
    r_t \;=\; \mathcal{L}_{\text{CE}}(\hat{\mathbf{y}}_t,\, \mathbf{y})
             - \mathcal{L}_{\text{CE}}(\hat{\mathbf{y}}_{t+1},\, \mathbf{y}),
    \qquad t = 1, \dots, T-1,
    \label{eq:reward}
\end{equation}
such that the reward grows with the size of the improvement. For simplicity, we credit each fixation with this reward directly, instead of
accumulating future rewards into a discounted return as is often done in
reinforcement learning. Equivalently, this can be regarded as a discounted return with a discount factor of zero. In our experiments, non-myopic return formulations generalized worse
when the model is unrolled for more steps at inference time (see \Cref{app:policy_loss_choices}).

We apply the following standardization to the rewards:
\begin{equation}
    \tilde{r}_t \;=\; \frac{r_t - \mathbb{E}\!\left[\,r_t\,\right]}
                           {\sigma\!\left[\,r_t\,\right]},
    \label{eq:reward_norm}
\end{equation}
where $\mathbb{E}[\,\cdot\,]$ and $\sigma[\,\cdot\,]$ are the empirical mean and standard deviation over the samples of the current batch at the same step. This equalizes the influence of early and late fixations, whose raw rewards differ considerably in magnitude because an early fixation reduces the task loss far more than a late one. It further prevents the reward scale from drifting as the task loss decreases over training.

\paragraph{Policy loss.}
The policy $\pi_\theta(\cdot \mid \mathbf{H}_t)$ of \Cref{eq:policy} defines a distribution over the $64\times64$ grid of candidate fixations, from which the next fixation $f_{t+1}$ is drawn. We optimize it via policy gradient~\citep{williams1992simple}. Intuitively, the update makes a fixation more probable when its standardized reward is positive, and less probable otherwise. Concretely, this is achieved by minimizing the following loss function,
\begin{equation}
    \mathcal{L}_{\text{policy}}(\theta) = -\sum_{t=1}^{T-1}
        \operatorname{sg}\!\left[\tilde{r}_t\right]
        \log \pi_\theta\!\left(f_{t+1} \mid \operatorname{sg}\!\left[\mathbf{H}_t\right]\right),
    \label{eq:policy_loss}
\end{equation}
where $\operatorname{sg}[\cdot]$ denotes the stop-gradient (stop-grad) operator. It is
applied to $\tilde{r}_t$ so that the reward only weights the gradient rather than receiving one itself. It further is applied to $\mathbf{H}_t$ to confine the policy gradient to the policy head, leaving backbone and memory to be shaped by the task loss alone. For a motivation of why we opted for this design, we refer to \cref{sec:gradient_conflicts}.

\paragraph{Total objective.}
The full training objective is the sum of the task
and policy loss:
\begin{equation}
    \mathcal{L}(\theta) = \mathcal{L}_{\text{task}}(\theta) + \mathcal{L}_{\text{policy}}(\theta).
    \label{eq:total_loss}
\end{equation}
Both terms are optimized simultaneously, but they act on disjoint sets of parameters because of the stop-grad in \Cref{eq:policy_loss}. The policy loss updates only the policy head, while the backbone, memory module, and task head are trained by the task loss alone. The foveation module is parameter-free.

\section{Experiments}
\label{sec:experiments}

\subsection{Datasets}
\label{sec:exp_datasets}

We train and evaluate on two image-classification datasets, ImageNet~\citep{deng2009imagenet} and Cluttered MNIST (C-MNIST), and two dense semantic-segmentation datasets, ADE20K~\citep{zhou2019semantic} and COCO-Stuff~\citep{caesar2018coco}. ImageNet is a large-scale natural-image recognition benchmark spanning \num[group-digits=integer]{1000} object categories, with \num[group-digits=integer]{1.28}M training and \num[group-digits=integer]{50000} validation images~\citep{deng2009imagenet}. C-MNIST is a custom $10$-class digit-recognition task in which one MNIST digit~\citep{lecun1998gradient} is placed as target on a larger canvas alongside five EMNIST letter distractors~\citep{cohen2017emnist}. Targets come from \num[group-digits=integer]{60000} training and \num[group-digits=integer]{10000} validation digits, distractors from \num[group-digits=integer]{124800} training and \num[group-digits=integer]{20800} validation letters. Samples are generated on the fly rather than drawn from a fixed pool, so there is no fixed image count. We refer to \cref{app:cmnist} for the exact generation procedure. Similar datasets derived from MNIST \citep{lecun1998gradient} have already been used in prior active vision works \citep{chen2021look,wang_EmulatingHumanlikeAdaptive_2025}. We include C-MNIST as a controlled setting in which the discriminative signal is small and spatially localized, so that the task can only be solved by actively fixating the target. ADE20K is a scene-parsing benchmark with $150$ semantic classes, comprising \num[group-digits=integer]{20210} training and \num[group-digits=integer]{2000} validation images. COCO-Stuff densely labels the COCO~2017 images with $171$ semantic classes ($80$ thing and $91$ stuff), comprising \num[group-digits=integer]{118287} training and \num[group-digits=integer]{5000} validation images.

\subsection{Trained models}
\label{sec:exp_setup}

All active vision experiments use the architecture and general training recipe described in \Cref{sec:method} with a fixed number of $T=4$ glimpses. The classification models were trained from random initialization, whereas the segmentation models were finetuned from the last checkpoint of the ImageNet classification model trained on random locations due to the limited size of the segmentation datasets. All models were trained three times under an identical training procedure. Complete optimization details, hyperparameters, and dataset preprocessing for all training settings can be found in \Cref{app:training_details}. All tasks are trained in two different settings:

The \emph{learned-policy} setting uses the full hybrid objective of \Cref{sec:model_training}, minimized concurrently: the policy-gradient term supervises the policy head and the task loss supervises the task head, memory and backbone. Since both objectives are minimized concurrently, this setting allows the policy to adapt to the task features and vice versa.

The \emph{random-location} setting instead draws fixations uniformly at random over the image, so the policy head is unused and the policy-gradient terms are absent during optimization. The backbone, memory, and task head are trained under the task loss. As in the learned-policy setting, the first fixation is fixed at the image center, and only the remaining $T-1$ fixations are drawn at random. The random-location setting was proposed by \citet{canvit2026activevision} as a way to train a policy-agnostic active vision model. We therefore include it in our work to compare its performance to the learned-policy setting.

We also train \emph{passive baselines} that consist of the Res-UNet backbone (\cref{sec:backbone}) and task head (\cref{sec:task_readout}) of FoAViS, but have no foveation, memory, or policy. Instead, they process the whole image at $128\times128$ pixels in a single forward pass. The baselines are thus architecturally comparable to FoAViS and receive as many pixels as it does across four $64\times64$ glimpses (of which fewer are unique for FoAViS due to foveal oversampling; \cref{app:logpolar_density}). They also share the training recipe of FoAViS, which includes initializing the segmentation baselines from the ImageNet baseline. Details are given in \Cref{app:passive_baseline}.

\section{Results and discussion}
\label{sec:results}

FoAViS processes visual scenes sequentially through a foveated sensory bottleneck, accumulates the resulting features into a coordinate-aligned 2D memory, and learns downstream task predictions concurrently with a fixation policy. We evaluate this framework across four benchmarks for classification and segmentation with a focus on how models perform when being supplied with their native (\Cref{sec:results_native}) vs. other policies (\cref{sec:results_transfer}) at test time. We then go on to analyze the emergent scanpath properties of our models, comparing them to results typically observed in human scanpath data (\cref{sec:results_properties}), and conclude with an analysis of problems encountered during joint policy and task optimization (\cref{sec:gradient_conflicts}).

In our work, we focus on policy and task learning in a biologically inspired architecture and investigate the properties of the learned policies, instead of optimizing benchmark performance. For completeness, we provide the results of prior active vision architectures on ImageNet and ADE20K alongside our models in \Cref{app:prior_work}. Direct comparability across these models is limited, however, as they differ substantially in input size, pixel budget, model size, pretraining schemes, and overall architecture.

\subsection{Native policy evaluation}
\label{sec:results_native}

\begin{table}[t]
  \centering
  \caption{\textbf{Native policy evaluation.} 
  A passive full-image baseline (top), models trained with random locations (middle), and models trained concurrently with a policy (bottom). Values report the mean $\pm$ std across three random seeds (single run for passive baselines). All active vision models were trained with a fixed horizon of $T=4$, but are evaluated at different \emph{\# Glimpses}. 
  For the learned-policy models, test-time fixations were selected deterministically via argmax from the policy's probability map. Passive baselines receive the full image at $128\times 128$ pixels.}
  \label{tab:main_combined_cur}
  \small
  \setlength{\tabcolsep}{5pt}
  \begin{tabular}{c c c cc cc}
    \toprule
    & C-MNIST & ImageNet & \multicolumn{2}{c}{ADE20K} & \multicolumn{2}{c}{COCO-Stuff} \\
    \cmidrule(lr){2-2} \cmidrule(lr){3-3} \cmidrule(lr){4-5} \cmidrule(lr){6-7}
    \# Glimpses & Acc (\%) & Acc (\%) & mIoU (\%) & Px Acc (\%) & mIoU (\%) & Px Acc (\%) \\
    \midrule
    \multicolumn{7}{c}{\textit{Baseline trained and evaluated with full images}} \\
    --          & 99.5 & 70.5 & 30.7 & 74.4 & 29.7 & 60.8 \\
    \midrule
    \multicolumn{7}{c}{\textit{Trained and evaluated with random locations}} \\
    1               & $76.6\pm0.2$ & $59.0\pm0.1$ & $19.7\pm0.3$ & $66.0\pm0.1$ & $20.2\pm0.2$ & $51.4\pm0.2$ \\
    4   & $90.6\pm0.3$ & $63.7\pm0.1$ & $21.8\pm0.4$ & $68.1\pm0.0$ & $23.2\pm0.2$ & $54.7\pm0.1$ \\
    6               & $93.4\pm0.2$ & $64.8\pm0.1$ & $22.7\pm0.0$ & $68.9\pm0.1$ & $24.0\pm0.1$ & $55.7\pm0.1$ \\
    \midrule
    \multicolumn{7}{c}{\textit{Trained and evaluated with its own learned policy}} \\
    1               & $76.4\pm0.2$ & $58.8\pm0.2$ & $19.4\pm0.4$ & $66.0\pm0.2$ & $20.3\pm0.2$ & $51.3\pm0.0$ \\
    4   & $99.4\pm0.0$ & $68.5\pm0.3$ & $24.3\pm0.1$ & $69.9\pm0.2$ & $24.9\pm0.3$ & $56.4\pm0.2$ \\
    6               & $99.4\pm0.1$ & $69.4\pm0.3$ & $24.9\pm0.2$ & $70.5\pm0.3$ & $25.9\pm0.3$ & $57.5\pm0.2$ \\
    \bottomrule
  \end{tabular}
  \par\vspace{2pt}
\end{table}

We first run each model with the fixation policy it was trained with -- the learned-policy model with its own policy head, and the random-location model with locations drawn uniformly across the image. This provides a direct measure of performance under the training conditions of each model.

Across all four benchmarks, models trained concurrently with a fixation policy consistently outperform their counterparts trained and evaluated on random locations (\Cref{tab:main_combined_cur}). At the training budget of $T=4$ glimpses, learning where to look provides a clear advantage across both image-level tasks (+8.8\% Acc on C-MNIST; +4.8\% Acc on ImageNet) and dense tasks (+2.5 mIoU on ADE20K; +1.7 mIoU on COCO-Stuff). This result shows that these models successfully learn to focus on informative regions, as is also supported by qualitative results (\Cref{fig:intro} and \Cref{app:devpanels}). With a single glimpse, i.\,e. only the initial center fixation, random-location and learned-policy models perform similarly, since no fixation has yet been chosen by the policy. Beyond the training budget, performance continues to scale upward for all benchmarks except C-MNIST, which saturates early (see \Cref{app:glimpse_sweep} for full scaling curves).

We compare FoAViS to passive full-image baselines with a similar architecture and pixel budget at $T=4$ glimpses (see \Cref{sec:exp_setup}) and find that it underperforms them ($-2.0\%$ Acc on ImageNet; $-6.4$ mIoU on ADE20K; $-4.8$ mIoU on COCO-Stuff). One contributing factor may be that fewer of the pixels it receives are unique due to foveal oversampling (\cref{app:logpolar_density}). The gap is larger for segmentation, possibly because this task requires a detailed semantic understanding of the full image, which can be hard to obtain from a sequence of foveated glimpses. Two design choices may also put FoAViS at a disadvantage relative to the passive baselines: (1)~Its backbone applies convolutions in log-polar coordinates, whereas convolutional architectures were developed for Cartesian images. (2)~The per-position LSTM memory updates each location independently, limiting how information is combined across glimpses.

\subsection{Policy transfer evaluation}
\label{sec:results_transfer}

\begin{table}[t]
  \centering
  \caption{\textbf{Policy transfer evaluation.} Models trained with random locations (top block) and models trained on a specific dataset with their own learned policy (bottom block) are evaluated with heuristic policies (top sub-block) and with learned policies (bottom sub-block). Shaded cells indicate evaluation with a model's native policy. Values report the mean $\pm$ std across three models trained with different random seeds. The heuristic policies are defined in \Cref{app:heuristics}.}
  \label{tab:policy_transfer_fused}
  \small
  \setlength{\tabcolsep}{5pt}
  \begin{tabular}{l c c cc cc}
    \toprule
    & & ImageNet & \multicolumn{2}{c}{ADE20K} & \multicolumn{2}{c}{COCO-Stuff} \\
    \cmidrule(lr){3-3} \cmidrule(lr){4-5} \cmidrule(lr){6-7}
    Evaluation policy & \# Glimpses & Acc (\%) & mIoU (\%) & Px Acc (\%) & mIoU (\%) & Px Acc (\%) \\
    \midrule
    \multicolumn{7}{c}{\textit{Trained with random locations}} \\
    random locations                 & 4 & \cellcolor{gray!15}$63.7\pm0.1$ & \cellcolor{gray!15}$21.8\pm0.4$ & \cellcolor{gray!15}$68.1\pm0.0$ & \cellcolor{gray!15}$23.2\pm0.2$ & \cellcolor{gray!15}$54.7\pm0.1$ \\
    triangle  & 4 & $65.0\pm0.1$ & $22.6\pm0.1$ & $68.9\pm0.1$ & $24.2\pm0.2$ & $55.8\pm0.1$ \\
    quadrant                & 5 & $65.9\pm0.1$ & $23.1\pm0.2$ & $69.5\pm0.1$ & $24.8\pm0.2$ & $56.6\pm0.2$ \\
    saliency + IoR & 4 & $65.6\pm0.2$ & -- & -- & -- & -- \\
    entropy + IoR     & 4 & -- & $22.4\pm0.3$ & $68.5\pm0.1$ & $23.3\pm0.2$ & $55.1\pm0.1$ \\
    \arrayrulecolor{black!30}\cmidrule(lr){1-7}\arrayrulecolor{black}
    ImageNet policy                  & 4 & $66.3\pm0.1$ & $22.8\pm0.3$ & $68.7\pm0.1$ & $24.0\pm0.1$ & $55.0\pm0.1$ \\
    ADE20K policy                    & 4 & $64.7\pm0.1$ & $22.8\pm0.3$ & $69.0\pm0.1$ & $23.6\pm0.0$ & $55.3\pm0.0$ \\
    COCO-Stuff policy                & 4 & $64.9\pm0.2$ & $22.9\pm0.3$ & $68.9\pm0.1$ & $24.1\pm0.2$ & $55.6\pm0.1$ \\
    \midrule
    \multicolumn{7}{c}{\textit{Trained with its own learned policy}} \\
    random locations                 & 4 & $63.3\pm0.1$ & $21.7\pm0.2$ & $68.3\pm0.1$ & $23.4\pm0.2$ & $54.7\pm0.1$ \\
    triangle  & 4 & $65.3\pm0.1$ & $22.9\pm0.2$ & $69.3\pm0.2$ & $24.5\pm0.1$ & $56.0\pm0.1$ \\
    quadrant                 & 5 & $66.1\pm0.2$ & $23.4\pm0.4$ & $69.9\pm0.2$ & $25.2\pm0.2$ & $56.9\pm0.1$ \\
    saliency + IoR & 4 & $66.7\pm0.2$ & -- & -- & -- & -- \\
    entropy + IoR     & 4 & -- & $22.9\pm0.4$ & $69.1\pm0.3$ & $23.9\pm0.1$ & $55.7\pm0.2$ \\
    \arrayrulecolor{black!30}\cmidrule(lr){1-7}\arrayrulecolor{black}
    ImageNet policy                  & 4 & \cellcolor{gray!15}$68.5\pm0.3$ & $23.7\pm0.2$ & $69.4\pm0.2$ & $24.9\pm0.1$ & $55.5\pm0.1$ \\
    ADE20K policy                    & 4 & $65.5\pm0.1$ & \cellcolor{gray!15}$24.3\pm0.1$ & \cellcolor{gray!15}$69.9\pm0.2$ & $24.3\pm0.1$ & $55.8\pm0.1$ \\
    COCO-Stuff policy                & 4 & $65.5\pm0.5$ & $24.0\pm0.4$ & $69.8\pm0.3$ & \cellcolor{gray!15}$24.9\pm0.3$ & \cellcolor{gray!15}$56.4\pm0.2$ \\
    \bottomrule
  \end{tabular}
\end{table}

Because our model learns to fixate on task-relevant regions, a baseline trained and evaluated on random locations naturally underperforms. However, \citet{canvit2026activevision} show that random-location models can benefit substantially from more sophisticated policies at test time. Conversely, it remains an open question how robust models trained concurrently with a policy are when evaluated with other fixation strategies. To investigate this, we evaluate every model under non-native policies. That is, a model receives externally supplied fixation points instead of selecting them with its native policy. We consider two types of such policy transfer.

In \emph{heuristic policy transfer}, fixations are selected either independently of image content or based on internal activation maps of the model. Specifically, we evaluate five strategies (see  \Cref{app:heuristics}): (1)~uniform random locations, (2)~a fixed triangle scan pattern, (3)~a fixed quadrant scan pattern, (4)~a saliency-based heuristic inspired by \citet{lukanov_BiologicallyInspiredDeep_2021} combined with a hardcoded IoR mechanism (suppression radius of a quarter of the image width), and (5)~an entropy-based heuristic inspired by \citet{canvit2026activevision} using the same hardcoded IoR mechanism. 

In \emph{learned policy transfer}, fixations come from another learned-policy model: it is run on the images to be evaluated, and the fixation sequences it selects (probability map argmax) are used as forced fixations for the receiving model. The receiving model can be another learned-policy model or the random-location model.

\paragraph{Policy transfer to the random-location model.}
We find that models trained on random locations can be successfully paired with more elaborate policies to achieve improved performance (\Cref{tab:policy_transfer_fused}). For example, on ImageNet, both the triangle policy ($65.0\%$) and the saliency policy ($65.6\%$) outperform random locations ($63.7\%$). The learned ImageNet policy yields further gains ($66.3\%$), surpassing the quadrant policy ($65.9\%$) despite the latter using an additional glimpse. While segmentation models also improve with heuristic and learned policies, the learned policies provide no additional benefit over the heuristic ones. Interestingly, even when evaluated with the corresponding learned policy, the model trained on random locations fails to match the performance of models trained concurrently with a policy on ImageNet ($66.3\%$ vs.\ $68.5\%$), ADE20K ($22.8$ vs.\ $24.3$ mIoU), and COCO-Stuff ($24.1$ vs.\ $24.9$ mIoU). This demonstrates that the co-adaptation of representations to the fixation policy during training is beneficial. We see two possible, non-exclusive explanations: (1)~an encoder trained concurrently with a policy is exposed to more informative data during training and thus learns better representations, and (2)~the memory can learn to aggregate information more effectively across glimpses when these are directed to semantically meaningful locations. The result also puts the findings of \citet{canvit2026activevision} into perspective, who report that policy-agnostic training benefits from sophisticated policies at test time, but do not investigate concurrent training of encoder and policy.

\paragraph{Policy transfer to the learned-policy model.}
When evaluating a model trained concurrently with a policy under other fixation strategies, we find that performance generally degrades. On ImageNet, this drop is especially pronounced: its native policy performance ($68.5\%$) substantially exceeds that of the closest alternatives, namely the saliency-based policy ($66.7\%$) and the learned ADE20K and COCO-Stuff policies ($65.5\%$ each). Evaluating it on random locations ($63.3\%$) yields worse performance than the model trained directly on random locations ($63.7\%$). Furthermore, under the triangle policy ($65.3\%$), it performs roughly on par with the random-location model evaluated under the same policy ($65.0\%$). A similar pattern is observed for the segmentation models. These results indicate that the backbone of a model trained concurrently with a policy specializes to the input distribution induced by that policy.

\subsection{Properties of learned policies}
\label{sec:results_properties}

\begin{figure}[!tbp]\centering
\includegraphics[width=\textwidth]{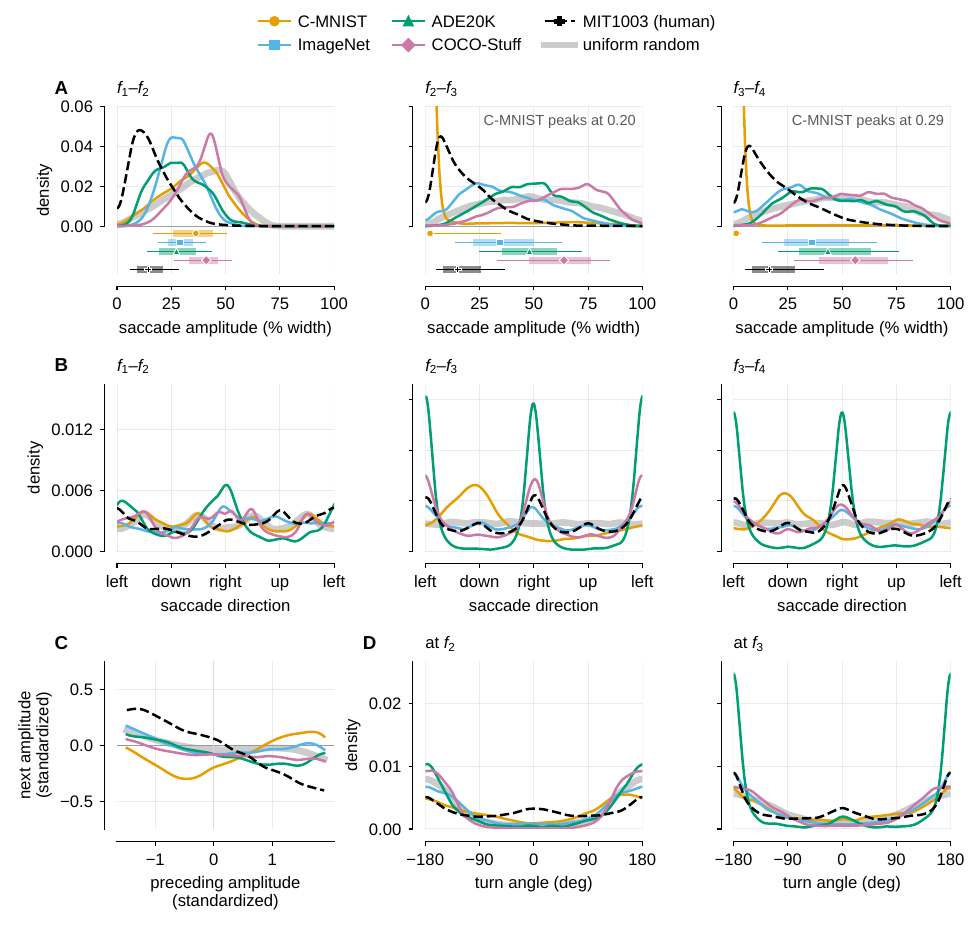}
\caption{ \textbf{Scanpath statistics of the learned policies.} Results are reported for a single seed, but remain consistent across training runs with different seeds (\cref{app:policy_stats_seeds}). Unless stated otherwise, metrics for our models are calculated for the standard training horizon of three fixations ($f_2, f_3, f_4$) after an initial center fixation ($f_1$). The human dataset (MIT1003) was also recorded with an initial center fixation and metrics were calculated for all scanpaths that consist of at least three fixations afterwards. 
\textbf{(A)}~Saccade amplitude across first three saccades. 
\textbf{(B)}~Saccade direction across first three saccades.
\textbf{(C)}~Mean amplitude of the next saccade as a function of the preceding one (amplitudes standardized within each scanpath). Our models were unrolled for nine fixations ($f_2\text{--}f_{10}$) starting from a center fixation ($f_1$). For MIT1003, we selected scanpaths with at least nine fixations after the center fixation and truncated them to these nine. The plot is based on the saccades between $f_2$ and $f_{10}$. The initial saccade from the center to $f_2$ is excluded because its starting point is fixed.
\textbf{(D)}~Turning angle between two consecutive saccade vectors, measured between the first and second saccade (at $f_2$) and between the second and third saccade (at $f_3$).}
\label{fig:policy_stats}
\end{figure}

Beyond task performance, we analyze the behavioral characteristics of the scanpaths generated by our models, contrasting them across tasks and comparing them against human scanpaths from MIT1003 \citep{judd2009learning}. These human scanpaths were recorded with a display diagonal of approximately $35^\circ$ of visual angle \citep{kummerer2022deepgaze}, roughly $28^\circ \times 21^\circ$ for typical $1024 \times 768$ pixel images. In contrast, our log-polar foveation architecture operates directly in pixel space without an explicit mapping to visual degrees. Consequently, this comparison should be regarded as an initial reference point to contextualize model behavior and motivate future research on human and model scanpaths. We first evaluate standard scanpath statistics following \citet{kummerer2022deepgaze}. We then quantify how consistent learned policies are across different random seeds for the same task, and how much policies differ across tasks. Finally, we examine the IoR characteristics of our trained policies.

\paragraph{Cumulative fixation distribution.}
A behavior common to all our trained models is to avoid fixating at the very edge of an image, reflected by the dark border in the cumulative fixation distributions (\cref{fig:intro}B). This can be explained by models learning to look around efficiently: Since the glimpse captures a region around the exact fixation point in sufficiently high resolution, fixating at the very edge would leave much of that informative region outside the image. Beyond this shared behavior, different tasks and datasets yield characteristic fixation distributions. For instance, segmentation models allocate fewer fixations to the top or bottom of an image, most notably on ADE20K, likely because these regions tend to contain less informative content such as sky or ground. Furthermore, segmentation models tend to explore peripheral regions, whereas ImageNet models concentrate fixations in a ring around the center, reflecting the object-centered nature of the dataset. A tendency to direct fixations towards the image center is well documented in humans and more pronounced than in our models (see \cref{app:shape_of_prior}), remaining robust across different visual inputs and tasks \citep{tatler2007central}. \Cref{app:shape_of_prior} shows these distributions for all three seeds, separately for each step, and with each policy applied to the images of the other datasets.

\paragraph{Saccade amplitude.}
We examine the distribution of saccade amplitudes \citep{bahill1975most} across fixations $f_2$ to $f_4$ following the initial center fixation $f_1$ (\cref{fig:policy_stats}A). For all models, the initial saccade from $f_1$ to $f_2$ is constrained in amplitude by the central starting position. After this initial saccade, classification models exhibit smaller amplitudes than segmentation models. This difference can be explained by their respective objectives: while classification requires focusing on a central object of interest, segmentation models need to obtain a comprehensive understanding of the scene. For C-MNIST in particular, exploration decreases sharply across fixations as the target digit is found in an increasing number of cases. Compared to our models, humans consistently make shorter saccades, which might be explained by several characteristics and constraints of human eye movements that favor shorter saccades and are not included in our model. First, saccade duration increases with amplitude \citep{bahill1975main}, prolonging the period during which vision is degraded \citep{vanbeers2008saccadic}. Second, larger saccades are subject to greater endpoint variability, making them less reliable \citep{vanbeers2008saccadic}. Third, in natural viewing, large gaze shifts are typically accomplished by combined eye and head movements \citep{bahill1975most,freedman2008coordination}. Lastly, short saccades in humans may serve local exploration or correct a preceding saccade, and tend to follow after shorter fixation durations \citep{tatler2008systematic}. Foveated vision has also been proposed to favor short saccades \citep{tatler2009prominence}, though the absence of similarly short saccades in our foveated models suggests that foveation alone might not be sufficient to produce such behavior.

\paragraph{Saccade amplitude autocorrelation.}
We study saccade amplitude autocorrelation \citep{tatler2008systematic,wilming2013saccadic} by computing the mean amplitude of the next saccade as a function of the preceding one, standardizing amplitudes using the mean and standard deviation within each scanpath to account for variability across images (\cref{fig:policy_stats}C). Due to the standardization, a negative association is observed even for a uniformly random policy when scanpaths are relatively short, growing stronger as length decreases (see \Cref{app:autocorr_by_length} for a more detailed analysis). We therefore benchmark against this random policy using scanpaths of nine fixations. On MIT1003 \citep{judd2009learning}, human observers display a clear negative association beyond that of the random policy. This finding aligns with prior work reporting a weak negative association between consecutive saccade amplitudes \citep{kummerer2022deepgaze,malemshinitski2020mathematical}, consistent with periods of local scanning separated by large relocations to new scene regions \citep{tatler2008systematic}. In contrast, our models exhibit no substantial negative association beyond that of the random policy. Thus, while our models do explore individual informative image regions in more detail (\Cref{fig:intro} and \Cref{app:devpanels}), they lack the human pattern of long saccades that relocate to a new region followed by short saccades that explore it. As short saccades following long ones may serve local exploration or correct a preceding saccade \citep{tatler2008systematic,kummerer2022deepgaze}, and tend to follow short fixations \citep{tatler2008systematic}, a framework that also models fixation durations might favor the emergence of this pattern.

\paragraph{Saccade direction.}
Regarding direction \citep{tatler2009prominence}, all models trained on natural images show a bias towards horizontal (left/right) and vertical (up/down) saccades, whereas diagonal saccades occur less frequently (\cref{fig:policy_stats}B). The strength of these biases differs between models: the horizontal bias is particularly strong for the ADE20K model, while the vertical bias is most pronounced for the ImageNet model. The most likely explanation for these biases is that information of interest in natural images is often arranged along horizontal or vertical axes. For example, objects frequently have a vertical orientation, which explains why the vertical bias is strongest for the object-centric ImageNet and why the synthetic C-MNIST dataset does not exhibit any horizontal or vertical biases. 
In the human scanpath literature, similar biases are well documented \citep{foulsham2008turning,kummerer2022deepgaze}. In fact, the pattern observed on MIT1003 closely resembles those that emerge naturally in a task-driven way in our models. Prior work likewise attributes the human horizontal bias partly to image content \citep{foulsham2008turning,kummerer2022deepgaze,reeves2026horizontal}. However, observer-related factors also contribute \citep{najemnik2008eye,reeves2022head,koevoet2025effort}. For instance, diagonal saccades require more effort than horizontal or vertical ones and therefore tend to be avoided by humans \citep{koevoet2025effort}. Our model serves as a flexible framework where such constraints could be added or omitted selectively to observe how different factors contribute to the observed pattern.

\paragraph{Saccade turn angle.}
We examine the turn angle between two consecutive saccades \citep{rothkegel2016influence,smith2009facilitation,wilming2013saccadic}, which represents the angular change in direction from one fixation to the next. Across all models, saccade probability increases with absolute turn angle, with exact reversals of $180^\circ$ being the most likely (\cref{fig:policy_stats}D). This tendency is particularly pronounced for the ADE20K model, which is also the only model that exhibits a slight continuation bias at $f_3$, favoring subsequent saccades in the same direction near $0^\circ$. On MIT1003, we also observe both biases, consistent with established findings in human scanpaths \citep{tatler2008systematic,wilming2013saccadic,kummerer2022deepgaze}. Reversals have been partly attributed to image boundaries \citep{wilming2013saccadic}: even a scanpath model based only on saliency and inhibition of return produces frequent turns near $180^\circ$, simply because reaching an edge naturally turns the gaze back \citep{schwetlick2020modeling}. In addition, humans tend to return to informative locations \citep{smith2009facilitation,wilming2013saccadic}. Continuation, in contrast, has been attributed to oculomotor bias \citep{smith2009facilitation,bays2012active}, small corrective saccades \citep{tatler2008systematic} and lingering presaccadic attention \citep{schwetlick2020modeling}, which stays at the same retinal position and thus points further along the same direction after a saccade has landed. For our models, image boundaries most likely explain the high rate of reversals, especially given their tendency towards large saccade amplitudes. The continuation bias on ADE20K likely reflects the model's tendency to explore images horizontally (\cref{fig:intro}B), which may also be one factor that contributes to the pattern observed in humans.

\begin{figure}[!t]\centering
\includegraphics[width=0.9\textwidth]{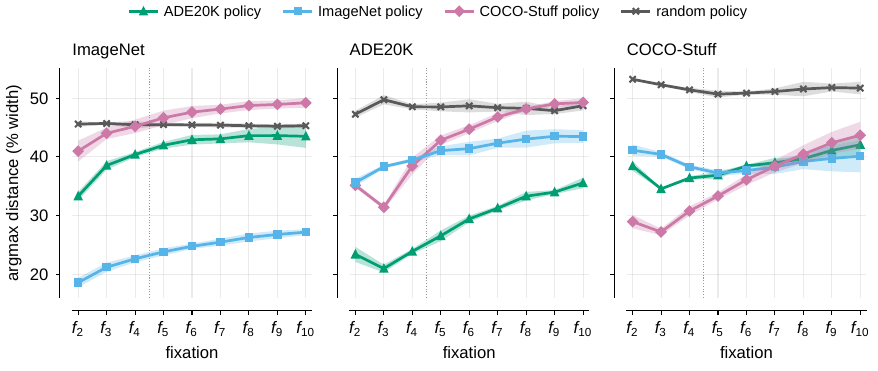}
\caption{\textbf{Argmax fixation distance.} Euclidean distance between fixations from the native reference model and fixations from evaluated models conditioned on the native model's fixation history, computed across nine fixations $f_2$--$f_{10}$ following the initial center fixation $f_1$ (see \Cref{sec:results_properties} for details). Distances are expressed as a percentage of image width (all images are square, so width equals height). Lower values indicate stronger policy alignment. Colored lines indicate comparison models (different tasks or the same task with different random seeds). The gray line denotes the expected distance when locations are drawn uniformly at random. Shaded bands show $\pm1$ std over three model pairs. The dotted line marks the $T = 4$ glimpses used in training.
}
\label{fig:argmax}
\end{figure}

\begin{figure}[!t]
  \centering
  \includegraphics[width=\textwidth]{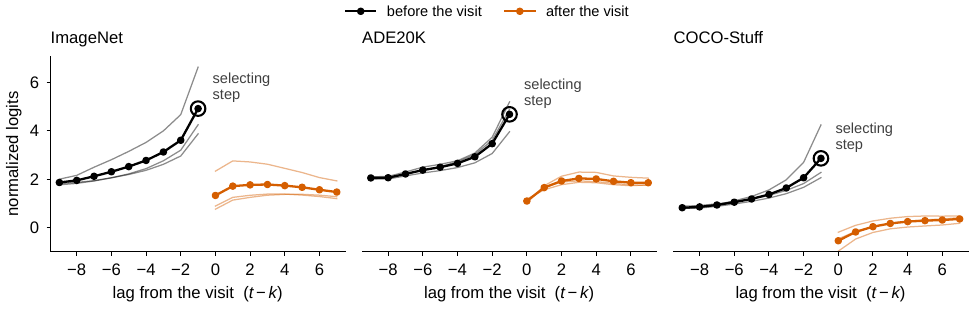}
  \caption{\textbf{Emergent inhibition of return.} 
z-standardized policy logits at visited locations as a function of relative lag $t - k$ (see \Cref{sec:results_properties} for details). Thin lines show individual models trained with different random seeds, while bold lines indicate their mean. Negative~/~positive lags correspond to steps before~/~after the fixation. Lower~/~higher logits reflect reduced~/~increased selection probability at that location.}
  \label{fig:inhibition}
\end{figure}

\paragraph{Argmax fixation distance.} 
Beyond aggregate scanpath statistics, we evaluate cross-task policy similarity and within-task consistency across random seeds using a conditional evaluation scheme adapted from scanpath modeling~\citep{kummererbethge2021sota, kummerer2022deepgaze}. In this scheme, models are scored on next-fixation predictions conditioned on identical fixation histories. While prior work performs the conditional evaluation using continuous predictions like log-likelihood, our policies are unregularized and highly peaked with clear maxima, making such density-based scores ill-conditioned. Instead, we evaluate spatial distance between predicted fixations directly under the following protocol: For each dataset, scanpaths are first generated by a native reference model trained on that dataset. Fixations are selected via argmax from policy probability maps across nine fixations $f_2\text{--}f_{10}$, starting from an initial center fixation $f_1$. These scanpaths are then fed as forced fixation sequences to models trained on (1)~other tasks or (2)~the same task with different random seeds. At each step, we measure the \emph{argmax distance} defined as the Euclidean distance between the argmax location predicted by that model's policy probability map and the reference fixation (\Cref{fig:argmax}).

Models trained on the same task across random seeds exhibit consistently lower argmax distances than models trained on different tasks, demonstrating stronger overall policy alignment. Nevertheless, within-task argmax distances remain non-trivial, measuring at least 20\% from $f_2$ onward. This suggests that no single canonical policy emerges, arguably because multiple distinct fixation strategies can effectively tackle the same task. Compared to a uniformly random policy, learned policies are in general more similar to each other, indicating that they share some general principles of where to look next given the same history of past fixations (e.\,g. saliency and inhibition of return). We also find that alignment decreases as forced fixation history grows, suggesting that the optimal next fixation becomes increasingly ambiguous. In \Cref{app:perglimpse_extra}, we compare policies based on two further similarity metrics and obtain consistent results.

\paragraph{Emergent inhibition of return.}
Human scanpaths naturally avoid looking back at recently visited locations, a mechanism known as inhibition of return \citep{itti1998model}. To test whether our models learn a similar suppression mechanism, we evaluate whether visiting a location depresses its policy logits at subsequent steps. Specifically, models trained with $T = 4$ glimpses are unrolled for an extended horizon of $T_\text{eval} = 10$ steps ($t = 1, \dots, 10$), starting from the initial center fixation $f_1$ at $t = 1$. We then track the policy logit at each fixation coordinate $f_k$ ($k \in \{2, \dots, 10\}$), which is visited at step $t = k$, across all steps $t$ of the episode. These values are then grouped by their relative lag $t - k$ to the visit and averaged over all fixations and images. To make values comparable across timesteps and images, we consider z-standardized logits over all spatial positions of the $64\times 64$ probability map.

For the three natural-image datasets, we find that the logits of a location $f_k$ gradually increase up until the point where the location is actually fixated (\Cref{fig:inhibition}). Right after the fixation, the logits for that location sharply decrease and then partially recover over the following steps, making a revisit more likely further into the future, a pattern similarly reported for human gaze behavior \citep{posner1984components}. The same pattern also holds for every visit index separately (\Cref{app:perifixation_split}).

\subsection{Challenges in joint task and policy optimization}
\label{sec:gradient_conflicts}

\begin{figure}[!t]
  \centering
  \includegraphics[width=\textwidth]{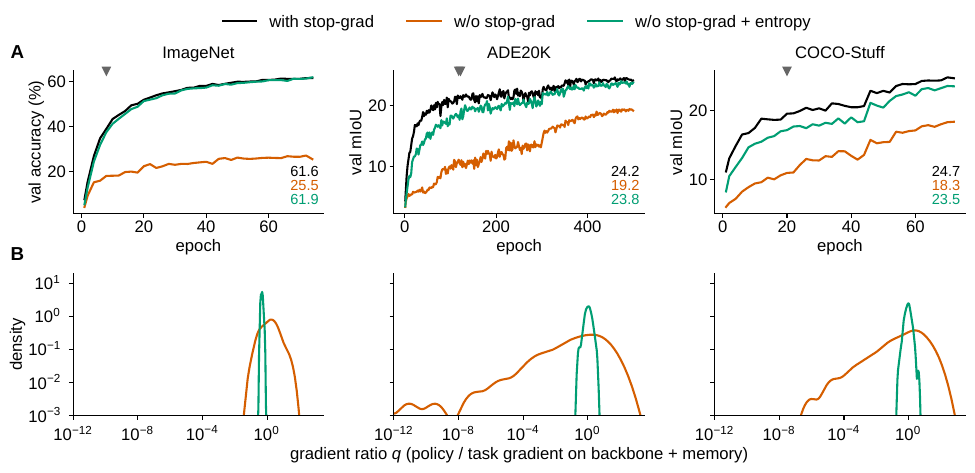}
  \caption{\textbf{Training dynamics under different optimization settings.}
  \textbf{(A)}~Task performance across models trained with stop-grad (see \cref{sec:model_training}), without stop-grad, and without stop-grad alongside entropy regularization. The triangle marks the checkpoint evaluated in panel (B), matched at roughly 150k training iterations. \textbf{(B)}~Distribution of the ratio $q$ of policy to task gradient magnitudes for backbone and memory parameters, as a kernel density estimate over $\log_{10} q$ with one value per minibatch, computed over 78 (ImageNet, batch size 64) or 312 (ADE20K and COCO-Stuff, batch size 16) training minibatches, i.\,e.\ about 5k images per model. The stop-grad model yields zero policy gradient to these components and is omitted.}
  \label{fig:joint_ratio_pdf}
\end{figure}

In primate vision, gaze is guided by priority maps represented in a dedicated oculomotor network of cortical and subcortical areas~\citep{fecteau2006salience, bisley2019neural},
that interacts with the ventral, object-recognition pathway in both directions~\citep{schall1995topography, moore2003selective}. In contrast, we train the policy on top of memory features that are optimized solely through the task loss (see \cref{sec:model_training}), as is also done in AdaptiveNN~\citep{wang_EmulatingHumanlikeAdaptive_2025}. As a result, FoAViS lacks a structural counterpart capable of self-organizing representations specifically for fixation selection. 
To allow the policy to shape intermediate representations, a natural first option is to let the policy gradient also backpropagate through the memory and backbone. We refer to this setting as \emph{joint} task and policy optimization, as both objectives then shape the backbone and memory. We experimented with this setting extensively and consistently observed optimization issues that we aim to raise awareness of in the following. These issues can be clearly illustrated by ablating the stop-grad operation while keeping all other components and objectives from \Cref{sec:method} fixed. In that case, we observe substantially degraded task performance across ImageNet, ADE20K, and COCO-Stuff compared to our default stop-grad setting (\cref{fig:joint_ratio_pdf}A). A likely cause is high variability in the policy gradient without stop-grad (\cref{fig:joint_ratio_pdf}B), though the precise underlying mechanism requires further investigation (see \Cref{app:detach_ablation} for a more detailed analysis). Empirically, we find that most of the performance can be recovered by adding entropy regularization to the probability distribution output by the policy head (see \Cref{app:entropy_new} for implementation details), which also stabilizes the relative magnitude between policy and task gradient. However, task performance under entropy regularization still tends to fall short of our default stop-grad setting (\cref{fig:joint_ratio_pdf}A), suggesting that the optimization issues are not fully resolved. At the same time, the resulting policies slightly exceed the performance of their stop-grad counterparts when transferred to a model trained on random locations (\Cref{tab:xpol_jointabl}). Our findings highlight the complexity of joint task and policy optimization, where performance trade-offs can be subtle, as illustrated by the entropy-regularized setting. Further research is needed to explore both architectural designs and optimization algorithms that enable stable, efficient, and effective learning of task and policy in active vision systems.

\section{Conclusion}
\label{sec:conclusion}

\subsection{Summary}
We presented FoAViS, an active vision model that consists of log-polar foveation, a backbone and 2D spatial memory, as well as a task and policy head (\Cref{sec:fov,sec:backbone,sec:memory,sec:sampling_mechanism,sec:task_readout}), and systematically studied its gaze behavior across different tasks. The system can simultaneously learn where to look and what it sees (\cref{sec:model_training}). We found that the learned policy selects more informative regions than random or heuristic baselines (\cref{sec:results_native}). Furthermore, the model's representations develop concurrently with its own learned policy, having a positive effect on model performance (\cref{sec:results_transfer}). Beyond task performance, our models exhibited structured, task-specific gaze behaviors that reproduce key aspects of human scanpath statistics while also showing notable divergences (\cref{sec:results_properties}). These empirical patterns, along with phenomena such as inhibition of return, emerged naturally from concurrent task-policy optimization without explicit supervision, providing a normative perspective on gaze behavior. Finally, we drew attention to optimization challenges encountered during joint task and policy training, raising the broader question of how to structure and optimize multi-component active vision systems (\cref{sec:gradient_conflicts}).

\subsection{Limitations and outlook}
FoAViS is built on top of a simple Res-UNet with roughly 13.5M parameters and is trained from scratch with a relatively low compute budget. Unsurprisingly, it therefore trails the raw performance of higher-capacity models that leverage larger, more powerful model architectures \citep[AdaptiveNN;][]{wang_EmulatingHumanlikeAdaptive_2025}, more extensive training, and/or strong distilled features \citep[CanViT; ][]{canvit2026activevision} as a starting point (see also \Cref{app:prior_work}). Our model would certainly benefit from integrating such techniques to obtain stronger downstream task performance, as a faithful computational model of primate vision should competently solve a broad range of visual tasks. However, evaluating active vision models purely on task performance relative to compute neglects other critical axes along which such systems need to be judged when serving as models of biological vision. Specifically, we identify three key dimensions: (1)~architectural alignment with biological vision (foveation, spatial memory, and cortical areas), (2)~behavioral alignment with human scanpaths, and (3)~representational alignment, which has so far primarily been studied in single-pass vision models.

\paragraph{Architectural alignment.}
Our model implements a log-polar image sampling and learns both image-level and dense vision tasks concurrently with an active vision policy. While these are important properties for an active vision system, our architecture still lacks several core characteristics of biological vision. First, our log-polar sampling approximates retinal sampling only qualitatively, sacrificing local isotropy to map the foveated representation onto a Cartesian grid for standard convolutions. Recent work provides more realistic foveation mechanisms that could be used instead~\citep{blauch2026fovi}. A further limitation is that our model has no notion of time: every glimpse receives the same amount of computation. In humans, fixation durations vary and partly adapt to the processing demands of the currently fixated region~\citep{henderson2008eye,nuthmann2010crisp,wilming2013saccadic}. This temporal dimension likely shapes several of the scanpath statistics analyzed in \Cref{sec:results_properties}~\citep{tatler2008systematic,smith2009facilitation,wilming2013saccadic}. To address this, the policy could explicitly predict fixation duration, using architectural mechanisms like variable network depth to allocate compute dynamically. Furthermore, our model currently encodes each glimpse strictly bottom-up before merging it with the 2D scene memory. Consequently, the backbone cannot leverage top-down feedback from information accumulated across previous glimpses, a core characteristic of primate vision~\citep{bar2006top,gilbert2013top}. Recent work has started to explore top-down feedback in an active vision architecture~\citep{canvit2026activevision}. Moreover, our 2D spatial memory is designed for static scenes. When dealing with dynamic environments, keeping an exhaustive map is neither practical nor biologically plausible, leaving open what such a memory should store and in which reference frame. Finally, our architecture does not faithfully reflect distinct cortical areas and their communication, a factor critical for representational alignment.

\paragraph{Behavioral alignment.}
Prior work has established that training fixation policies concurrently with global or dense tasks is feasible across a wide range of foveation mechanisms and architectures~\citep{mnihRecurrentModelsVisual2014,baMultipleObjectRecognition2015,pardyl_AdaGlimpseActiveVisual_2025,wang_EmulatingHumanlikeAdaptive_2025}, which is further corroborated by our results. However, core architectural and optimization questions remain open: Which parts of the network should be responsible for the task prediction, for fixation selection, or for both, and how should information flow between them, potentially inspired by the dorsal and ventral streams? How do we optimize these objectives together without running into optimization instability? And what algorithms should we use to train policies for various tasks in the most efficient and effective way? Answering these questions will bring us closer to achieving an unbiased normative account of gaze behavior under different task constraints, providing a powerful tool to study similarities and differences between machine and human gaze behavior. Observed behavioral differences can serve as guidance to uncover systematic biases and inefficiencies as well as their underlying causes, which may in turn help us design systems that behave more like humans. Notable differences could potentially be reverse-engineered into a machine vision architecture through dedicated pretraining schemes, structural changes, or additional loss terms. Accurate behavioral alignment is arguably also important for adequate representational alignment, as the incoming visual inputs shape how representations emerge in the brain.

\paragraph{Representational alignment.}
While correspondence between deep artificial networks and the ventral visual stream has been widely studied in passive models~\citep{yamins2014,cadena2024}, researchers have only recently begun addressing neural questions in active vision~\citep{pourrahimi_EmergentBrainlikeRepresentations_2025,thorat2025predicting,sulewski2026active}. A primary obstacle has been the scarcity of datasets that jointly record gaze trajectories and multi-area neural population activity during unconstrained active viewing. Future experiments in the visual system of non-human primates should aim to close this gap, which could pave the way for a new generation of computational models of the primate visual system. With our work, we aim to provide a foundation for developing such a system, while also highlighting the challenges that remain on the way. Ultimately, this line of research opens a path towards comprehensive in silico models, enabling systematic manipulations to investigate the interactions between eye movements, internal representations, and behavioral goals in biological vision.

\subsubsection*{AI Usage Statement}
We used generative models and agentic tools for writing and debugging code, running experiments, and editing the manuscript. We thoroughly verified all results produced by these models and take full responsibility for the content of this paper.

\makeatletter
\if@accepted
\subsubsection*{Acknowledgments}%
Funded by the Deutsche Forschungsgemeinschaft (DFG, German Research Foundation) -- GRK2906 -- project number 502807174.
Yoha\"i-Eliel Berreby was funded by a Unifying Neuroscience and Artificial Intelligence - Qu\'ebec (UNIQUE) PhD Excellence Scholarship and by a Natural Sciences and Engineering Research Council of Canada (NSERC) Canada Graduate Research Scholarship -- Doctoral program.
Haoze Zhu was funded by the German Federal Ministry of Research, Technology and Space (BMFTR) under grant 01GQ2501A (PRISMA-V).
The authors gratefully acknowledge the computing time granted by the Resource Allocation Board and provided on the supercomputer Emmy/Grete at NHR-Nord@G\"ottingen as part of the NHR infrastructure.
The calculations for this research were conducted with computing resources under the project nim00018.
\fi
\makeatother

\bibliographystyle{tmlr}
\bibliography{literature}

\appendix
\crefalias{section}{appendix} 
\crefalias{subsection}{appendix}
\counterwithin{figure}{section}
\counterwithin{table}{section}
\clearpage

\section{Implementation details}
\label{app:implementation_details}
This appendix provides further information about the architecture and training procedure. The architecture is presented in more detail in \Cref{app:architecture}. Quantitative characteristics of the log-polar sampling are presented in \Cref{app:logpolar_density}. Further information on the training and evaluation protocol is given in \Cref{app:training_details}.

\subsection{Architecture}\label{app:architecture}
\Cref{tab:architecture_backbone,tab:architecture_heads} summarize the
backbone and the recurrent memory with its task heads. The ImageNet classification model has about $15.3$M parameters: $13.5$M backbone, $1.2$M
classification head, $0.44$M policy head and $0.13$M recurrent memory. The C-MNIST model has $15.0$M
($0.95$M classification head), and the semantic segmentation models have $14.2$M ($0.17$M
segmentation head). The backbone uses instance normalization
throughout to cope with input statistics that differ across glimpses.

\begin{table}[tp]
\centering
\small
\caption{\textbf{Backbone architecture.} The backbone takes a log-polar
  transformation of the input image. Each ResBlock uses $3\times3$ convolutions,
  with a $1\times1$ projection on the skip connection where needed. In a ResBlock, each
  convolution is followed by Instance Normalization and the first also by a ReLU; the
  second ReLU follows the residual addition. The
  bottleneck operates at the encoder's final resolution. In the decoder,
  each ``Conv $1\times1$ fuse'' row concatenates the incoming features with the
  corresponding encoder feature map (channel counts in parentheses) and
  projects them to the decoder width, again followed by
  Instance Normalization and ReLU.}
\label{tab:architecture_backbone}
\begin{tabular}{@{}llccc@{}}
\toprule
Component & Layer & Res.\ change & Out ch. & Output size \\
\midrule
\multirow{8}{*}{Encoder}
 & ResBlock (stride 1) & --           & 64  & $64\times64$ \\
 & ResBlock (stride 1) & --           & 64  & $64\times64$ \\
 & ResBlock (stride 2) & $\downarrow2$ & 128 & $32\times32$ \\
 & ResBlock (stride 1) & --           & 128 & $32\times32$ \\
 & ResBlock (stride 2) & $\downarrow2$ & 256 & $16\times16$ \\
 & ResBlock (stride 1) & --           & 256 & $16\times16$ \\
 & ResBlock (stride 2) & $\downarrow2$ & 512 & $8\times8$ \\
 & ResBlock (stride 1) & --           & 512 & $8\times8$ \\
\midrule
Bottleneck
 & ResBlock (stride 1) & --           & 128 & $8\times8$ \\
\midrule
\multirow{11}{*}{Decoder}
   & Conv $1\times1$ fuse ($128\!+\!512$) & --          & 128 & $8\times8$ \\
   & ResBlock (stride 1)                  & --          & 128 & $8\times8$ \\
   & UpConv (stride 2)                    & $\uparrow2$ & 128 & $16\times16$ \\
   & Conv $1\times1$ fuse ($128\!+\!256$) & --          & 128 & $16\times16$ \\
   & ResBlock (stride 1)                  & --          & 128 & $16\times16$ \\
   & UpConv (stride 2)                    & $\uparrow2$ & 128 & $32\times32$ \\
   & Conv $1\times1$ fuse ($128\!+\!128$) & --          & 128 & $32\times32$ \\
   & ResBlock (stride 1)                  & --          & 128 & $32\times32$ \\
   & UpConv (stride 2)                    & $\uparrow2$ & 128 & $64\times64$ \\
   & Conv $1\times1$ fuse ($128\!+\!64$)  & --          & 128 & $64\times64$ \\
   & ResBlock (stride 1)                  & --          & 128 & $64\times64$ \\
\midrule
Inverse log-polar
   & Log-polar $\to$ Cartesian $64\times64$ grid (bilinear) & -- & 128 & $64\times64$ \\
\bottomrule
\end{tabular}

\vspace{1.0em}

\caption{\textbf{Recurrent memory and task heads.} The recurrent memory takes its past state and the current backbone output. It is an LSTM cell applied independently at each of the $64\times64$ spatial positions
and unrolled over the $T$ glimpses; the weights are shared across positions, and each position
carries its own state. The policy, classification, and segmentation
heads take the memory features $\mathbf{H}_t$ as input, the policy head through a stop-gradient. $K$ is the number of classes.}
\label{tab:architecture_heads}
\begin{tabular}{@{}llccc@{}}
\toprule
Component & Layer & Res.\ change & Out ch. & Output size \\
\midrule
Recurrent memory
 & Standard LSTM cell applied at every position & -- & 128 & $64\times64$ \\
\midrule
\multirow{4}{*}{Policy head}
 & Conv $3\times3$ + ReLU & -- & 128 & $64\times64$ \\
 & Conv $3\times3$ + ReLU & -- & 128 & $64\times64$ \\
 & Conv $3\times3$ + ReLU & -- & 128 & $64\times64$ \\
 & Conv $1\times1$ $\to$ probability map & -- & 1 & $64\times64$ \\
\midrule
\multirow{5}{*}{Classification head}
 & Conv $3\times3$ (stride 2) + ReLU & $\downarrow2$ & 256 & $32\times32$ \\
 & Conv $3\times3$ (stride 1) + ReLU & --            & 256 & $32\times32$ \\
 & Global avg-pool                   & $\downarrow32$            & 256 & $1\times1$ \\
 & Linear + ReLU                     & --            & 256 & $1\times1$ \\
 & Linear $\to$ classes              & --            & $K$ & $1\times1$ \\
\midrule
\multirow{3}{*}{Segmentation head}
 & Conv $3\times3$ + ReLU        & --         & 128 & $64\times64$ \\
 & Conv $1\times1$ $\to$ classes & --         & $K$ & $64\times64$ \\
 & Bilinear upsample ($\times8$) & $\uparrow8$ & $K$ & $512\times512$ \\
\bottomrule
\end{tabular}
\end{table}

\subsection{Details on the retinal layout of a glimpse}
\label{app:logpolar_density}
This subsection describes how the foveation module (\Cref{sec:fov}) distributes its samples over the image. Each
glimpse has $64 \times 64$ samples: $64$ rings spaced evenly in $\log r$, from $r = 1$\,px to the image width $S$,
with $64$ samples spaced evenly in angle on each ring. Because the rings are spaced evenly in $\log r$, the
distance between neighboring samples grows in proportion to the distance from the fixation.

Close to the fixation, neighboring samples are less than a pixel apart, so the image is oversampled: up to
$11$\,px from the fixation for $224 \times 224$ images and $10$\,px for $512 \times 512$ images. This disk covers
less than $1\%$ of the image (about $360$ pixels) but receives $44\%$ of the samples, so about $1430$ of the
$4096$ samples ($35\%$) read pixels that are already sampled ($29\%$ for $512 \times 512$). We did not optimize
this allocation, since the exact design of the foveation is not the focus of this work. A different sampling scheme would leave the principles of the architecture unchanged. Beyond the disk,
neighboring samples are more than a pixel apart. To reduce aliasing there, each sample is the mean of a
$4 \times 4$ block of a grid sampled $4\times$ finer, rather than a single read at its position. Because the
outermost ring has radius $S$, part of the grid falls outside the image and reads zeros: $11\%$ of the samples
for a fixation at the image center ($10\%$ for $512 \times 512$), $53\%$ at the middle of an edge and $75\%$ in
a corner.

\begin{figure}[t]
\centering
\includegraphics[width=0.65\textwidth]{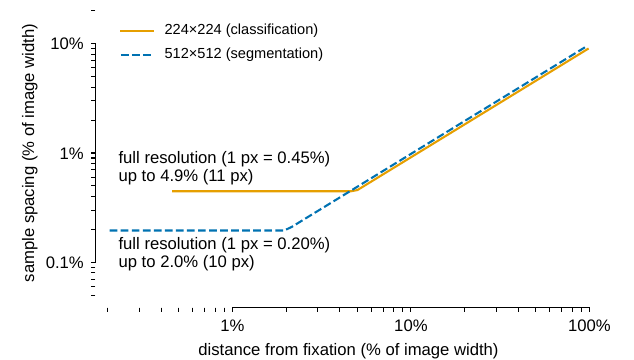}
\caption{\textbf{Sampling density of the glimpse} at the classification
($224 \times 224$) and segmentation ($512 \times 512$) resolutions. The spacing between neighboring
samples, roughly the smallest detail a glimpse resolves, is plotted against the distance from the
fixation, both as a percentage of the image width. Up to $11$\,px ($224$) and $10$\,px
($512$) from the fixation, the image is oversampled: neighboring samples are less than a pixel apart. The
plot floors the spacing at one pixel there (plateaus, labeled \emph{full resolution}), since samples closer
than a pixel add no detail.}
\label{fig:logpolar_density}
\end{figure}

Both resolutions follow the same rule. Because the rings run from $1$\,px to $S$, the gap between neighboring
rings at a given distance from the fixation depends on the image size only through $\log S$, which grows by only
$15\%$ from $224$ to $512$ although the image is more than twice as wide. At the same distance from the fixation
in pixels, the rings of the $512$ grid are therefore only about $15\%$ further apart, and the spacing along a ring
is the same, so in pixels the two grids sample the image nearly alike. Relative to the image they differ, because
a pixel is a smaller part of a larger image: the oversampled disk covers $2.0\%$ of the image width at $512$ and
$4.9\%$ at $224$ (\Cref{fig:logpolar_density}).

\subsection{Training and evaluation details}
\label{app:training_details}

\paragraph{Optimization.}
\Cref{tab:hparams} lists the hyperparameters. The learning rate rises linearly during warmup, stays constant and is then
halved at fixed intervals.

\begin{table}[t]
\centering
\small
\caption{\textbf{Training hyperparameters.} Random initialization uses PyTorch's defaults. The segmentation models take over the backbone and memory of the ImageNet random-location model; their policy and segmentation heads are newly initialized.}
\label{tab:hparams}
\setlength{\tabcolsep}{5pt}
\begin{tabular}{lcccc}
\toprule
Hyperparameter & ImageNet & C-MNIST & ADE20K & COCO-Stuff \\
\midrule
Initialization          & random & random & \multicolumn{2}{c}{ImageNet random-location model} \\
Optimizer               & \multicolumn{4}{c}{AdamW, weight decay $0.01$ (all parameters)} \\
$\beta_1, \beta_2, \epsilon$ & \multicolumn{4}{c}{$0.99,\ 0.999,\ 10^{-8}$} \\
Learning rate           & $10^{-4}$ & $10^{-4}$ & $2 \times 10^{-4}$ & $2 \times 10^{-4}$ \\
Warmup                  & \multicolumn{4}{c}{$2000$ steps (linear)} \\
Epochs                  & $150$ & $400$ & $500$ & $72$ \\
LR constant for (epochs) & $89$ & $233$ & $300$ & $44$ \\
then halved every (epochs) & $13$ ($5\times$) & $33$ ($6\times$) & $50$ ($4\times$) & $8$ ($4\times$) \\
GPUs                    & $4$ & $1$ & $1$ & $1$ \\
Batch size (total)      & $64$ ($16$ per GPU) & $64$ & $16$ & $16$ \\
Input resolution        & $224 \times 224$ & $224 \times 224$ & $512 \times 512$ & $512 \times 512$ \\
Glimpses ($T$)          & $4$ & $4$ & $4$ & $4$ \\
Gradient-norm clip      & $20$ & $20$ & $20$ & $20$ \\
Precision               & bf16 mixed & bf16 mixed & fp32 & fp32 \\
\bottomrule
\end{tabular}
\end{table}

\paragraph{Fixations and policy.}
The first fixation is the image center. During training, each further fixation is drawn from the policy's probability map
(\Cref{eq:policy}) and placed uniformly at random inside the drawn cell of the $64\times64$ grid. The rewards are standardized
per step over the samples on each GPU (\Cref{eq:reward_norm}; $16$ samples for ImageNet due to multi-GPU training, the whole batch otherwise). In the
random-location setting, every fixation after the first is drawn uniformly from the whole image, and the policy head is
neither used nor trained. Validation during training draws fixations as above; reported task
metrics are instead evaluated with the center of the most probable cell as fixation (argmax, see Evaluation below).

\paragraph{Data.}
ImageNet images are normalized with ImageNet statistics. In training, they are randomly resized and cropped to $224 \times 224$
(scale $[0.08, 1]$) and flipped horizontally; in evaluation, the shorter side is resized to $256$ and the central
$224 \times 224$ crop is taken. C-MNIST canvases (\Cref{app:cmnist}) are normalized with MNIST statistics and not augmented.
Segmentation images are augmented in training as follows. A rectangular region covering $40$ to $100\%$ of the image area,
with a width-to-height ratio between $0.75$ and $1.33$, is cut out at a random position and resized to $512 \times 512$.
The region is thus stretched or squashed by up to a factor of $1.33$, and its aspect ratio is not preserved. The result is flipped horizontally with probability $0.5$, color-jittered (brightness, contrast
and saturation $0.4$, hue $0.1$) with probability $0.8$, converted to grayscale with probability $0.1$, and rotated by a
random angle of up to $\pm 10^{\circ}$, which leaves the corners empty (filled with label $0$, see below). In evaluation,
the whole image is resized to $512 \times 512$, again without preserving the aspect ratio. Label maps undergo the same geometric transforms, with nearest-neighbor interpolation.

\paragraph{Segmentation loss and labels.}
For segmentation, the task loss and the reward (\Cref{eq:task_loss,eq:reward}) use the per-pixel cross-entropy, averaged over the image.
Unlabeled pixels (label $0$ in ADE20K, $255$ in COCO-Stuff) are excluded from loss, reward and metrics. ADE20K has
$150$ classes and COCO-Stuff $171$.
For COCO-Stuff, two settings were by mistake taken over from datasets in which label $0$ is a background class. In
COCO-Stuff, label $0$ is instead the class \emph{person}, while unlabeled pixels have label $255$ and are excluded as
described above. (1)~Label $0$ is weighted by $0.1$ in the cross-entropy loss (a setting from an earlier setup in which
the background class was down-weighted by this factor instead of excluded, applied here by mistake). In COCO-Stuff, this
down-weights person pixels, in the task loss and hence in the reward. (2)~The image corners that the training-time
rotation leaves empty are labeled $0$, i.e.\ as background. In COCO-Stuff, they are labeled as person. Both mistakes
affect all COCO-Stuff models in the same way, so comparisons between these models remain valid, and we therefore did
not retrain them. Evaluation is not affected.

\paragraph{Evaluation.}
Every reported task metric comes from the last checkpoint of a run, evaluated with argmax fixations on the full validation set
(ImageNet \num[group-digits=integer]{50000}, C-MNIST \num[group-digits=integer]{10000}, ADE20K \num[group-digits=integer]{2000}
and COCO-Stuff \num[group-digits=integer]{5000} images). Classification is scored by accuracy. Segmentation is scored by mIoU and pixel accuracy, computed from one confusion
matrix accumulated over the whole validation set; mIoU averages over the classes that occur in the labels or the
predictions. Both are computed at $512 \times 512$, the resolution the models see, not at the original image resolution.
Because every image is squashed to a square, this differs from the standard protocol, which resizes the predicted logits
to each image's native resolution before scoring. Scored that way, mIoU changes by $-0.5$ on ADE20K and by $+0.1$ on
COCO-Stuff. The shift is almost the same for models trained with a learned policy and with random locations, so their
difference barely moves (from $2.49$ to $2.59$~mIoU on ADE20K and from $1.76$ to $1.77$ on COCO-Stuff), and since all models
are scored the same way, no comparison in this paper changes. \Cref{tab:prior_work}, which compares with published numbers,
uses the native-resolution score. Where a cell reports mean $\pm$ std, it is over three models trained with different random
seeds. Some analyses in \Cref{app:detach_ablation,app:loss_choices} deviate from the evaluation protocol described above (see \Cref{app:regimes} for details).

\paragraph{Compute.}
One ImageNet model trains in about three days on four A100 or H100 GPUs, one ADE20K or COCO-Stuff model in about two days and one
C-MNIST model in about $20$ hours, each on a single GPU.

\clearpage
\section{Cluttered MNIST}
\label{app:cmnist}
C-MNIST (\Cref{sec:exp_datasets}) is generated on the fly.
Each image is a $224 \times 224$ single-channel canvas ($224 = 8 \times 28$) with a black background; there is no stored pool of images. The target is one MNIST digit at its native $28 \times 28$
size, placed at a uniformly random position that keeps it fully inside the canvas (top-left corner uniform in
$\{0, \dots, 196\}^2$), and its digit class is the label, giving $10$ classes. The target covers
$28^2 / 224^2 \approx 1.6\%$ of the canvas. Five clutter characters are added, each a $28 \times 28$ letter drawn
uniformly from EMNIST letters and placed independently in the same way; letters are never a class. Target and clutter
are combined by a pixel-wise maximum, without rejecting overlaps, so letters can partly cover the target and each
other.

\begin{figure}[t]
\centering
\includegraphics[width=\textwidth]{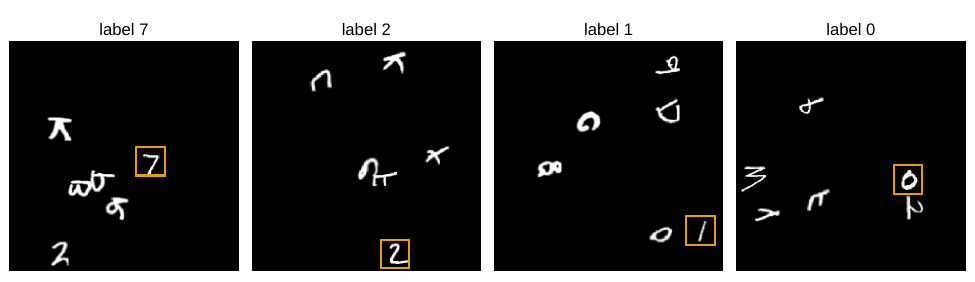}
\caption{\textbf{C-MNIST validation images}, the first four of the validation set, with the target digit boxed and its
label above. The other characters are EMNIST letters.}
\label{fig:cmnist_examples}
\end{figure}

Training images take their targets from the MNIST training split (\num[group-digits=integer]{60000} images) and their
letters from the EMNIST letters training split. Position and clutter are drawn anew every time a sample is loaded, so
each of the $400$ training epochs, of \num[group-digits=integer]{60000} images each, sees new images. Validation images
use the MNIST test split (\num[group-digits=integer]{10000} images) and the EMNIST letters test split. Each is a
deterministic function of its sample index, which seeds the choice and placement of its clutter, so all evaluations and
models see identical validation images. Pixel values are scaled to $[0, 1]$ and normalized with the MNIST mean and standard deviation ($0.1307$ and
$0.3081$); no augmentation is applied.
\Cref{fig:cmnist_examples} shows four validation images.

\clearpage
\section{Passive baseline}
\label{app:passive_baseline}
\Cref{tab:main_combined_cur} compares the active models with a passive baseline that sees the whole image at once.
The passive baseline is the non-foveated counterpart of FoAViS. It follows the design of our backbone and uses
the same data pipeline (splits, image sizes and augmentations) as the active models, and differs from them in seeing
the whole image in a single forward pass, without foveation, memory or policy.

For classification, the baseline is our full Res-UNet backbone (\Cref{tab:architecture_backbone}) followed by the
classification head of the active model (\Cref{tab:architecture_heads}), which reads the backbone output instead of
the memory ($14.5$M parameters for C-MNIST and $14.7$M for ImageNet, against $15.0$M and $15.3$M for the active
models, which add memory and policy head; \Cref{app:architecture}). For segmentation, the baseline is the same
backbone with the segmentation head of the active model ($13.7$M parameters, against $14.2$M).

Both baselines receive the image downsampled to $128 \times 128$, the pixel count of four glimpses, so they use as
many pixels and as much backbone computation as four glimpses of the active model (but see \cref{app:logpolar_density} where foveal oversampling is quantified which leads to fewer unique pixels seen in FoAViS). The Res-UNet predicts at its input
resolution, so the task head reads a $128 \times 128$ grid, twice the resolution of the $64 \times 64$ memory of the
active model; the segmentation logits are upsampled to $512 \times 512$ in the same way. Like the active segmentation models, the segmentation baselines are initialized from an ImageNet model, here the
classification baseline: its backbone (encoder and decoder) initializes the Res-UNet, and only the segmentation head
starts from random initialization. The active segmentation models additionally take over the memory of the ImageNet
random-location model (\Cref{tab:hparams}).

All baselines are trained with the same learning rate schedule, number of epochs and batch
size as the corresponding active model (\Cref{tab:hparams}): AdamW (weight decay $0.01$) and
cross-entropy, a learning rate of $10^{-4}$ (classification) or $2 \times 10^{-4}$ (segmentation)
with a linear warmup over the first $2000$ steps, held constant and then halved at the same epochs
as for the active model, for $400$ (C-MNIST), $150$ (ImageNet), $500$ (ADE20K) and $72$
(COCO-Stuff) epochs, with batch size $64$ (classification) or $16$ (segmentation) on the same number of GPUs.
The baselines use AdamW with $\beta_1 = 0.9$ (PyTorch's default) instead of the active models' $0.99$.

\clearpage
\section{Published results of prior active vision models}
\label{app:prior_work}
\Cref{tab:prior_work} places our results among published active vision models. Each model is listed with its best
published result, together with the input resolution, number of glimpses and pixel budget at which it was obtained.

\begin{table}[t]
  \centering
  \caption{\textbf{Published results of prior active vision models on ADE20K and ImageNet.}
  Each model is listed with its \emph{best published} result. These works cannot be compared in a controlled way: backbone,
  initialization, glimpse budget, input resolution and evaluation protocol all differ. \emph{Pixels}: pixels used for the
  task, summed over all glimpses and divided by the pixels of the model's own input image (\emph{Input res.}). Pixels are counted as delivered, after any down-scaling, not by the area they cover. A glance or overview
  pass, if a model uses one, counts at its own size and is included in \emph{Glimpses}. Overlapping glimpses are not
  de-duplicated, so a budget can exceed $100\%$. Two values in a cell are
  segmentation / classification. A dash marks a benchmark the model does not report.}
  \label{tab:prior_work}
  \small
  \setlength{\tabcolsep}{4pt}
  \resizebox{\textwidth}{!}{%
  \begin{tabular}{llllcccc}
    \toprule
    \textbf{Model} & \textbf{Backbone} & \textbf{Init.} & \textbf{Input res.} & \textbf{Pixels (\%)} & \textbf{Glimpses}
      & \textbf{ADE20K mIoU (\%)} & \textbf{ImageNet Acc (\%)} \\
    \midrule
    DRAM~\citep{baMultipleObjectRecognition2015}                  & ResNet-50         & ImageNet (3-stage)     & $224^2$         & 94.5$^{\mathrm{a}}$ & 8       & --  & 67.5$^{\mathrm{b}}$ \\
    Saccader~\citep{elsayed_SaccaderImprovingAccuracy_2019}       & NASNet            & ImageNet (3-stage)     & $331^2$         & 69.9$^{\mathrm{a}}$ & 6       & --  & 75.0 \\
    GFNet~\citep{wangGlanceFocusDynamic2020}                      & ResNet-50         & ImageNet               & $224^2$         & 91.8        & 5       & --  & 75.9 \\
    TNet~\citep{papadopoulosHardAttentionScalable2021}            & BagNet-77         & from scratch           & $224^2$         & 70.9        & 6       & --  & 74.6 \\
    FoveaTer~\citep{jonnalagadda2023foveater}                     & PatchConvNet      & ResNet-18 + DeiT-S     & $224^2$         & \emph{n/a}  & 3       & --  & 76.7 \\
    \citet{lukanov_BiologicallyInspiredDeep_2021}              & Xception-based CNN & from scratch           & $256^2$         & 10.0$^{\mathrm{a}}$ & 1       & --  & 65.2$^{\mathrm{c}}$ \\
    STAM~\citep{rangrej2022stam}                                  & DeiT-B            & DeiT (distilled)       & $224^2$         & 55.1        & 27      & --  & 80.8 \\
    AME (MAE)~\citep{pardyl2023active}                            & ViT-L             & MAE                    & $128\times256$  & 18.8        & 8       & 24.4 & -- \\
    AME (SETR)~\citep{pardyl2023active}                           & ViT-L             & SETR                   & $128\times256$  & 18.8        & 8       & 27.6 & -- \\
    \citet{liu_ImprovedEfficiencyBased_2023}                    & ResNet-101        & ImageNet (not tuned)   & $224^2$         & 41.3        & 5       & --  & 73.7 \\
    AdaGlimpse~\citep{pardyl_AdaGlimpseActiveVisual_2025}         & ViT-B             & MAE + glimpse pretr.   & $224^2$         & 36.7 / 28.6 & 8 / 14  & 25.7 & 77.5 \\
    AdaptiveNN~\citep{wang_EmulatingHumanlikeAdaptive_2025}       & DeiT-S            & from scratch           & $288^2$         & 62.8$^{\mathrm{d}}$ & 4.15$^{\mathrm{d}}$ & --  & 82.2 \\
    CanViT-B~\citep{canvit2026activevision}                       & ViT-B             & DINOv3 distillation    & $512^2$         & 131 / 100   & 21 / 16 & 45.9 & 84.5 \\
    \midrule
    FoAViS (ours), ImageNet model                                & Res-UNet          & from scratch           & $224^2$         & 32.7        & 4       & --  & 68.5 \\
    FoAViS (ours), ADE20K model                                  & Res-UNet          & ImageNet$^{\mathrm{e}}$     & $512^2$         & 6.3         & 4       & 23.8 & --  \\
    \bottomrule
  \end{tabular}%
  }
  \par\vspace{3pt}
  {\footnotesize $^{\mathrm{a}}$Only the glimpses that reach the classifier are counted; glimpse locations are chosen from a view of the whole image (DRAM, Saccader) or from an earlier glimpse (Lukanov).
  $^{\mathrm{b}}$Re-implementation by the Saccader authors, as reported by TNet.
  \emph{n/a}: FoveaTer foveates the convolutional feature map of the full image instead of sampling the image, so no pixel
  budget applies. $^{\mathrm{c}}$ImageNet ILSVRC2010.
  $^{\mathrm{d}}$Averages: AdaptiveNN stops adaptively, after $3.15$ fixations on average plus its glance.
  $^{\mathrm{e}}$Initialized from our ImageNet model trained with random locations.}
\end{table}

\paragraph{Additional information on individual entries of \cref{tab:prior_work}.} Saccader and DRAM choose their glimpse locations from a
view of the whole image: Saccader's network processes the full image at $224^2$, and DRAM's location network also receives
the two wider, lower-resolution crops of each glimpse, the widest spanning the whole image. Only the glimpses reach the classifier, and only they are counted (note a of
\cref{tab:prior_work}). Saccader itself reports $29.5\%$, the image area its six glimpses cover after removing overlaps;
counted as in the other rows (six $113^2$ glimpses on $331^2$ images), it is $69.9\%$. DRAM was evaluated by the
Saccader authors~\citep{elsayed_SaccaderImprovingAccuracy_2019}, whose appendix describes its ImageNet training in
three stages (two pretraining stages, then full training), counted as for Saccader; its accuracy is taken from the table of
TNet~\citep{papadopoulosHardAttentionScalable2021}. Where a paper gives some variants only in plots (GFNet's other
backbones, Saccader's other glimpse counts), we list its best printed result. For FoveaTer we list its main model; a variant in its appendix that applies six
transformer layers to the full feature map before foveating reaches $78.3\%$. Lukanov's accuracy and budget are from its
own paper, on ImageNet ILSVRC2010. Its result table does not restate the protocol, so we take the one the paper gives
for all its models: two $81^2$ foveated glimpses on $256^2$ images, the first at the center and the second placed by one saccade. The
model keeps no memory, so only the second glimpse is classified, and only it is counted, as for DRAM and Saccader
(note a); counting both would give $20\%$. \citet{liu_ImprovedEfficiencyBased_2023} list the same result at $27\%$. STAM
reports pixels rather than glimpses; its $27$ glimpses are $27.7$K pixels divided by its $32^2$ glimpse size. AME's
retina-like glimpses are counted after their peripheral down-scaling: eight $48^2$ glimpses on $128\times256$ images
deliver $18.75\%$ of the pixels while covering $56.25\%$ of the image area. AME's supplementary states this setting for
segmentation without naming the weights; we assume it for both variants, which share one table and the same retinal
baselines. AdaGlimpse lists the run as full-resolution glimpses with $56.25\%$; we follow AME's own materials.
AdaptiveNN uses no fixed number of fixations but concludes observation adaptively once the evidence suffices, so
its glimpse count and budget are averages, computed from the $3.15$ ``\#Visual Fixations (average)'' its paper reports
at this operating point. Its paper states ImageNet-pretrained initialization only for its other tasks; for ImageNet it
follows the full-length DeiT and ResNet training schedules, and its released code builds the model without pretrained
weights, so we list it as trained from scratch. CanViT's ADE20K result uses a coarse-to-fine viewing policy chosen at
inference, a $64^2$ canvas instead of the trained $32^2$ and a frozen linear probe, at $21$ glimpses; its ImageNet result is fine-tuned and peaks at $16$ glimpses.

\paragraph{Our models.} Our rows use the learned policy at the trained budget of four glimpses, averaged over three
models trained with different random seeds. The budget counts the $4 \cdot 64^2$ samples of the four glimpses against
the $224^2$ (ImageNet) or $512^2$ (ADE20K) pixels of the image. The ADE20K mIoU of the table is scored at each
image's native resolution, the standard protocol; elsewhere segmentation numbers are
scored at $512 \times 512$ (\Cref{app:training_details}), which is why the numbers in \cref{tab:prior_work} are not the same as the ones reported in \cref{tab:main_combined_cur}.

\paragraph{Comparability.} The numbers in \cref{tab:prior_work} are not comparable. The models differ at once in backbone, pretraining, input
resolution, number and size of glimpses and evaluation protocol. Nor does a small pixel budget imply low
computational cost: How much compute a model uses depends as much on the architecture that processes the pixels as on how many it
samples. A fair comparison would need controlled benchmarks at
matched input resolutions, pixel budgets and number of glimpses with clearly defined performance metrics. Beyond task
performance and efficiency, comparisons of active vision models that aim to be biologically faithful may also benefit from the axes
of alignment discussed in \cref{sec:conclusion}.

\clearpage
\section{Unrolling beyond the training horizon}
\label{app:glimpse_sweep}
\begin{figure}[t]
  \centering
  \includegraphics[width=\textwidth]{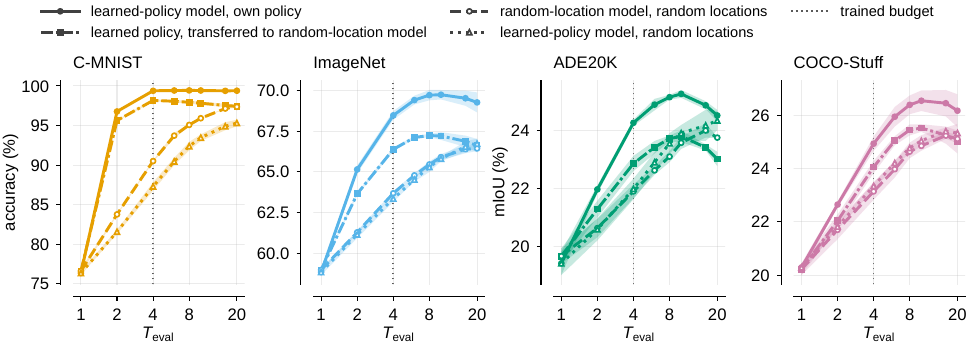}
  \caption{\textbf{Unrolling beyond the training horizon.} Every model is trained at $T=4$ (dotted rule) and unrolled to each budget at
  inference; nothing is retrained. One color per dataset, one line style per setting.
  \textbf{Solid} is each dataset's own learned policy, \textbf{dashed} a model trained and
  evaluated with random locations, \textbf{dash-dotted} each seed's learned policy transferred to
  that same seed's random-location model, and \textbf{dotted} the learned-policy model evaluated with
  random locations. All four are measured by one protocol (last checkpoints; learned policies fixate the argmax of their probability map,
  random locations are drawn uniformly once per evaluation, so these points can differ slightly from the corresponding table cells in \cref{tab:main_combined_cur,tab:policy_transfer_fused}). At $T_\text{eval}{=}1$ the four curves collapse to two, since the
  first fixation is the fixed image center and no policy has acted yet. Shading is $\pm1$ std over three
  models trained with different random seeds, measured on this protocol in every panel.}
  \label{fig:horizon_replay}
\end{figure}

This appendix evaluates how models trained with four glimpses perform when they are unrolled for more glimpses at inference.
The number of glimpses is fixed during training, but nothing in the architecture fixes it at
inference. Because the memory is a spatial feature map rather than a vector, extra glimpses keep
writing into the locations the policy selects, so the model can simply be unrolled for longer. \Cref{fig:horizon_replay} evaluates
this: every model is trained with $T=4$ glimpses per episode and then unrolled to each budget $T_\text{eval}$ with no
retraining. Fixations were chosen via argmax of the probability map of the policy.

The figure shows each learned-policy model and each random-location model with its native policy, and
two policy transfers (\Cref{sec:results_transfer}): each seed's learned policy transferred to that same seed's
random-location model, and the learned-policy model evaluated with random locations.

With their own learned policy, the models on ImageNet, ADE20K and COCO-Stuff keep improving past the budget they were
trained at, peak at $T_\text{eval} = 8$--$10$ and then decay slightly, staying above their value at the trained budget up
to $T_\text{eval} = 20$; C-MNIST is saturated from the trained budget on.

With more glimpses, random locations catch up: the margin of the learned policy over random locations narrows on every
dataset. On the learned-policy model it stays positive over the whole range; transferred to the random-location model, it
falls to about zero or below at the largest budgets on ADE20K and COCO-Stuff.

\clearpage
\section{Visualizations}
\label{app:visualizations}
This appendix shows single episodes: what the model receives at each step (\Cref{app:transform_panels}) and how its policy, memory and prediction develop (\Cref{app:devpanels}).

\paragraph{Selection of the examples.} The examples are chosen, not drawn at random, to show how the task
prediction can change across glimpses. For one learned-policy model per dataset (the one shown in \Cref{fig:intro}), we
ran one episode on each image of a validation pool (C-MNIST: all \num[group-digits=integer]{10000}, ImageNet: a random \num[group-digits=integer]{5000}, ADE20K:
all \num[group-digits=integer]{2000}, COCO-Stuff: a random \num[group-digits=integer]{3000}). Each of several criteria keeps the episodes that meet a condition
and ranks them by a score, and the twenty highest-ranked episodes of each criterion became candidates (up to $120$ per
classification dataset and $100$ per segmentation dataset). For classification:
\begin{enumerate}[nosep]
  \item wrong after the first glimpse and correct at the end, ranked by the rise in the ground-truth probability from
  the first to the last glimpse;
  \item the predicted class changes at the last glimpse, ranked by how much the ground-truth and final-class
  probabilities change at that glimpse;
  \item the final class differs from the first prediction and gains at least $40\%$ of probability in
  one step, ranked by that gain;
  \item at least three different classes are predicted over the episode, ranked by their number;
  \item correct at some glimpse but wrong at the end, ranked by the drop in the ground-truth probability from its
  highest value before the last glimpse;
  \item never correct, with a final class that differs from the first, ranked by the rise in the final-class probability.
\end{enumerate}
For segmentation:
\begin{enumerate}[nosep]
  \item all episodes, ranked by the gain in per-image mIoU (plus half the gain in pixel accuracy) from the first to
  the last glimpse;
  \item final pixel accuracy higher than after the second glimpse, ranked by the largest share of pixels relabeled at
  a later glimpse;
  \item pixel accuracy reaches at least $30\%$ at some glimpse, ranked by the share of pixels relabeled over the episode;
  \item images with at least $70\%$ labeled pixels and at least three ground-truth classes, ranked by the final
  pixel accuracy;
  \item all episodes, ranked by the drop in pixel accuracy from its highest value before the last glimpse.
\end{enumerate}
From the candidates, we picked four examples per dataset by hand, so not every criterion is represented. One of the
four examples was always selected to be a failure. In the order of \Cref{app:devpanels}, the examples come from
criteria 3, 2, 4 and 5 for ImageNet, 2, 2, 4 and 5 for C-MNIST, 4, 1, 1 and 5 for ADE20K and 3, 1, 4 and 5 for
COCO-Stuff. The panels therefore show unusual episodes rather than typical ones. On average, the task prediction improves
with every glimpse.

\subsection{Visualization of the foveation}
\label{app:transform_panels}
\begin{figure}[p]
    \centering
    \includegraphics[width=0.82\textwidth]{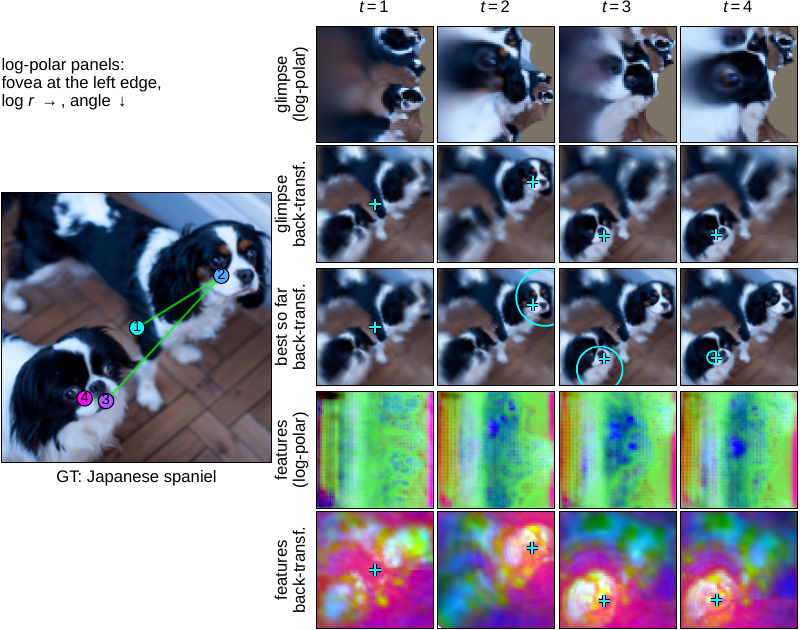}\par\vspace{0.2em}
    \includegraphics[width=0.82\textwidth]{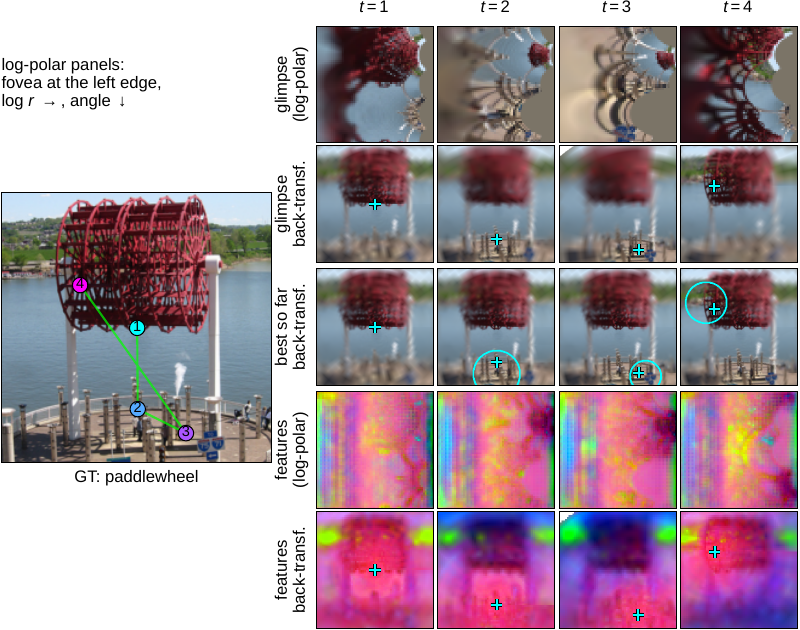}
    \caption{\textbf{Foveation, ImageNet.} Same examples and scanpaths as \Cref{fig:devpanel_imagenet}
    (top: \emph{Japanese spaniel}, bottom: \emph{paddlewheel}). Rows as described in
    \Cref{app:transform_panels}.}
    \label{fig:transform_imagenet}
\end{figure}

\begin{figure}[p]
    \centering
    \includegraphics[width=0.82\textwidth]{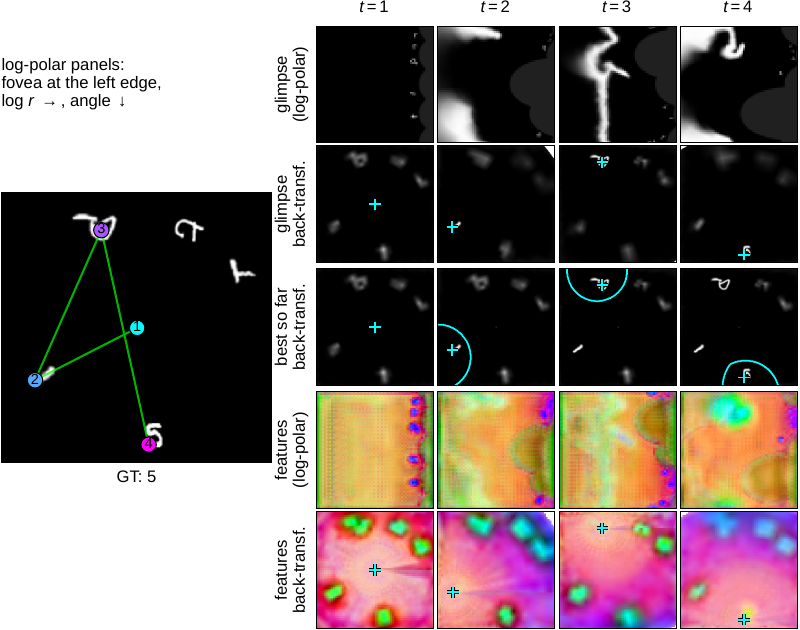}\par\vspace{0.2em}
    \includegraphics[width=0.82\textwidth]{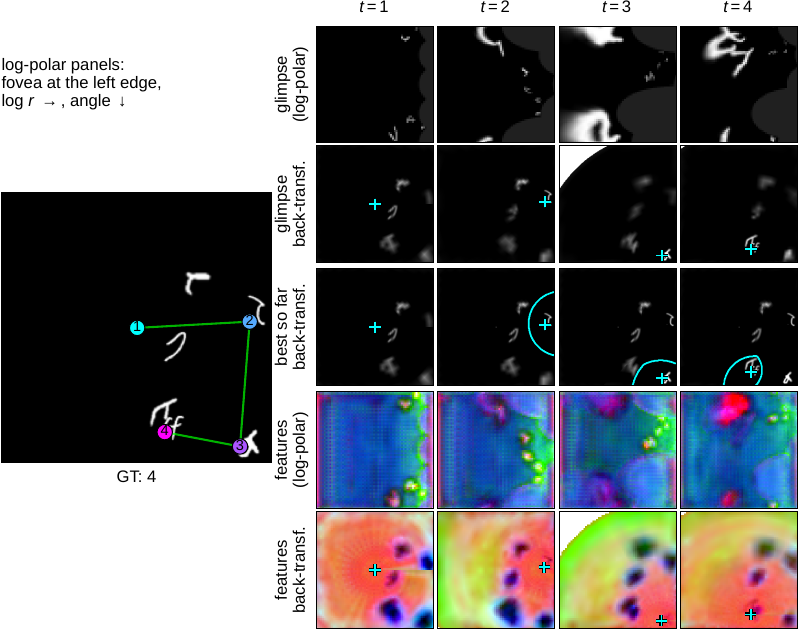}
    \caption{\textbf{Foveation, C-MNIST.} Same examples and scanpaths as
    \Cref{fig:devpanel_mnist} (ground truth \emph{5} and \emph{4}). Rows as described in
    \Cref{app:transform_panels}.}
    \label{fig:transform_mnist}
\end{figure}

\begin{figure}[p]
    \centering
    \includegraphics[width=0.82\textwidth]{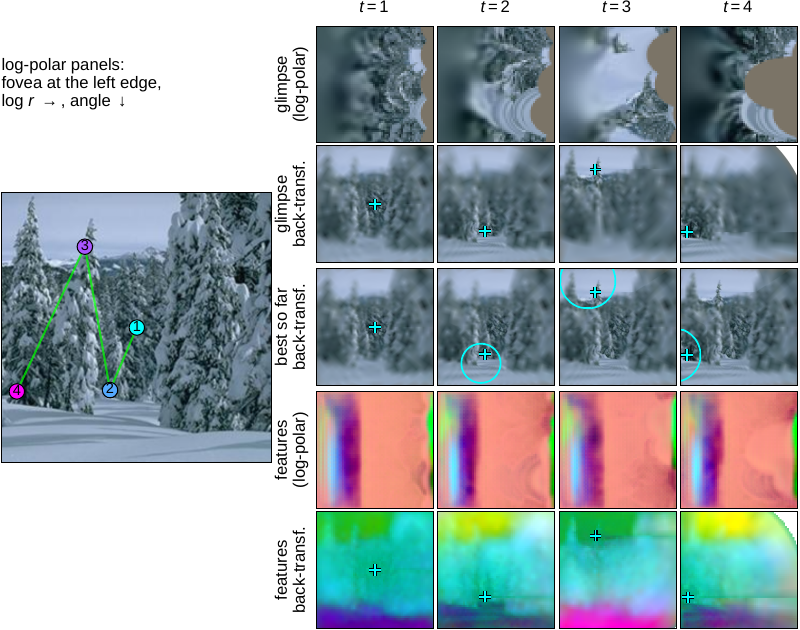}\par\vspace{0.2em}
    \includegraphics[width=0.82\textwidth]{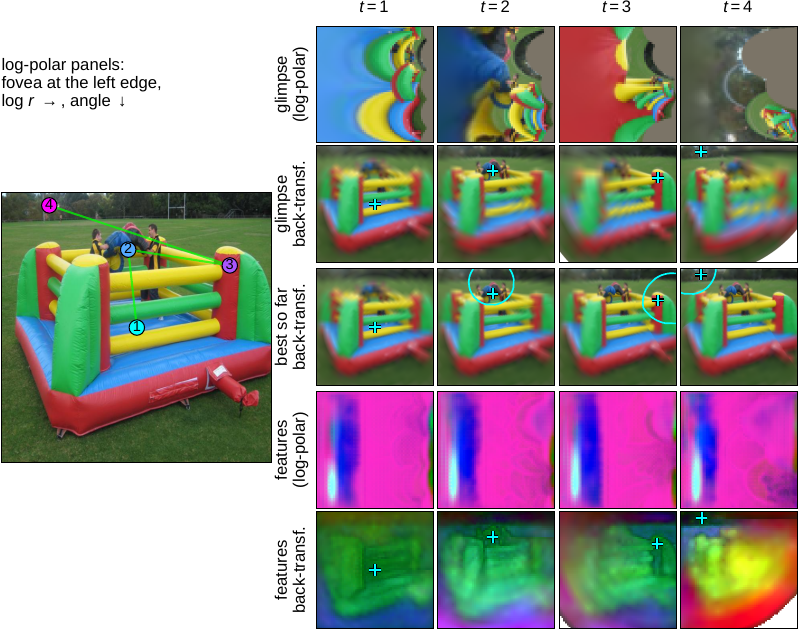}
    \caption{\textbf{Foveation, ADE20K.} Same examples and scanpaths as
    \Cref{fig:devpanel_ade20k}. Rows as described in \Cref{app:transform_panels}.}
    \label{fig:transform_ade20k}
\end{figure}

\begin{figure}[p]
    \centering
    \includegraphics[width=0.82\textwidth]{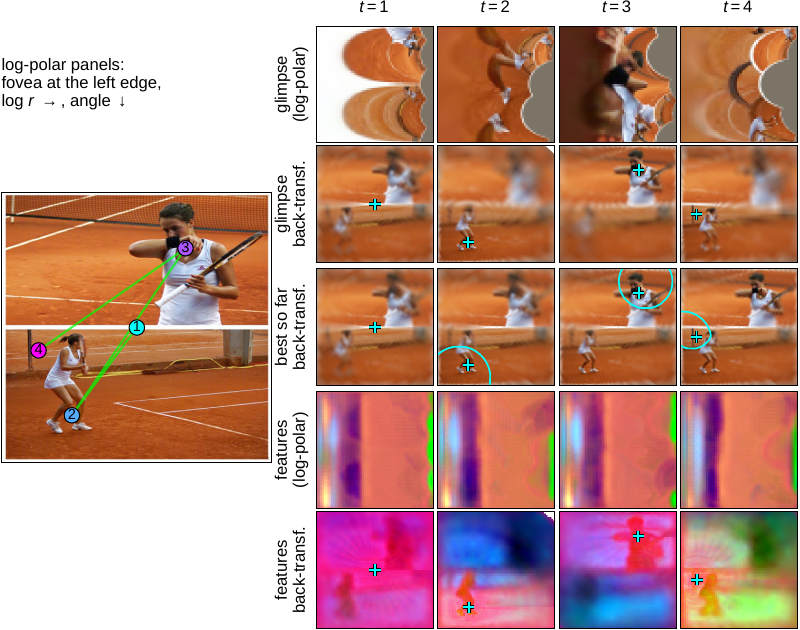}\par\vspace{0.2em}
    \includegraphics[width=0.82\textwidth]{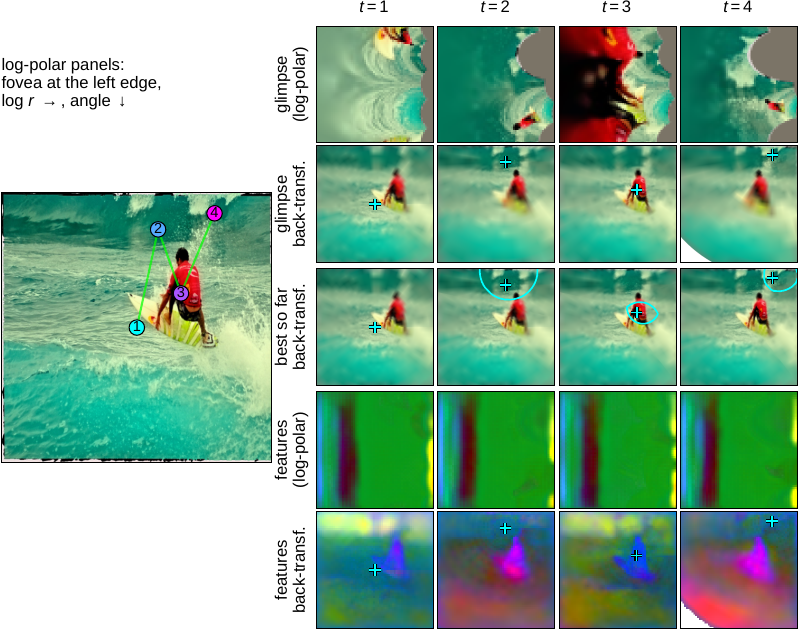}
    \caption{\textbf{Foveation, COCO-Stuff.} Same examples and scanpaths as
    \Cref{fig:devpanel_coco}. Rows as described in \Cref{app:transform_panels}.}
    \label{fig:transform_coco}
\end{figure}

\Cref{fig:transform_imagenet,fig:transform_mnist,fig:transform_ade20k,fig:transform_coco} show what the model receives at each step for the first two examples of each dataset in
\Cref{app:devpanels} (same scanpaths, one column per glimpse). The rows are
(1)~the glimpse in log-polar coordinates, i.e.\ the backbone input (log-radius from left to right, so the fovea is the
left edge; gray: outside the image);
(2)~the glimpse mapped back to image coordinates, for illustration only;
(3)~the best view so far: each pixel from the glimpse whose fixation is closest to it, which sampled it most finely,
as the sample spacing grows with the distance from the fixation (\Cref{app:logpolar_density}); the outline marks where
the current glimpse samples at least twice as finely as all earlier ones, a disk around the new fixation pushed away
from the earlier ones;
(4)~the backbone features in log-polar coordinates;
(5)~the same features mapped into image coordinates, the input of the memory (\Cref{sec:memory}).
Rows (4) and (5) show the first three principal components of the features, one basis per row and example. The cyan
cross marks the fixation.

\clearpage
\subsection{Visualization of policy, memory and prediction across glimpses}
\label{app:devpanels}

Each figure follows one episode: a validation example unrolled along a scanpath drawn from its policy, one column
per step. The rows show the policy heatmap (as logits, since the probabilities are too peaked to show more than their
maximum; the cyan cross marks the fixation), the memory features projected onto their first three principal
components (one basis per example), the change of the memory $|\Delta \mathbf{H}|$, and the task prediction after that glimpse.
Unlike the other rows, the heatmap in column $t$ is the probability map that chose $f_t$, computed at step $t-1$; column $t = 1$ has none, since $f_1$ is the image center.

\paragraph{Memory updates.}
\label{app:gating_visual}
The $|\Delta \mathbf{H}|$ row is $|\mathbf{H}_t - \mathbf{H}_{t-1}|$ averaged over the $128$ channels, on one color scale per
example. At $t = 1$ the change covers the whole image, since $\mathbf{H}_0 = \mathbf{0}$. In the later columns of the
ImageNet and C-MNIST examples, the largest changes lie near the fixation of that glimpse: a disk around it on ImageNet
(\Cref{fig:devpanel_imagenet}, top) and the characters close to it on C-MNIST (\Cref{fig:devpanel_mnist}, top). On the
segmentation examples, the later changes are faint compared with the first one on the shared color scale
(\Cref{fig:devpanel_ade20k,fig:devpanel_coco}).

\paragraph{Classification.}
The task prediction is shown as the top-5 class probabilities, with the ground-truth class in green. Percentages
refer to the four predictions in order.

\paragraph{Semantic segmentation.}
The task prediction is shown as the predicted map with its pixel accuracy, followed by a row that marks which pixels
changed label relative to the previous step: green where a pixel became correct, orange where it
became wrong, gray where it changed between two wrong labels. Percentages are rounded and refer to
the four predictions in order.

\begin{figure}[p]
    \centering
    \includegraphics[width=\textwidth]{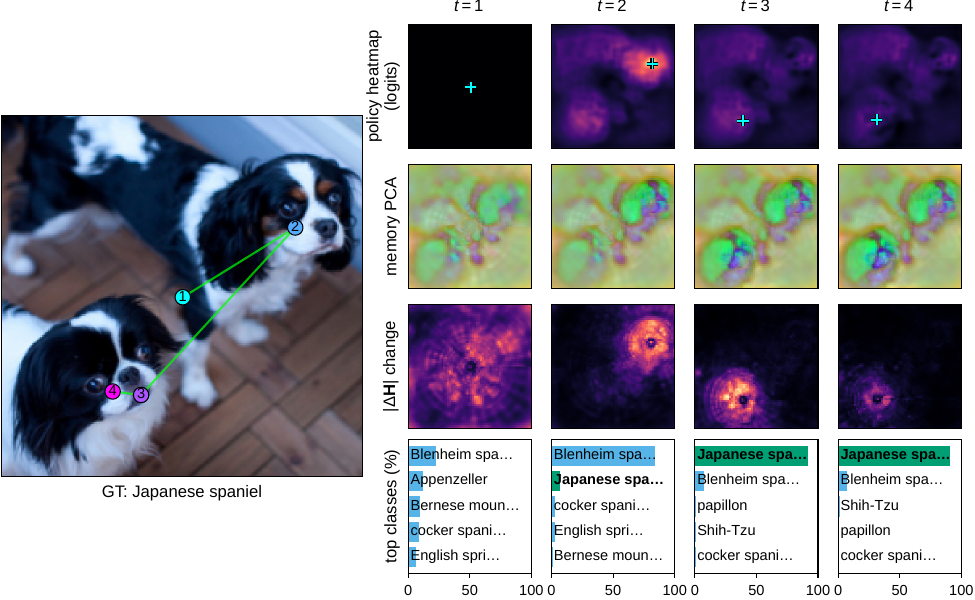}\par\vspace{0.2em}
    \includegraphics[width=\textwidth]{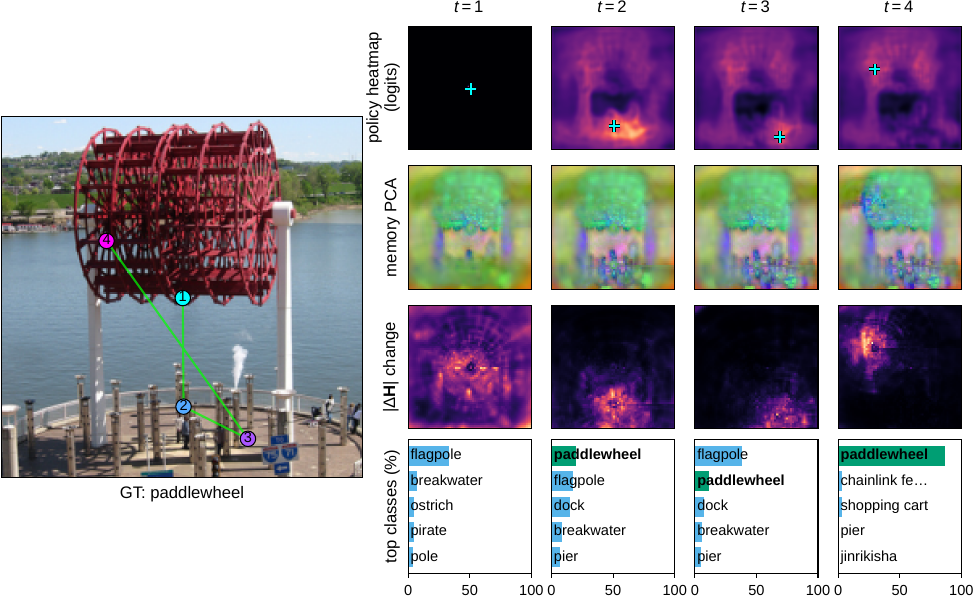}
    \caption{\textbf{Policy, memory and prediction, ImageNet ($T=4$).} \textbf{Top}: ground truth \emph{Japanese spaniel}. The first two
    predictions name the related \emph{Blenheim spaniel} ($23$, $84\%$); after the third glimpse
    the prediction switches to \emph{Japanese spaniel} ($92\%$) and keeps it ($91\%$).
    \textbf{Bottom}: the predicted class alternates between \emph{flagpole} ($33\%$),
    \emph{paddlewheel} ($20\%$) and \emph{flagpole} again ($39\%$); the last glimpse raises the
    ground truth \emph{paddlewheel} to $86\%$.
}
    \label{fig:devpanel_imagenet}
\end{figure}
\begin{figure}[p]\ContinuedFloat
    \centering
    \includegraphics[width=\textwidth]{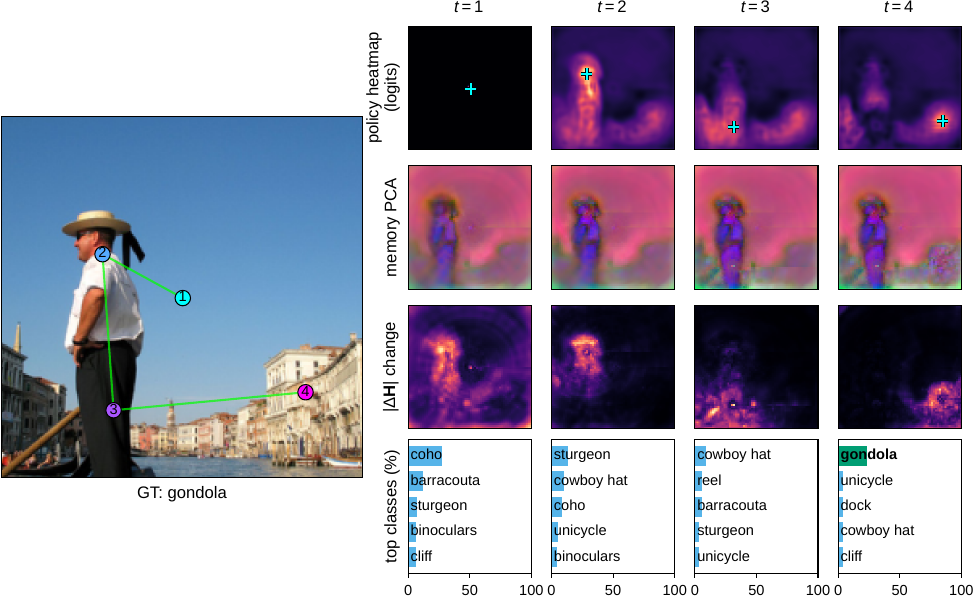}\par\vspace{0.2em}
    \includegraphics[width=\textwidth]{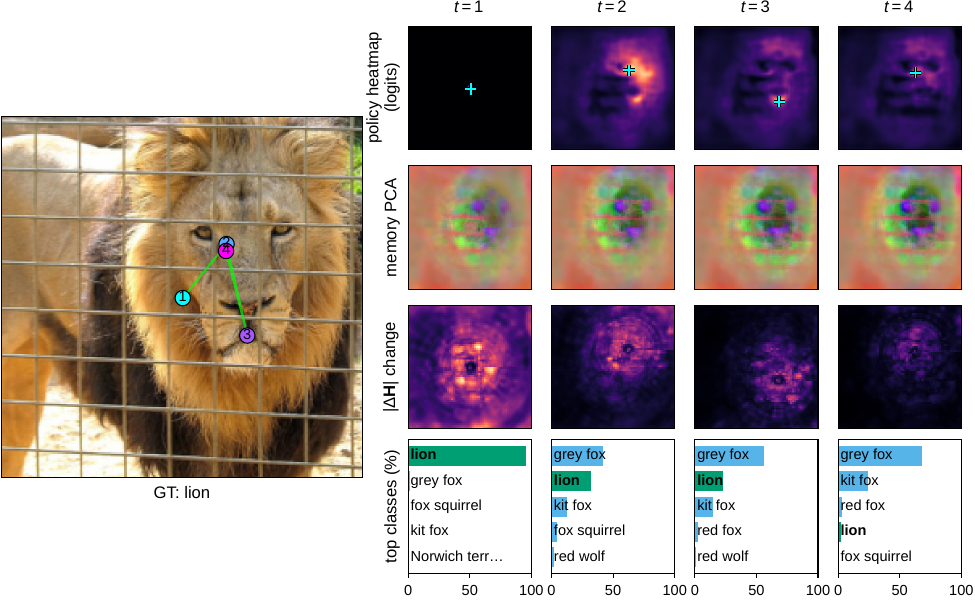}
    \caption{\textbf{Policy, memory and prediction, ImageNet (continued).} \textbf{Top}: four predictions, four classes, all at low
    confidence: \emph{coho} ($28\%$), \emph{sturgeon} ($13\%$), \emph{cowboy hat} ($9\%$), and
    the ground truth \emph{gondola} ($23\%$) only at the last glimpse.
    \textbf{Bottom, failure}: \emph{lion} at $95\%$ after the first glimpse loses probability at
    every step ($32$, $23$, $2\%$) and \emph{grey fox} is the predicted class from the second glimpse on
    ($42$, $56$, $68\%$).
}
\end{figure}

\begin{figure}[p]
    \centering
    \includegraphics[width=\textwidth]{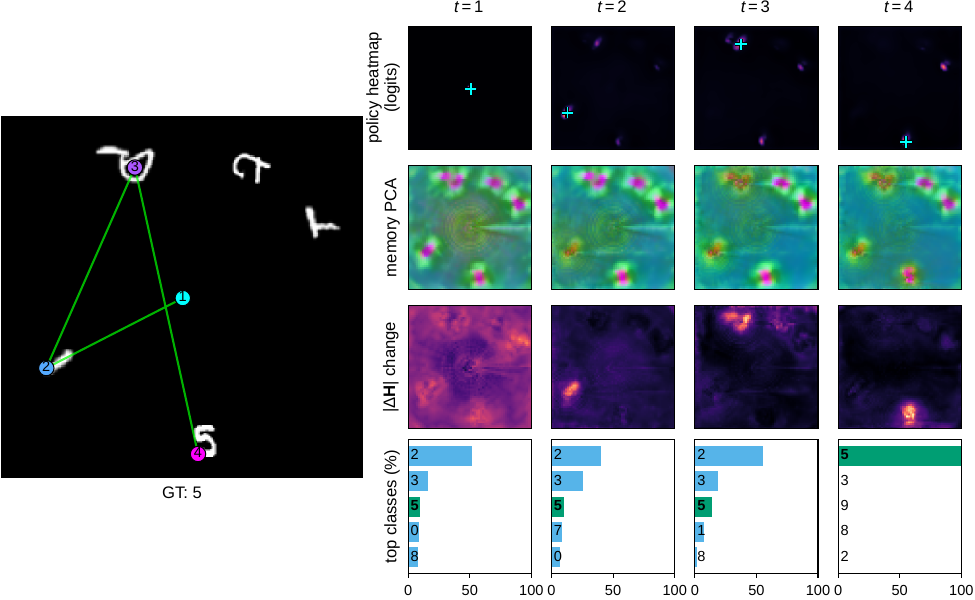}\par\vspace{0.2em}
    \includegraphics[width=\textwidth]{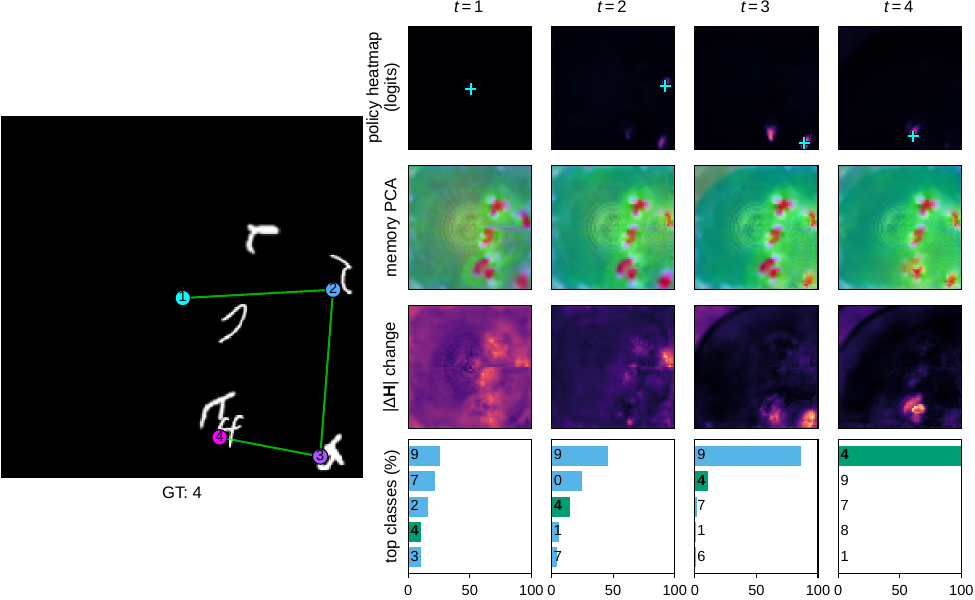}
    \caption{\textbf{Policy, memory and prediction, C-MNIST ($T=4$).} \textbf{Top}: \emph{2} for the first three predictions
    ($52$, $40$, $55\%$) while the ground truth \emph{5} stays at $9$--$14\%$; the last glimpse
    switches to \emph{5} at $100\%$.
    \textbf{Bottom}: \emph{9} with rising confidence ($26$, $46$, $87\%$) while the ground
    truth \emph{4} stays at $10$--$15\%$; the last glimpse switches to \emph{4} at $100\%$.
}
    \label{fig:devpanel_mnist}
\end{figure}
\begin{figure}[p]\ContinuedFloat
    \centering
    \includegraphics[width=\textwidth]{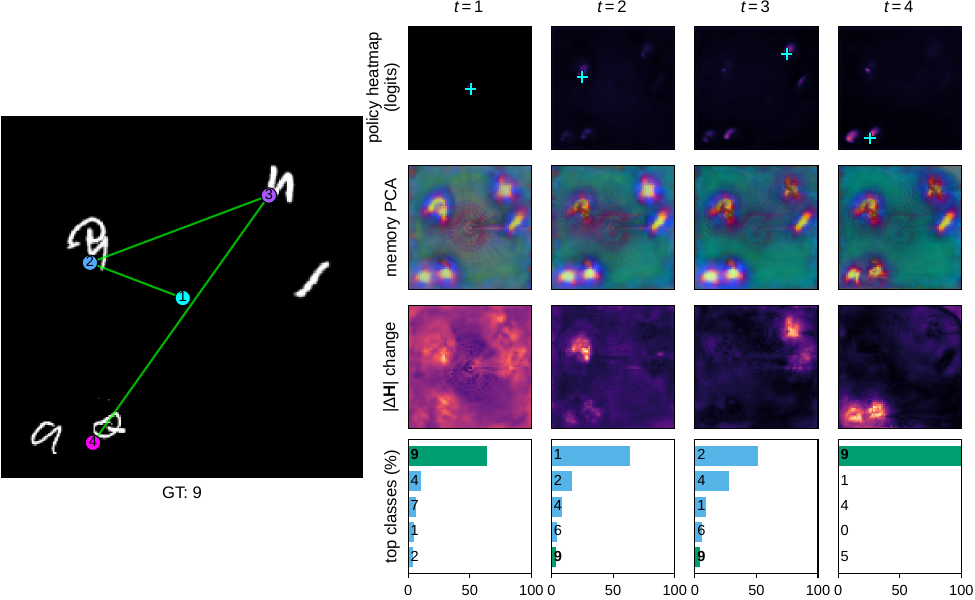}\par\vspace{0.2em}
    \includegraphics[width=\textwidth]{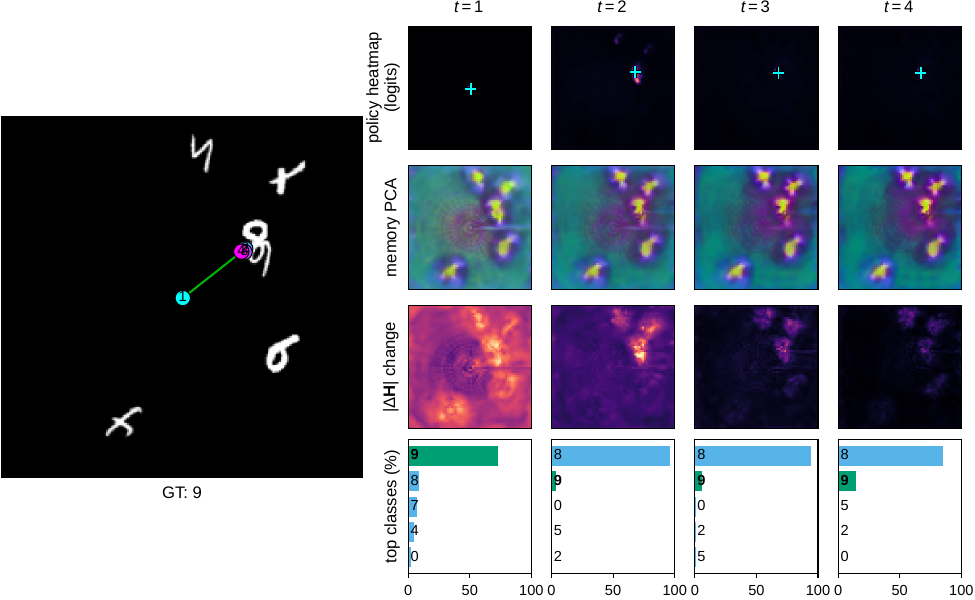}
    \caption{\textbf{Policy, memory and prediction, C-MNIST (continued).} \textbf{Top}: \emph{9} ($64\%$), \emph{1} ($63\%$),
    \emph{2} ($51\%$), then the ground truth \emph{9} again at $100\%$.
    \textbf{Bottom, failure}: \emph{9} at $73\%$ after the first glimpse; the three later
    fixations land on the same location and the prediction moves to \emph{8} ($96$, $94$, $85\%$),
    with the ground truth at $4$--$15\%$.
}
\end{figure}

\begin{figure}[p]
    \centering
    \includegraphics[width=0.82\textwidth]{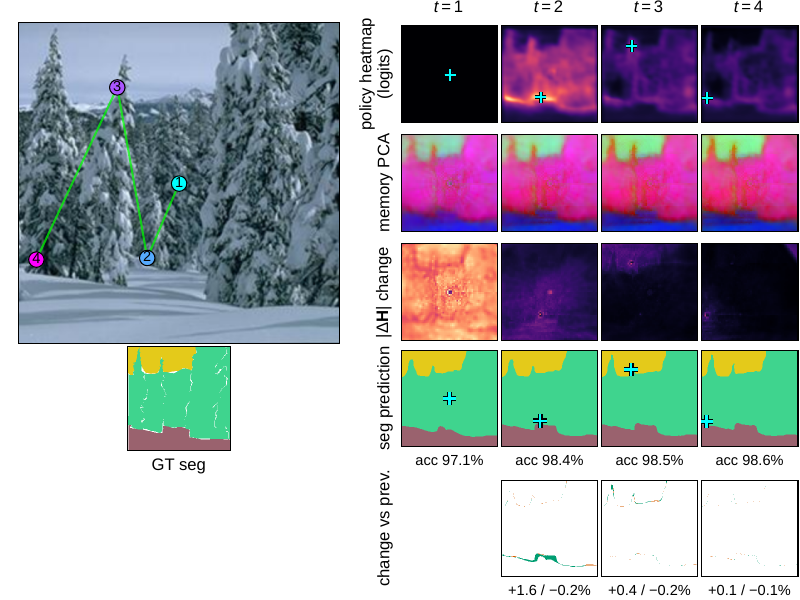}\par\vspace{0.2em}
    \includegraphics[width=0.82\textwidth]{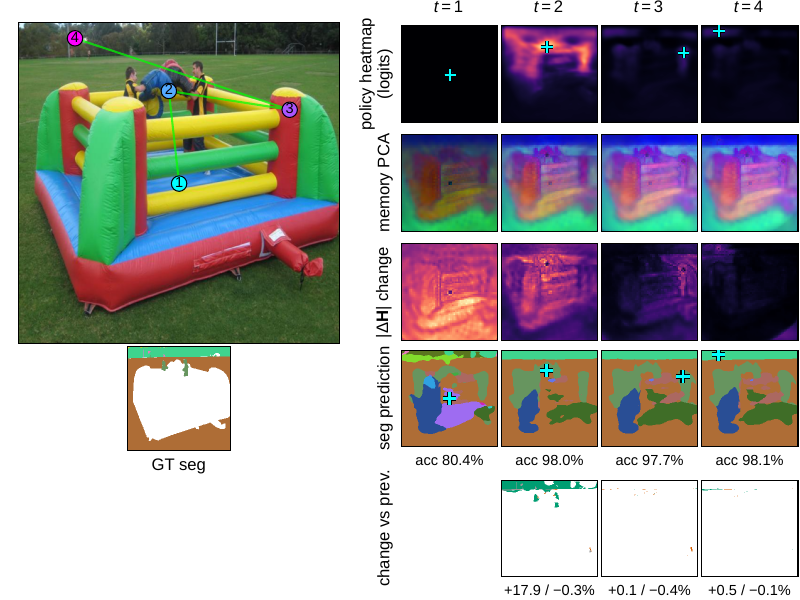}
    \caption{\textbf{Policy, memory and prediction, ADE20K ($T=4$).} \textbf{Top}: pixel accuracy $97$--$99\%$ at every step and
    per-image mIoU from $93\%$ to $96\%$; later glimpses relabel at most $2\%$ of the pixels.
    \textbf{Bottom}: pixel accuracy $80\%$ after the first glimpse and $98\%$ after the second,
    which relabels $30\%$ of the pixels; per-image mIoU $12\to40\%$.
}
    \label{fig:devpanel_ade20k}
\end{figure}
\begin{figure}[p]\ContinuedFloat
    \centering
    \includegraphics[width=0.82\textwidth]{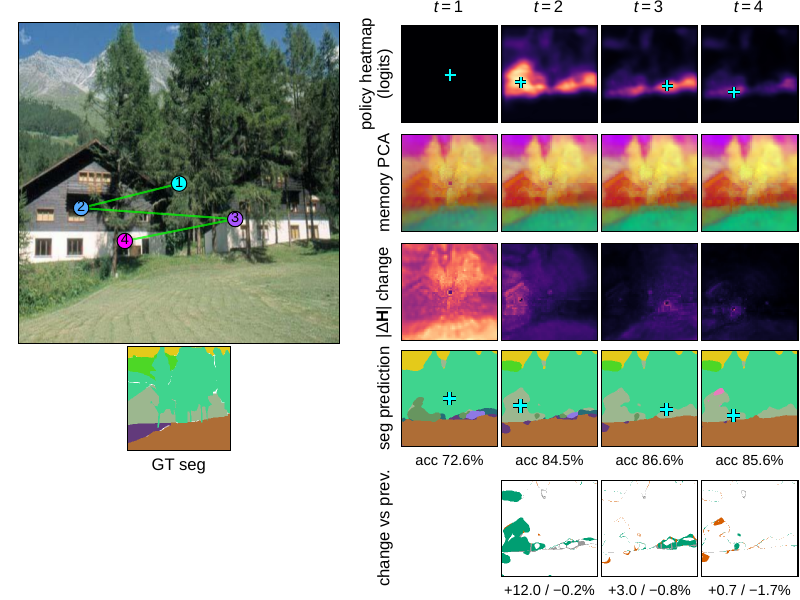}\par\vspace{0.2em}
    \includegraphics[width=0.82\textwidth]{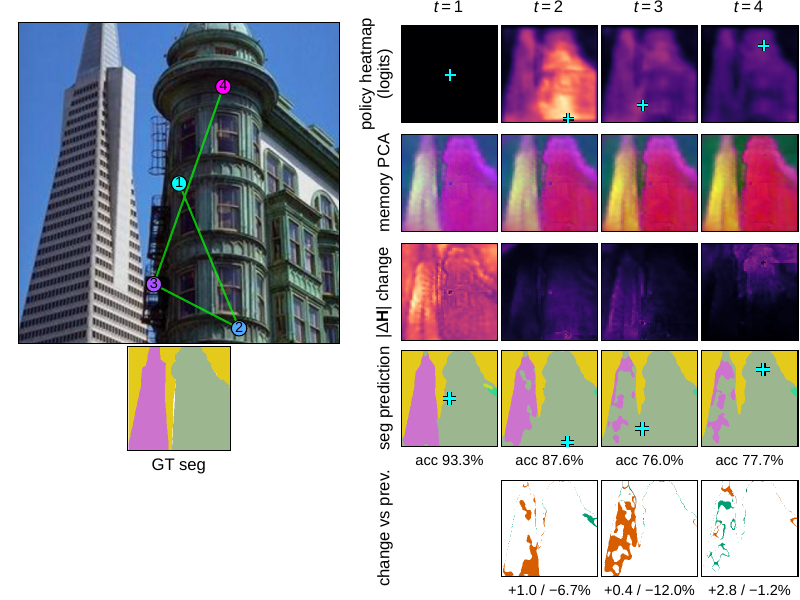}
    \caption{\textbf{Policy, memory and prediction, ADE20K (continued).} \textbf{Top}: pixel accuracy $73$, $85$, $87$, $86\%$ and
    per-image mIoU $22\to49\%$, with less relabeling at each step ($14$, $5$, $3\%$ of the
    pixels).
    \textbf{Bottom, failure}: $93\%$ pixel accuracy after the first glimpse; the next two
    glimpses relabel $8\%$ and $13\%$ of the pixels, mostly from correct to wrong, and accuracy
    falls to $88\%$ and $76\%$ ($78\%$ at the end).
}
\end{figure}

\begin{figure}[p]
    \centering
    \includegraphics[width=0.82\textwidth]{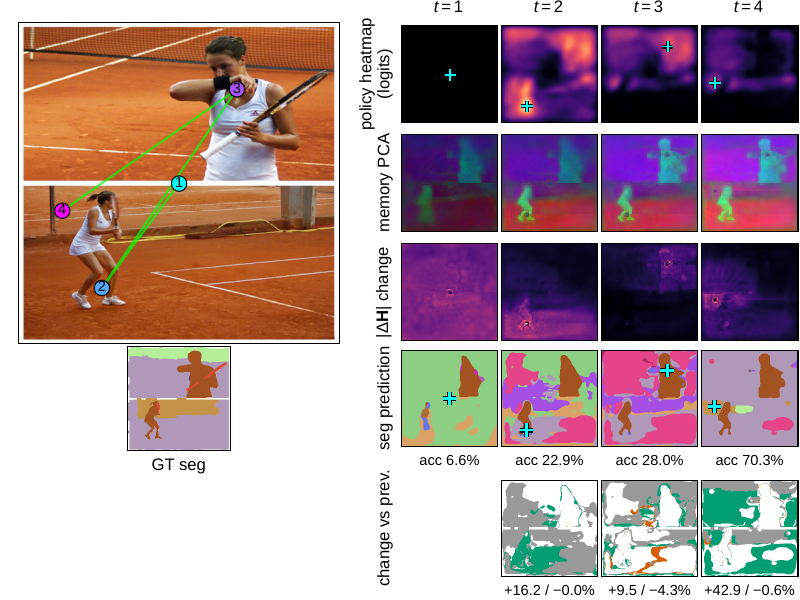}\par\vspace{0.2em}
    \includegraphics[width=0.82\textwidth]{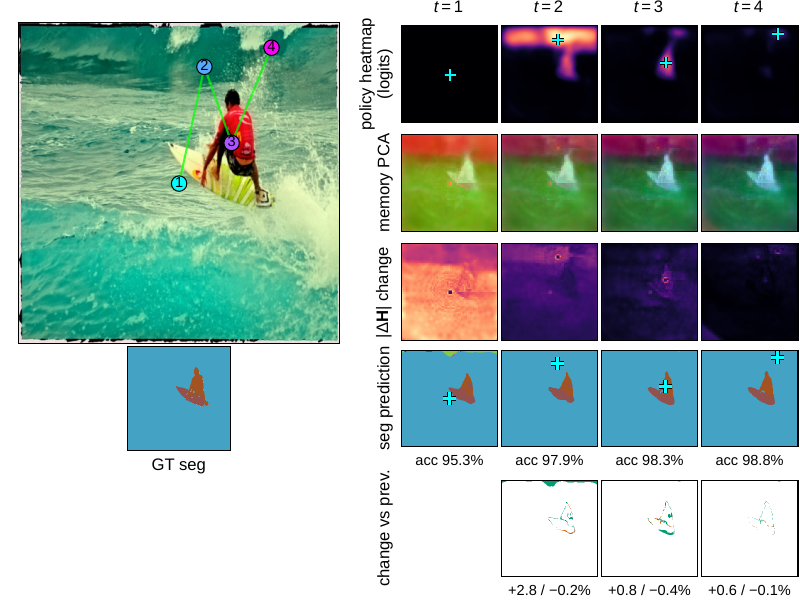}
    \caption{\textbf{Policy, memory and prediction, COCO-Stuff ($T=4$).} \textbf{Top}: $60$, $49$ and $63\%$ of the pixels change label
    at the three later glimpses; pixel accuracy $7$, $23$, $28$, $70\%$.
    \textbf{Bottom}: pixel accuracy changes little ($95$, $98$, $98$, $99\%$) while per-image
    mIoU rises from $33\%$ to $82\%$.
}
    \label{fig:devpanel_coco}
\end{figure}
\begin{figure}[p]\ContinuedFloat
    \centering
    \includegraphics[width=0.82\textwidth]{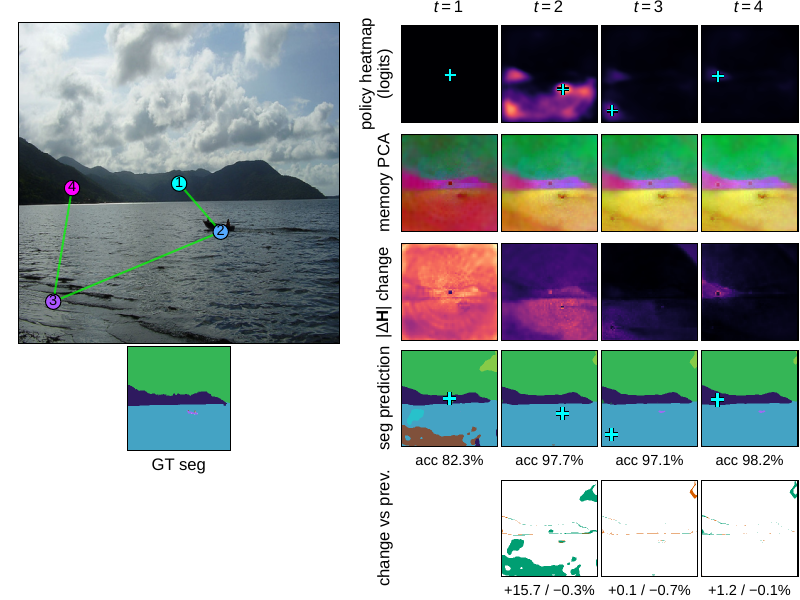}\par\vspace{0.2em}
    \includegraphics[width=0.82\textwidth]{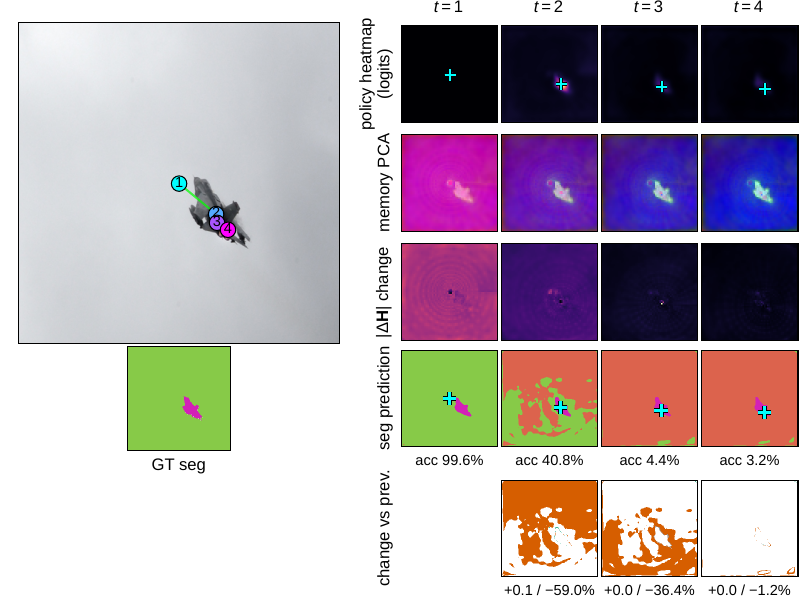}
    \caption{\textbf{Policy, memory and prediction, COCO-Stuff (continued).} \textbf{Top}: pixel accuracy $82\%$ after the first glimpse
    and $98\%$ after the second, which relabels $16\%$ of the pixels ($98\%$ at the end);
    per-image mIoU $27\to55\%$.
    \textbf{Bottom, failure}: pixel accuracy above $99\%$ after the first glimpse, then $41$,
    $4$ and $3\%$; the second and third glimpses relabel $59\%$ and $37\%$ of the pixels.
}
\end{figure}

\clearpage
\section{Heuristic fixation policies}
\label{app:heuristics}
\begin{figure}[t]
  \centering
  \includegraphics[width=\textwidth]{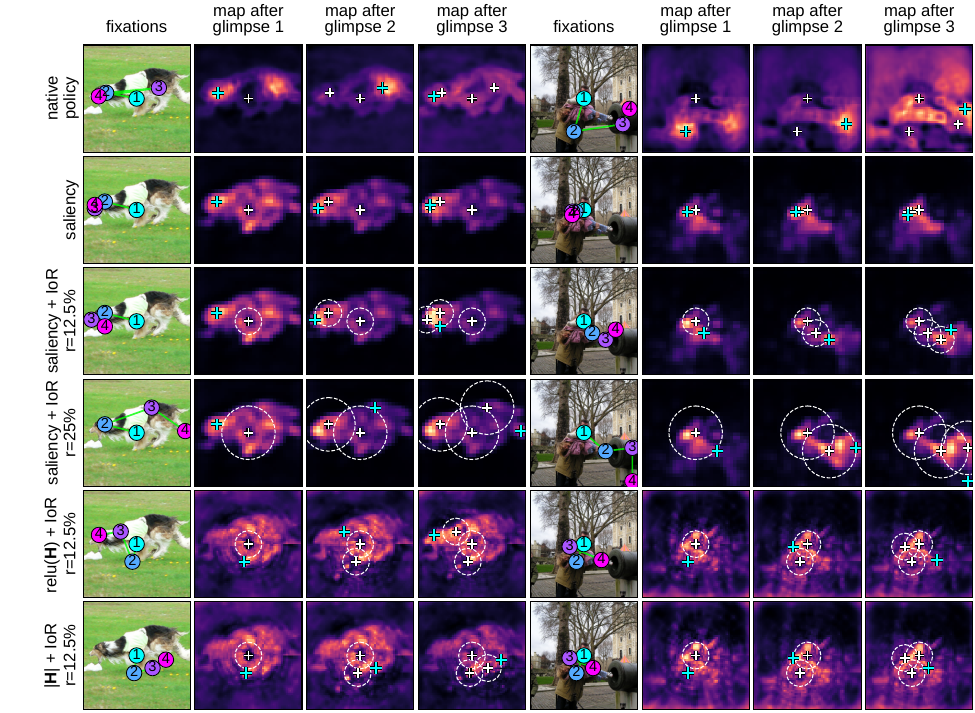}
  \caption{\textbf{Heuristic policies on ImageNet.} The learned-policy model evaluated with
    different policies on two validation images (left and right). Each row is one evaluation policy: the model's own
    (native) policy, saliency, saliency with IoR at $r = 12.5\%$ and at $r = 25\%$, and the two memory-saliency variants
    $\mathrm{relu}(\mathbf{H})$ and $|\mathbf{H}|$, both with IoR at $r = 12.5\%$. For each image, the first column shows
    the four fixations of the policy, numbered in order (1 is the initial center fixation). The next three columns show
    the map from which the next fixation was chosen after glimpses 1, 2 and 3, with brighter colors for higher values.
    In these maps, the cyan cross marks the chosen fixation (the argmax of the map, or in the native-policy row the
    fixation drawn from the probability map), white crosses mark earlier fixations, and dashed circles the regions
    excluded by IoR. The native-policy row shows the policy logits.}
  \label{fig:heuristic_panels_imagenet}
\end{figure}

\begin{figure}[t]
  \centering
  \includegraphics[width=\textwidth]{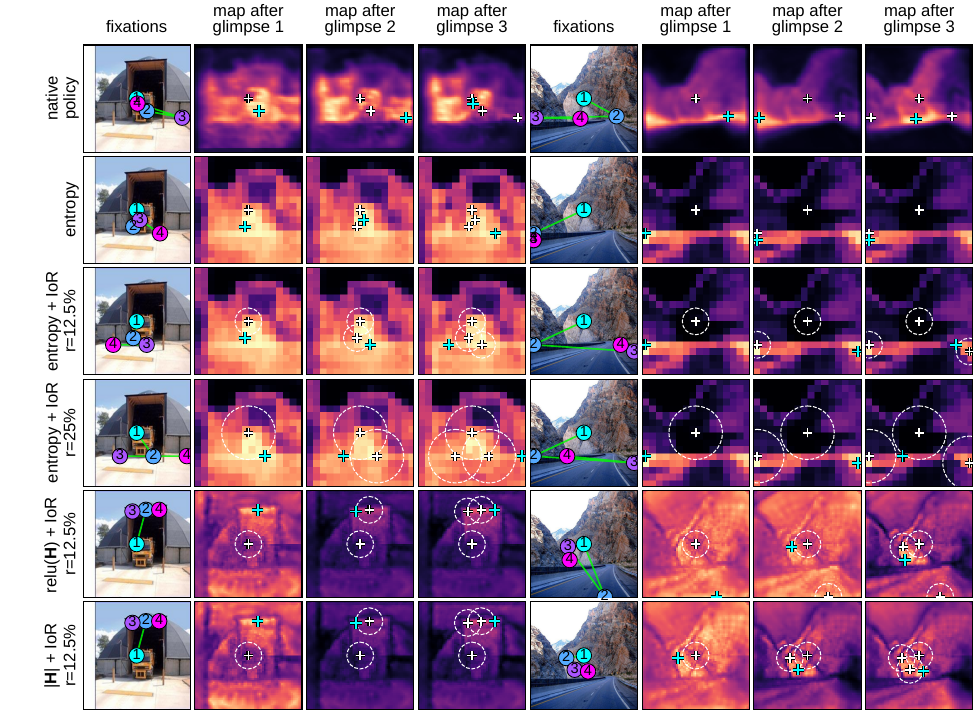}
  \caption{\textbf{Heuristic policies on ADE20K.} As in \cref{fig:heuristic_panels_imagenet}, with entropy in
    place of saliency in rows 2 to 4. The entropy map is shown on its $16{\times}16$ block grid.}
  \label{fig:heuristic_panels_ade20k}
\end{figure}

\begin{figure}[t]
  \centering
  \includegraphics[width=\textwidth]{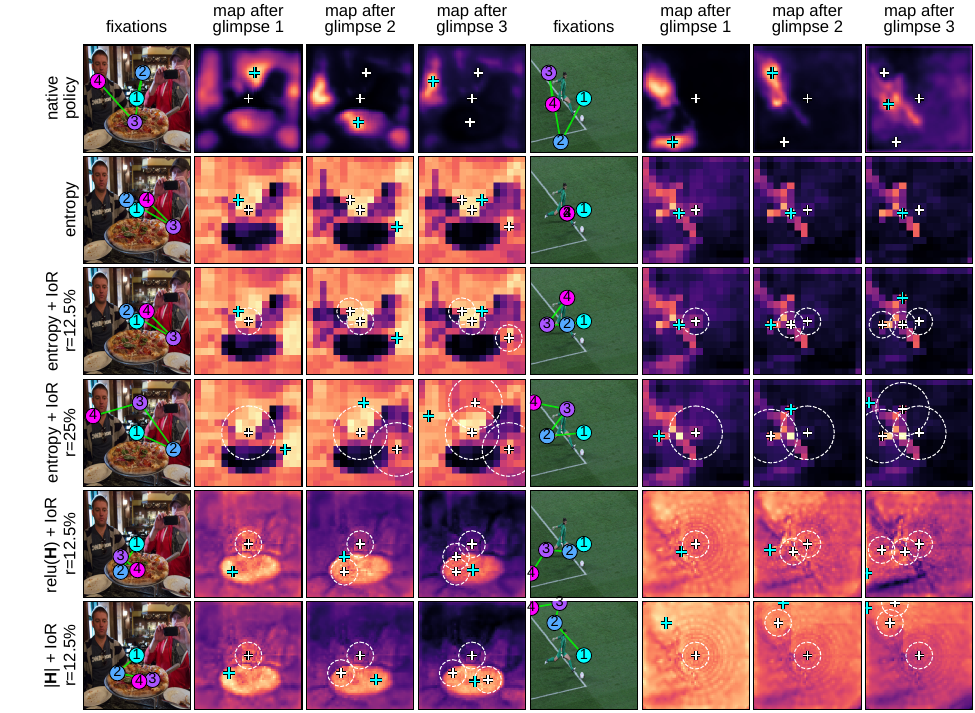}
  \caption{\textbf{Heuristic policies on COCO-Stuff}, as in \cref{fig:heuristic_panels_ade20k}.
}
  \label{fig:heuristic_panels_coco}
\end{figure}

Policy transfer (\Cref{sec:results_transfer}) compares learned policies with hand-designed ones; this appendix defines the heuristics.
\Cref{tab:heuristic_policies} gives the variants that the main text does not report, after four glimpses, and
\Cref{fig:heuristic_panels_imagenet,fig:heuristic_panels_ade20k,fig:heuristic_panels_coco} show, for two validation images
per dataset, the fixations of each heuristic and the map each choice was made from.

\begin{table}[t]
  \centering
  \caption{\textbf{Further heuristic policies.} Accuracy (ImageNet) and mIoU (ADE20K, COCO-Stuff) after $T{=}4$ glimpses for the heuristic variants that \Cref{tab:policy_transfer_fused} does not show. The IoR radius $r$ is given as a fraction of the image width (\Cref{app:heuristics}). The $r{=}25\%$ rows are the ones reported in \Cref{tab:policy_transfer_fused}, repeated here to compare the two IoR radii. Saliency is run on ImageNet only and entropy on the segmentation datasets only (\Cref{app:heuristics}). \emph{Learned}: learned-policy models, \emph{random}: random-location models, each evaluated with the heuristic of the row. Mean $\pm$ std over three models trained with different random seeds, last checkpoints.}
  \label{tab:heuristic_policies}
  \small
  \setlength{\tabcolsep}{3pt}
  \resizebox{\textwidth}{!}{%
  \begin{tabular}{l cc cc cc}
    \toprule
    & \multicolumn{2}{c}{ImageNet Acc (\%)} & \multicolumn{2}{c}{ADE20K mIoU (\%)} & \multicolumn{2}{c}{COCO-Stuff mIoU (\%)} \\
    \cmidrule(lr){2-3} \cmidrule(lr){4-5} \cmidrule(lr){6-7}
    Evaluation policy & learned & random & learned & random & learned & random \\
    \midrule
    saliency & $64.5\pm0.2$ & $63.7\pm0.2$ & -- & -- & -- & -- \\
    saliency + IoR ($r{=}12.5\%$) & $66.7\pm0.2$ & $65.2\pm0.2$ & -- & -- & -- & -- \\
    saliency + IoR ($r{=}25\%$) & $66.7\pm0.2$ & $65.6\pm0.2$ & -- & -- & -- & -- \\
    \midrule
    entropy & -- & -- & $21.9\pm0.4$ & $21.8\pm0.2$ & $23.2\pm0.2$ & $22.7\pm0.1$ \\
    entropy + IoR ($r{=}12.5\%$) & -- & -- & $22.8\pm0.3$ & $22.3\pm0.2$ & $23.7\pm0.1$ & $23.1\pm0.1$ \\
    entropy + IoR ($r{=}25\%$) & -- & -- & $22.9\pm0.4$ & $22.4\pm0.3$ & $23.9\pm0.1$ & $23.3\pm0.2$ \\
    \midrule
    memory saliency, $\mathrm{relu}(\mathbf{H})$ & $62.5\pm0.3$ & $62.5\pm0.6$ & $20.3\pm0.3$ & $20.8\pm0.4$ & $21.2\pm0.2$ & $21.3\pm0.1$ \\
    memory saliency, $\mathrm{relu}(\mathbf{H})$ + IoR ($r{=}12.5\%$) & $64.2\pm0.3$ & $63.7\pm0.4$ & $20.7\pm0.3$ & $21.1\pm0.4$ & $21.9\pm0.2$ & $22.1\pm0.2$ \\
    memory saliency, $|\mathbf{H}|$ & $62.8\pm0.0$ & $62.2\pm0.6$ & $20.3\pm0.3$ & $20.8\pm0.4$ & $21.2\pm0.1$ & $21.3\pm0.1$ \\
    memory saliency, $|\mathbf{H}|$ + IoR ($r{=}12.5\%$) & $64.2\pm0.2$ & $63.4\pm0.4$ & $20.9\pm0.2$ & $21.0\pm0.4$ & $22.0\pm0.2$ & $22.1\pm0.2$ \\
    \bottomrule
  \end{tabular}}
\end{table}

\paragraph{Common to all heuristics.}
The heuristics are used at evaluation only and replace the policy head; nothing is retrained, and they
are applied to learned-policy and random-location models alike. The first fixation $f_1$ is the image center. After
glimpse $t$, a heuristic computes a score map $M_t$ from the model's own state, and the next fixation $f_{t+1}$ is the
center of the highest-scoring cell. All maps are computed from the memory features
$\mathbf{H}_t \in \mathbb{R}^{128 \times 64 \times 64}$ after glimpse $t$ (\Cref{eq:memory}), which
are aligned with the image and are the input of every head.

\paragraph{Content-independent policies.}
These policies use no score map, so their fixations depend neither on the image nor on the model's state.
\emph{Random locations} draws every fixation after $f_1$ independently and uniformly from the whole image.
\emph{Triangle} ($T = 4$) visits the three vertices of an equilateral triangle around the image center, each $32.5\%$ of
the image width from the center (the model sees square images, so width and height are equal): first the vertex below the center, then the upper left and the upper right one.
\emph{Quadrant} ($T = 5$) visits the centers of the four image quadrants, in the order top left, top right, bottom left,
bottom right.

\paragraph{Saliency (ImageNet only).}
The saliency heuristic of the main text (\cref{tab:policy_transfer_fused}) is the bottom-up map of
\citet{lukanov_BiologicallyInspiredDeep_2021}: the channel mean of the last convolutional feature map before global
pooling. In our classification head this is the output $\mathbf{U}_t$ of its second convolution after the ReLU
($256 \times 32 \times 32$, \Cref{tab:architecture_heads}), so $M_t = \frac{1}{256}\sum_c \mathbf{U}_{t,c}$ on a
$32 \times 32$ grid. The map is non-negative due to the ReLU.

\paragraph{Memory saliency, $\mathrm{relu}(\mathbf{H})$ and $|\mathbf{H}|$.}
These variants take the same channel mean directly on the memory features $\mathbf{H}_t$ ($64 \times 64$). Because
$\mathbf{H}_t$ is signed, they treat negative activations in two ways: $\mathrm{relu}(\mathbf{H})$,
$M_t = \frac{1}{128}\sum_c \max(\mathbf{H}_{t,c}, 0)$, counts only positive activations, as after the ReLU of the
original, and $|\mathbf{H}|$, $M_t = \frac{1}{128}\sum_c |\mathbf{H}_{t,c}|$, counts both signs as activity.

\paragraph{Entropy (ADE20K, COCO-Stuff).}
The entropy heuristic of the main text (\cref{tab:policy_transfer_fused}) fixates where the current segmentation is
least certain. The segmentation head is applied to $\mathbf{H}_t$ at $64 \times 64$, before upsampling to the image
size, and a softmax over classes gives $p_t(k \mid x, y)$, the probability of class $k$ at cell $(x, y)$ after glimpse
$t$. The score of a cell is the entropy $-\sum_k p_t(k \mid x, y) \log p_t(k \mid x, y)$ in nats, averaged over
non-overlapping $4 \times 4$ blocks into $16 \times 16$ candidate locations, and the next fixation is the center of the
block with the highest score. The heuristic uses the model's own prediction only, not the label. It is inspired by the
entropy-guided policy of CanViT~\citep{canvit2026activevision}, which computes the entropy in the same way. The
difference is how the next fixation is chosen: CanViT only considers tiles it has not visited yet and moves from large
to small tiles over the episode, whereas we simply fixate the location with the highest entropy in the whole image,
always with a glimpse of the same size.

\paragraph{Inhibition of return (+IoR).}
Every map can be combined with inhibition of return: cells whose center lies within radius $r$ of any earlier
fixation, the initial center fixation included, are excluded from the argmax. We give $r$ as a fraction of the image
width. Every map is run with $r = 12.5\%$. The best heuristic per task (saliency on ImageNet, entropy on ADE20K and
COCO-Stuff) is also run with $r = 25\%$, and the main text reports these runs (\cref{tab:policy_transfer_fused}).

\clearpage
\section{Properties of learned policies}
\label{app:policy_properties}
Supporting analyses for \Cref{sec:results_properties}: the scanpath statistics for all three seeds
(\Cref{app:policy_stats_seeds}); each policy's spatial fixation prior across seeds, image domains and steps
(\Cref{app:shape_of_prior}); the saccade-amplitude autocorrelation for different scanpath lengths
(\Cref{app:autocorr_by_length}); the per-fixation policy comparison under two further metrics
(\Cref{app:perglimpse_extra}); and inhibition of return split by visit index (\Cref{app:perifixation_split}).

\subsection{Scanpath statistics across seeds}
\label{app:policy_stats_seeds}
\begin{figure}[t]\centering
  \includegraphics[width=\textwidth]{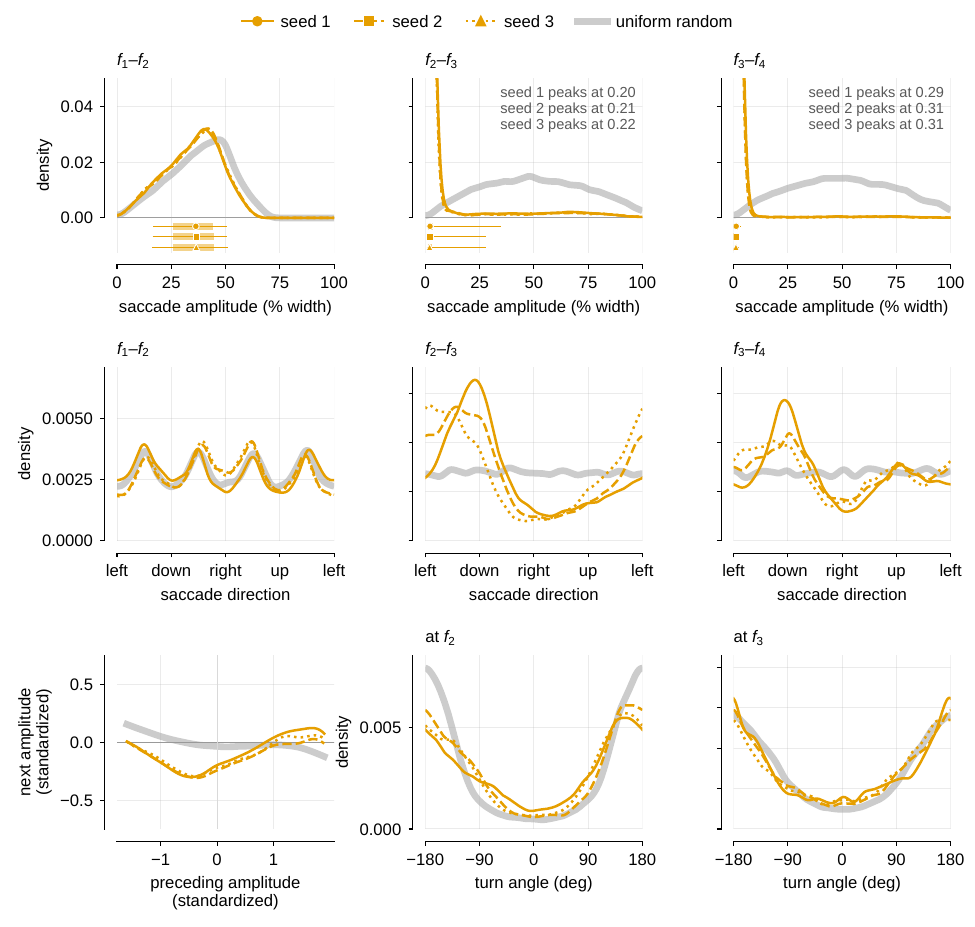}
  \caption{\textbf{Scanpath statistics across seeds, C-MNIST} (\num[group-digits=integer]{10000} validation images per seed; fixations drawn from the policy). Panels as in \cref{fig:policy_stats}; the three models trained with different random seeds share the dataset color and differ by line style (seed 1 solid, 2 dashed, 3 dotted) and marker. Thick gray line, in every panel: a uniform random policy (initial center fixation, then independent uniform fixations; simulated), the geometric reference for an image of this shape, the same simulation in every dataset's figure and in \cref{fig:policy_stats}. The $y$-axes are scaled per dataset, so the same gray line looks flatter where the models' curves are more peaked (e.g.\ the direction row of ADE20K against that of ImageNet).
}
  \label{fig:policy_stats_seeds_mnist}
\end{figure}

\begin{figure}[t]\centering
  \includegraphics[width=\textwidth]{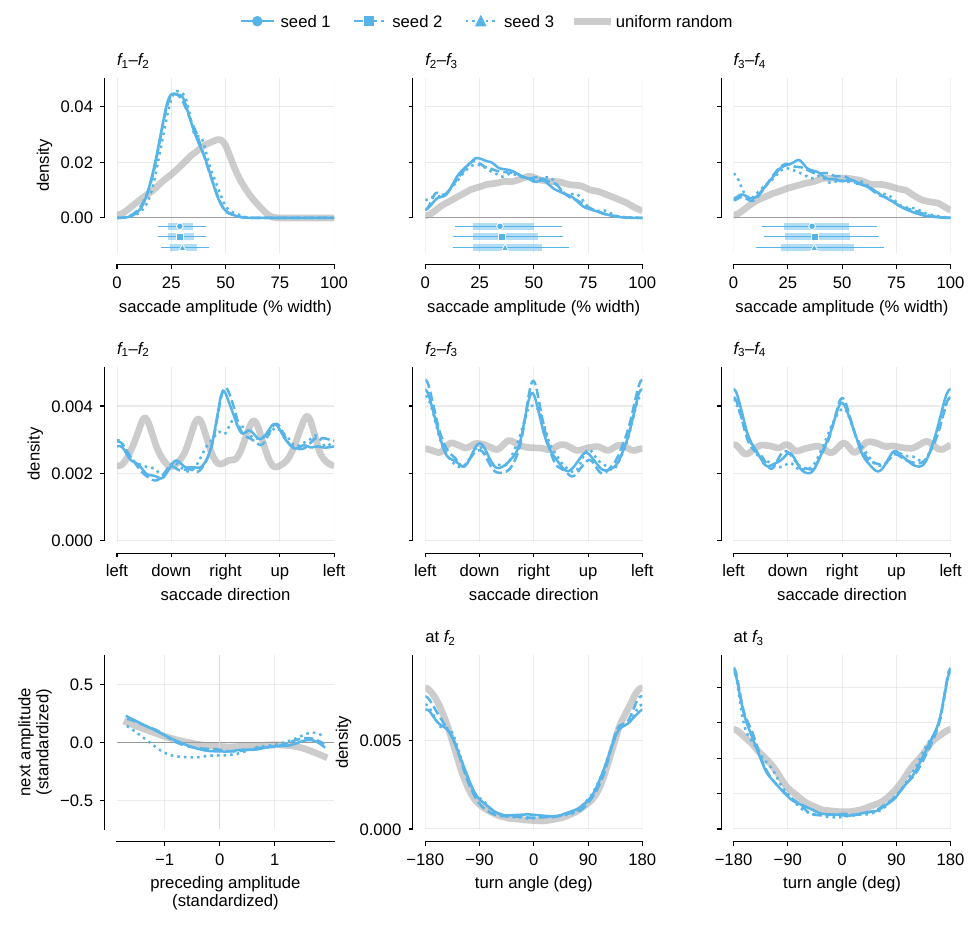}
  \caption{\textbf{Scanpath statistics across seeds, ImageNet} (\num[group-digits=integer]{10000} validation images per seed; fixations drawn from the policy). Panels, line styles and the random-policy reference as in \cref{fig:policy_stats_seeds_mnist}.
}
  \label{fig:policy_stats_seeds_imagenet}
\end{figure}

\begin{figure}[t]\centering
  \includegraphics[width=\textwidth]{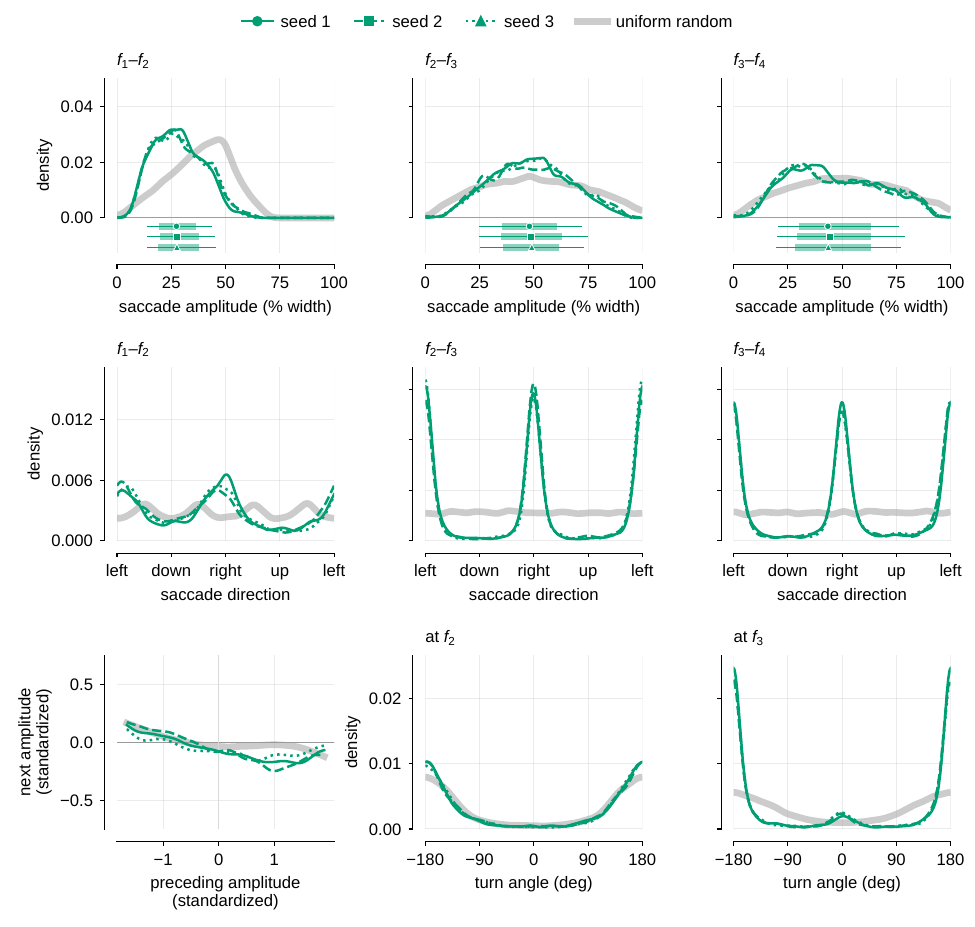}
  \caption{\textbf{Scanpath statistics across seeds, ADE20K} (\num[group-digits=integer]{2000} validation images per seed; fixations drawn from the policy). Panels, line styles and the random-policy reference as in \cref{fig:policy_stats_seeds_mnist}.
}
  \label{fig:policy_stats_seeds_ade20k}
\end{figure}

\begin{figure}[t]\centering
  \includegraphics[width=\textwidth]{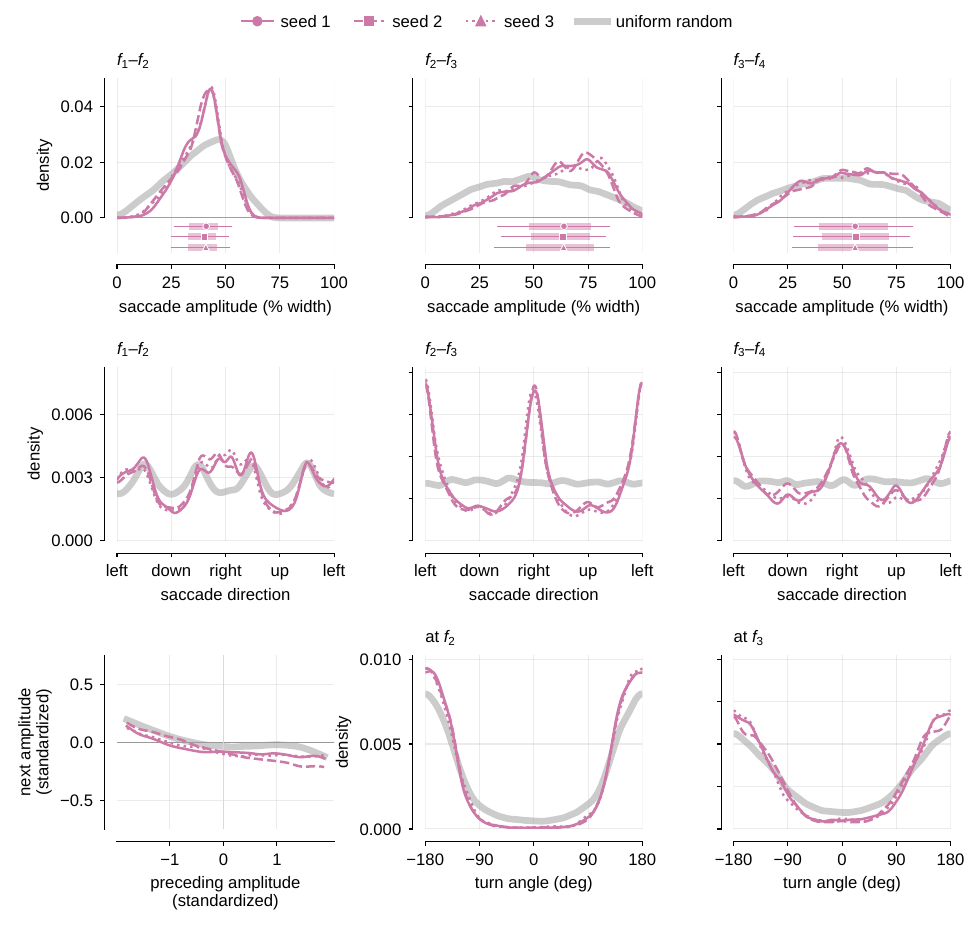}
  \caption{\textbf{Scanpath statistics across seeds, COCO-Stuff} (\num[group-digits=integer]{5000} validation images per seed; fixations drawn from the policy). Panels, line styles and the random-policy reference as in \cref{fig:policy_stats_seeds_mnist}.
}
  \label{fig:policy_stats_seeds_cocostuff}
\end{figure}

The main text reports scanpath statistics for one seed per dataset; \Cref{fig:policy_stats_seeds_mnist,fig:policy_stats_seeds_imagenet,fig:policy_stats_seeds_ade20k,fig:policy_stats_seeds_cocostuff}
show all three. On every dataset, the statistics of the three seeds are similar.

\clearpage
\subsection{More detailed analyses on cumulative fixation distributions}
\label{app:shape_of_prior}
\begin{figure}[t]\centering
  \includegraphics[width=\textwidth]{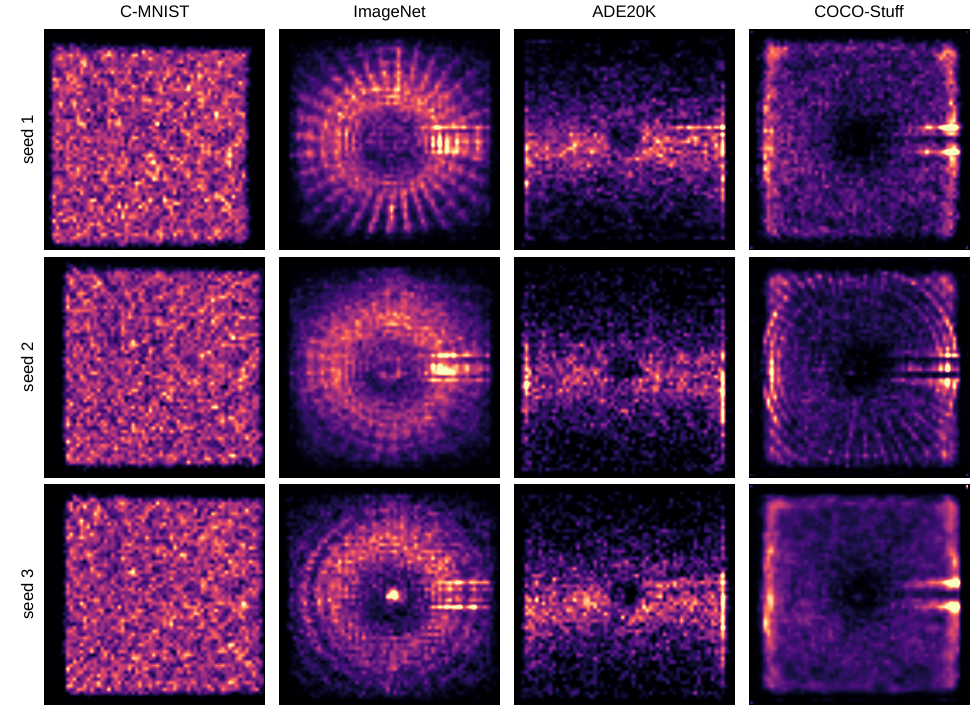}
  \caption{\textbf{Cumulative fixation distribution across seeds.} \Cref{fig:intro}B
  for the three seeds (rows) of our default models; seed 1 is the model shown in
  \cref{fig:intro}. Each panel is the mean probability map over the validation pool, with fixations drawn from
  the policy, averaged over the maps after glimpses 1 to 3, which choose fixations 2 to 4 (the initial center fixation is
  fixed and therefore left out, as is the map after the fourth glimpse). Colors are clipped per panel at the $99.5$th percentile. Images per seed: C-MNIST \num[group-digits=integer]{10000}, ImageNet \num[group-digits=integer]{10000}, ADE20K \num[group-digits=integer]{2000},
  COCO-Stuff \num[group-digits=integer]{5000}.
}
  \label{fig:cumulative_seeds}
\end{figure}

\begin{figure}[t]
  \centering
  \includegraphics[width=\textwidth]{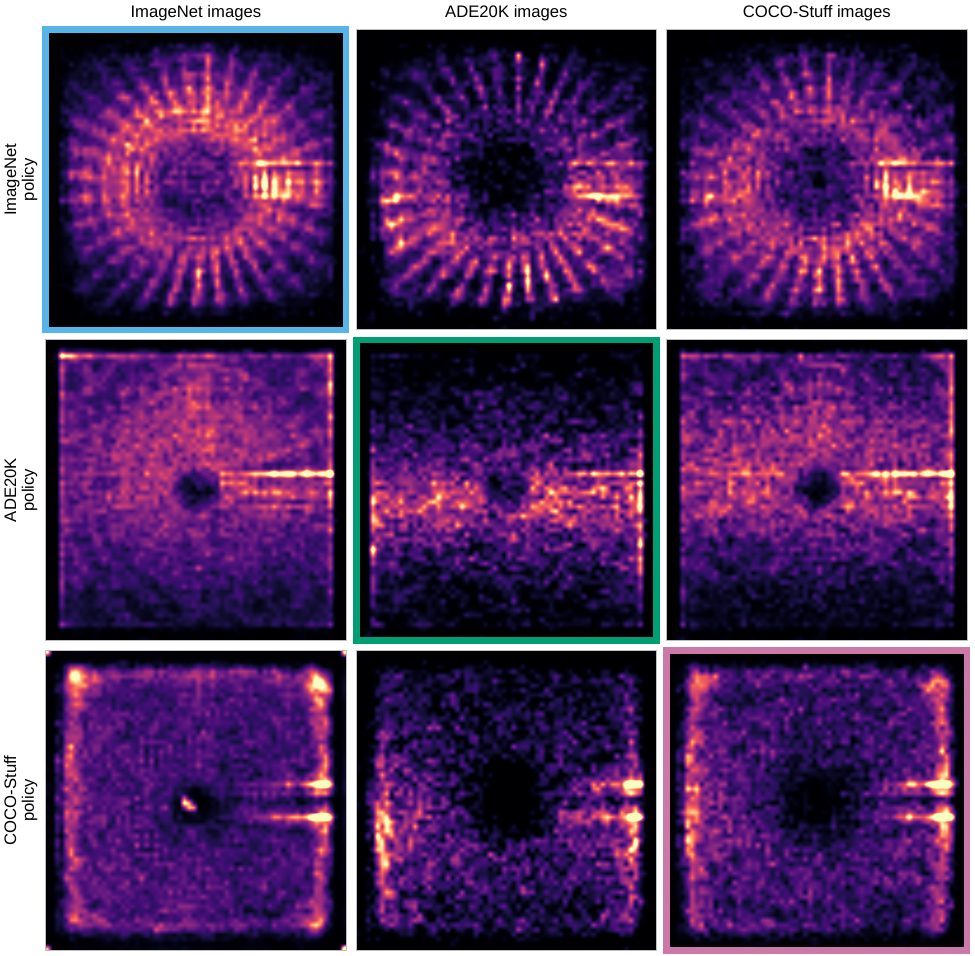}
  \caption{\textbf{Cumulative fixation distributions with the policy moved across image domains.}
  Each cell is the mean probability map over the image pool, with fixations drawn from the policy, averaged over the maps
  after glimpses 1 to 3, which choose fixations 2 to 4 (the initial center fixation is fixed and therefore left out, as is the map after the fourth glimpse), i.e.\ the quantity
  \cref{fig:intro}B plots per dataset. Rows fix the \textbf{policy} (the learned-policy
  models shown in \cref{fig:intro}), columns fix the \textbf{image domain}; images per column: ImageNet \num[group-digits=integer]{10000}, ADE20K \num[group-digits=integer]{2000}, COCO-Stuff \num[group-digits=integer]{5000}.
  Each model sees its own training resolution and all priors are compared in the shared
  $64\times64$ memory space. The diagonal (colored frame) is each policy on its own
  training domain.}
  \label{fig:centre_transfer}
\end{figure}

\begin{figure}[t]\centering
  \includegraphics[width=\textwidth]{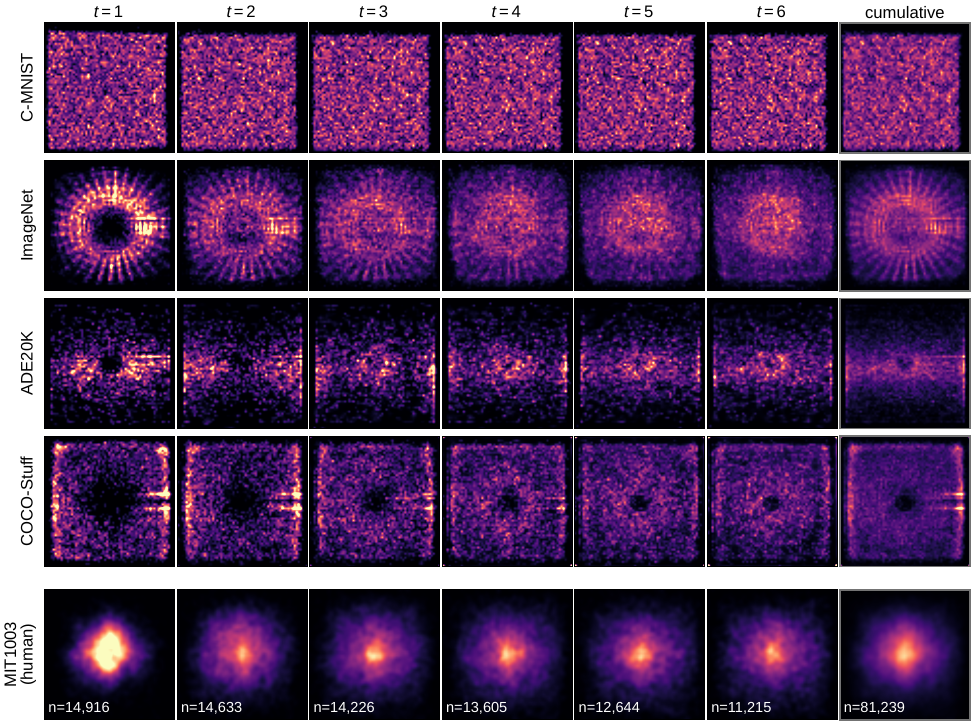}
  \caption{\textbf{Fixation distribution at each step, for the models and human observers.} \Cref{fig:intro}B,
  split by step: one row per dataset (top four: models; bottom: human observers), one column per step from $t = 1$ to
  $t = 6$, and the cumulative map. The models are trained with $T = 4$ and unrolled to $T_\text{eval} = 7$ glimpses with fixations drawn from the policy: after the initial center fixation $f_1$, the map at step $t$ chooses $f_{t+1}$, so the six steps place $f_2$ to $f_7$. Each map is the mean of
  the per-step distributions over the validation images (C-MNIST and ImageNet \num[group-digits=integer]{10000}, ADE20K
  \num[group-digits=integer]{2000}, COCO-Stuff \num[group-digits=integer]{5000}); the color scale is shared within a row
  and clipped at the $99.5$th percentile. The cumulative column averages all six steps ($t = 1$ to $6$), more than
  the cumulative map of \Cref{fig:intro}B ($t = 1$ to $3$). \emph{Bottom row:} MIT1003
  free viewing; column $t$ is the density of the fixation $f_{t+1}$, as a $64\times64$
  histogram blurred by one cell, and $n$ is the number of fixations per panel.
}
  \label{fig:perglimpse_prior}
\end{figure}

\Cref{fig:intro}B in the main paper shows each policy's cumulative fixation prior on its
own dataset. \Cref{fig:cumulative_seeds} repeats it for all three seeds. \Cref{fig:centre_transfer} applies each policy
to each image domain, policy along the rows, image domain along the columns, so the fixation distribution of one policy
can be compared across image domains, and the distributions of different policies on the same images.
\Cref{fig:perglimpse_prior} splits the prior of \cref{fig:intro}B by step, including human observers.

\clearpage
\subsection{Saccade-amplitude autocorrelation by scanpath length}
\label{app:autocorr_by_length}
\begin{figure}[t]
  \centering
  \includegraphics[width=\textwidth]{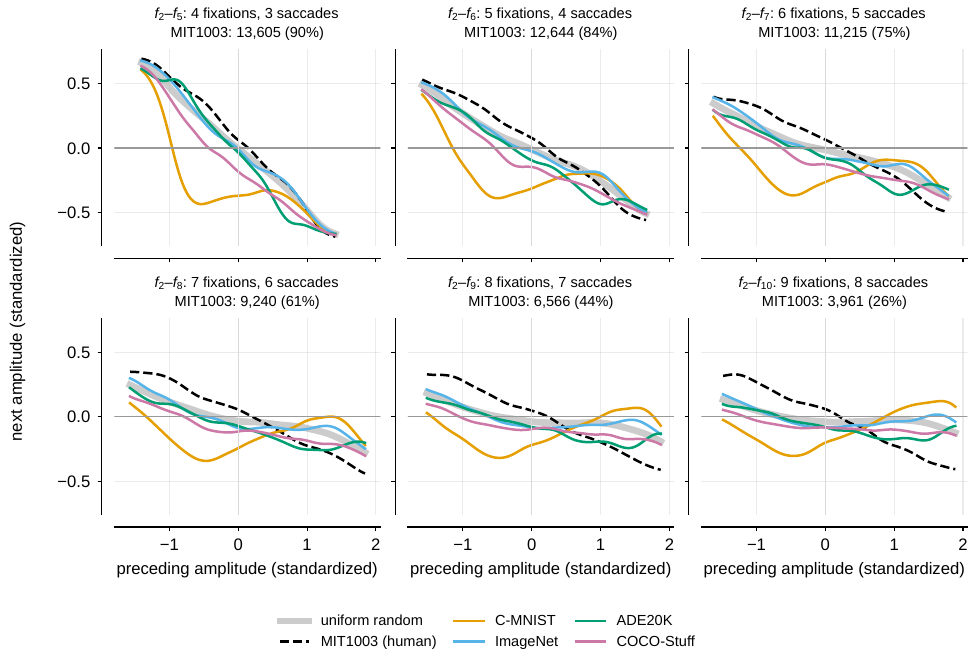}
  \caption{\textbf{Autocorrelation of saccade amplitudes.} Mean amplitude of the next saccade as a
  function of the preceding one, both z-standardized within each scanpath (own mean subtracted, divided by own
  std), smoothed with a Gaussian kernel (bandwidth $0.15$) and drawn over the range that lies within
  the $2$nd to $98$th percentile of the preceding amplitude for every source in the panel. Each panel uses the
  fixations $f_2$ to $f_{k+1}$ for one $k = 4, \dots, 9$, i.e.\ $k-1$ saccades. The initial saccade from the center $f_1$ to the next fixation $f_2$ is excluded since its starting position is fixed. For MIT1003, $f_2$ to $f_{k+1}$ are the first $k$ fixations after the initial center fixation, and each panel uses all scanpaths with at least $k$ of
  them (counts in the panel titles). One learned-policy model per dataset (the one shown in \cref{fig:intro}, trained with four glimpses) is unrolled for nine fixations ($f_2$ to
  $f_{10}$), drawn from the policy after the initial center fixation, and cut to $f_2$ to $f_{k+1}$. Thick gray line: uniform random policy.}
  \label{fig:after_long_by_length}
\end{figure}

The autocorrelation plot of \cref{fig:policy_stats}C standardizes the saccade amplitudes within each scanpath, which makes the results dependent on the scanpath length (see next paragraph). Standardization is still necessary in depicting the relationship to account for differences in saccade amplitudes across images. \Cref{fig:after_long_by_length} shows the same plot as in \cref{fig:policy_stats}C for different scanpath lengths. For all plots, the mean amplitude of the next saccade is shown as a
function of the preceding one, both z-standardized within their scanpath. The last panel is identical to \cref{fig:policy_stats}C. It can be clearly seen that the slopes of the curves depend on the scanpath length, which is why the uniformly random policy is needed as a baseline to interpret the results. An intuition of this dependence is illustrated below.

The curve of the random policy slopes downward, more steeply for short scanpaths. Its slope reflects the correlation between
consecutive amplitudes after they are standardized within each scanpath. This correlation results from two opposing
effects: (1)~Consecutive saccades share a fixation, so even when fixations are placed at uniformly random locations, their raw amplitudes are
positively correlated  at roughly $+0.11$.
(2)~Standardizing within a scanpath works against this: subtracting the scanpath's own mean makes its deviations
sum to zero, so a saccade above that mean has to be balanced by others below it. For $n$ independent
amplitudes this gives every pair of deviations a correlation of $-1/(n-1)$, i.e.\ $-0.5$ for three saccades
and $-0.14$ for eight, and the bias vanishes as $n$ grows. For short scanpaths it dominates: under the random
policy the correlation of consecutive standardized amplitudes is $-0.48$ with three saccades, $-0.19$ with five
and $-0.06$ with eight. It crosses zero at about twelve saccades and approaches the raw $+0.11$ for long
scanpaths ($+0.10$ with $100$). Part of the downward slope of every curve therefore comes from the standardization itself, which is why the random
policy at each scanpath length is needed for comparison.

\clearpage
\subsection{Additional per-fixation policy comparisons}
\label{app:perglimpse_extra}
\begin{figure}[t]
  \centering
  \includegraphics[width=\textwidth]{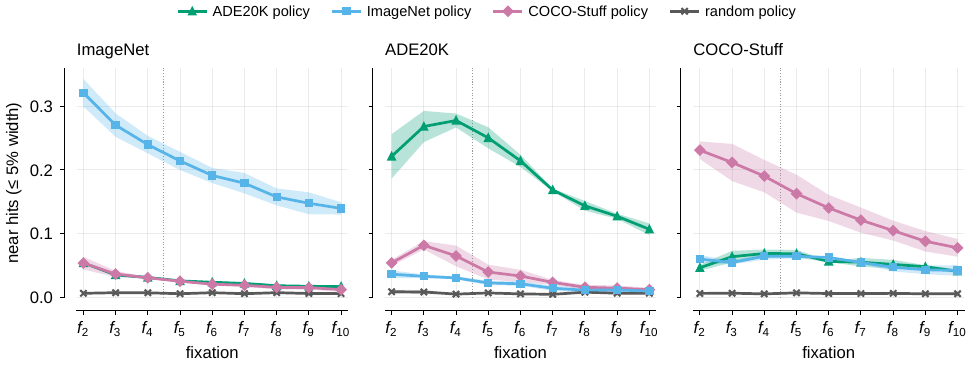}
  \caption{%
    \textbf{Per-fixation policy alignment: near hits.}
    Fraction of images on which the two policies' argmax fixations lie within $5\%$ of the image
    width of each other (about three cells of the $64\times64$ probability map; higher = more aligned).
    It thresholds the per-image distances that \cref{fig:argmax} averages. On every dataset and at every fixation,
    the two seeds of the same task give the most near hits, the other tasks' policies fewer and
    uniform fixations the fewest. Layout as \cref{fig:argmax}: every model is unrolled for nine fixations
    ($f_2$--$f_{10}$) after the initial center fixation; the dotted line marks the end of the $T = 4$ glimpses used in training.
    Shading: $\pm1$ std over three model pairs, each averaged over the validation images (ImageNet
    \num[group-digits=integer]{10000}, ADE20K \num[group-digits=integer]{2000}, COCO-Stuff \num[group-digits=integer]{5000}).
  }
  \label{fig:perglimpse_hit5}
\end{figure}

\begin{figure}[t]
  \centering
  \includegraphics[width=\textwidth]{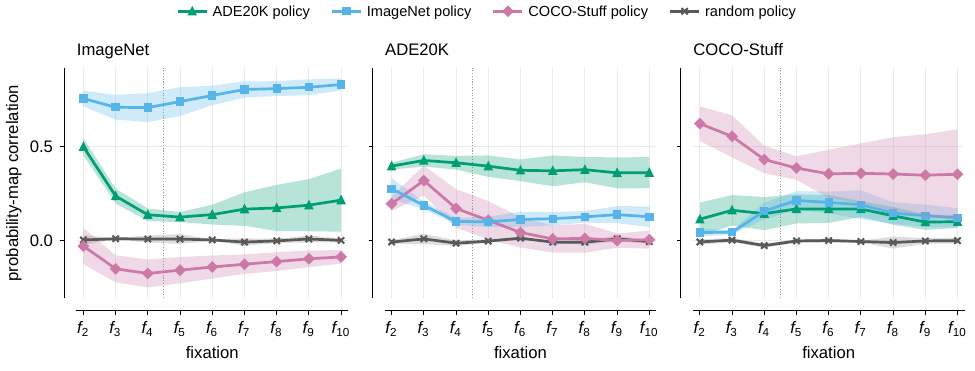}
  \caption{%
    \textbf{Per-fixation policy alignment: probability-map correlation.}
    Correlation between the two policies' next-fixation probability maps, each first averaged
    over the images (higher = more aligned): the
    least localized of the three metrics, and a general impression of where each policy
    puts its mass rather than of where its peak falls for a specific image. Layout and shading as in \cref{fig:perglimpse_hit5}.
  }
  \label{fig:perglimpse_marg_cc}
\end{figure}

\Cref{fig:perglimpse_hit5,fig:perglimpse_marg_cc} compare the policies under the protocol of \Cref{fig:argmax} with two
further similarity metrics. Near hits (\cref{fig:perglimpse_hit5}) take the same per-image argmax distances as
\Cref{fig:argmax} but only count whether the two fixations lie within $5\%$ of the image width of each other. The
probability-map correlation (\cref{fig:perglimpse_marg_cc}) compares the policies globally: it first averages probability maps across all images, and then correlates these averaged probability maps, so it reflects where each policy puts its mass rather than where it fixates on a given image.

\subsection{Inhibition of return, split by visit index}
\label{app:perifixation_split}
\begin{figure}[t]
  \centering
  \includegraphics[width=\textwidth]{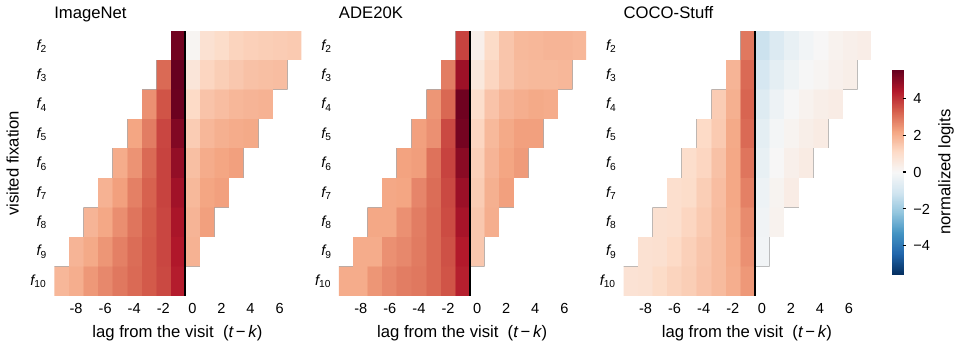}
  \caption{\textbf{Inhibition of return, split by visit index.}
  The quantity of \cref{fig:inhibition} with one row per visited fixation $f_k$ ($k = 2, \dots, 10$),
  one panel per dataset and a shared color scale. Fixations are drawn from the policy.
  Values are means over the three seeds of each of our default models.
  The black vertical line separates the steps before the visit (negative lags) from the visit and the steps after it.}
  \label{fig:perifixation}
\end{figure}

\begin{figure}[t]
  \centering
  \includegraphics[width=\textwidth]{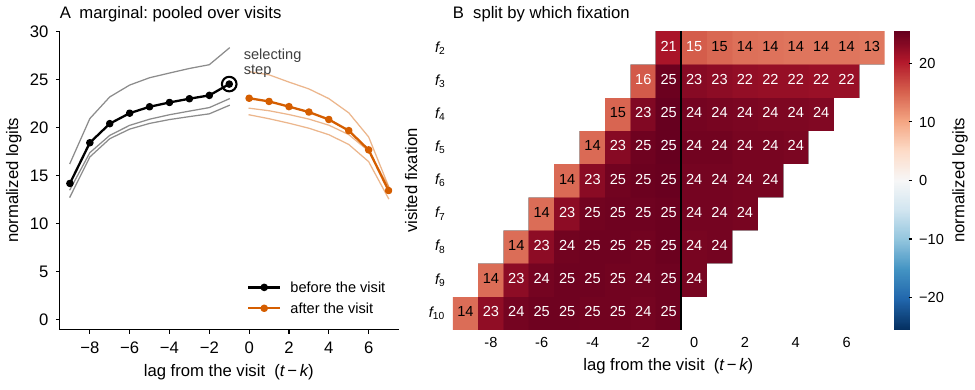}
  \caption{\textbf{Inhibition of return on C-MNIST}, shown on its own because its
  values are an order of magnitude larger: its probability maps put nearly all their mass on a single
  cell.
  The policy logit $z$ at the location of $f_2$, the first fixation after the initial center fixation, tends to drop sharply at the visit ($z=20.9\!\to\!15.4$)
  and stays low afterwards (seen in panel B). From $f_3$ on, the policy tends to keep returning to the same
  location, and the logit at the visited cell drops by at most $1.7$ (seen in panel B), so the marginal
  (seen in panel A) shows little suppression. This reflects the fact that the policy has a tendency to keep looking at the digit once it found it. Inhibition of return therefore does not emerge in the same way in this setting where discriminatory information is highly localized. Values are means over the three seeds of each of our default models. In (A), thin lines are the individual seeds.
  The black vertical line separates the steps before the visit (negative lags) from the visit and the steps after it.}
  \label{fig:perifixation_mnist}
\end{figure}

The main paper reports this effect marginally, averaged over visits. \Cref{fig:perifixation} shows the
same quantity with one row per visit index: the drop at the visit and the partial recovery afterwards
appear in every row. \Cref{fig:perifixation_mnist} shows C-MNIST separately.

\clearpage
\section{Overview of the training-recipe ablations}
\label{app:recipe_ablations}
\Cref{app:detach_ablation,app:loss_choices} examine alternatives to choices of the training recipe that the main text
presents. \Cref{app:choices_examined} lists the choices that these appendices vary on purpose, each with the setting of our default models and the alternative examined. \Cref{app:regimes} lists all the ways in which a run in
\Cref{app:detach_ablation,app:loss_choices} can differ from our default models. Besides the choices varied
on purpose, these include differences due to the design of an analysis or to saving compute, for example by using
runs from earlier stages of model development.

\subsection{The choices examined}
\label{app:choices_examined}

Our default models, i.e.\ the models of the main results
(\Cref{sec:results_native,sec:results_transfer,sec:results_properties}), are trained and evaluated as described in
\Cref{sec:model_training} and \Cref{app:training_details}. This main protocol is summarized at the
beginning of \Cref{app:regimes}. \Cref{app:detach_ablation,app:loss_choices} vary four main choices of the training
recipe of the main protocol, listed in \Cref{tab:reward_choices}: (1)~whether the policy
gradient reaches backbone and memory, (2)~whether entropy regularization is added, (3)~how the task is supervised, and
(4)~how a fixation is rewarded. For each choice, the table gives the setting of our default models, the
alternative we examine, and the appendix in which the alternative is analyzed. By default, the stop-gradient
keeps the policy gradient out of backbone and memory, since letting it through was shown to be detrimental in
\Cref{sec:gradient_conflicts}. \Cref{app:detach_ablation} examines this choice in more detail, together with entropy
regularization (choices (1) and (2)). As further remedies without the stop-gradient, \Cref{app:entropy_new} also tests
a smaller policy-loss weight and a smaller policy head. \Cref{app:loss_choices} compares the task supervision and the reward with their
end-of-episode alternatives (choices (3) and (4)).

\begin{table}[t]
  \centering
  \caption{\textbf{Alternatives to the training recipe investigated in \Cref{app:detach_ablation,app:loss_choices}.} For each choice of the training recipe, the table gives the setting used in our default models, the alternative we compare it with, and the subsection that reports the results. $\mathrm{CE}_t = \mathcal{L}_{\text{CE}}(\hat{\mathbf{y}}_t, \mathbf{y})$ is the task loss after glimpse $t$.}
  \label{tab:reward_choices}
  \small
  \resizebox{\linewidth}{!}{%
  \begin{tabular}{llll}
    \toprule
    Choice & Default & Alternative & Analyzed in \\
    \midrule
    Policy gradient   & stop-gradient on $\mathbf{H}_t$~(\cref{eq:policy_loss}) & full gradient & \cref{app:detach_ablation} \\
    Entropy regularization & none & entropy target ($\mathcal{H}^{\star} = 3.5$ nats) & \cref{app:entropy_new} \\
    Task supervision  & mean per-step CE~(\cref{eq:task_loss}) & last step only & \cref{app:task_loss_choices} \\
    Reward            & $r_t = \mathrm{CE}_t - \mathrm{CE}_{t+1}$~(\cref{eq:reward}) & $-\mathrm{CE}_T$ for every fixation & \cref{app:policy_loss_choices} \\
    \bottomrule
  \end{tabular}%
  }

\end{table}

\subsection{Deviations from the main protocol}
\label{app:regimes}

\textbf{The main protocol} is how our default models are trained and evaluated. Training uses the recipe of \Cref{sec:model_training} with the hyperparameters of \Cref{tab:hparams}: the
task loss at every step (\Cref{eq:task_loss}), the per-step reward of \Cref{eq:reward} without a baseline, its
standardization per step (\Cref{eq:reward_norm}), no entropy regularization, and the stop-gradient on the memory
features (\Cref{eq:policy_loss}). Evaluation uses the last checkpoint with argmax fixations
(\Cref{app:training_details}).

\textbf{Deviations.} The runs in \Cref{app:detach_ablation,app:loss_choices} differ from the main protocol for three
reasons: (1)~The alternative under investigation, such as the policy gradient on backbone and memory. (2)~The
design of an analysis, such as scoring policies on a common frozen model. (3)~Saving compute: for example, some runs are
trained only as long as needed to show a difference, or some runs come from earlier stages of model development (where the protocol differed) and were
not rerun under the main protocol to save compute.
\Cref{tab:regimes} lists all deviations.

\begin{table}[t]
  \centering
  \caption{\textbf{Numbered deviations from the main protocol.} Every figure and table caption in \Cref{app:detach_ablation,app:loss_choices} lists the numbers of the deviations that apply to its runs. Investigated settings without a number (the small policy head, the entropy target and the policy-loss weight $\lambda$) are named in the captions themselves. A result given only in the text either belongs to the runs of a cited figure or table, or lists its deviations in a footnote.}
  \label{tab:regimes}
  \small
  \setlength{\tabcolsep}{4pt}
  \begin{tabular}{p{0.04\textwidth}p{0.30\textwidth}p{0.56\textwidth}}
    \toprule
    & Deviation & Description \\
    \midrule
    D1 & Terminal reward & Every fixation receives the same reward, $-\mathrm{CE}$ at the last step,
    instead of its own per-step improvement (\Cref{eq:reward}) \\
    D2 & Terminal supervision & The task loss is computed only on the prediction after the last glimpse,
    instead of after every glimpse (\Cref{eq:task_loss}) \\
    D3 & Level normalizer & The reward is divided by its mean absolute value, pooled over batch
    and step, instead of being standardized per step (\Cref{eq:reward_norm}) \\
    D4 & Gradient on backbone and memory & The policy gradient reaches backbone and memory, instead of
    stopping at the memory features (stop-gradient in \Cref{eq:policy_loss}) \\
    D5 & Frozen representation & Only the policy head trains, on a frozen backbone, memory and task head,
    instead of all parameters; accuracy then depends on the fixations alone \\
    D6 & Early epoch & Evaluated at a matched epoch before convergence, instead of at the end of the
    full schedule (\Cref{tab:hparams}) \\
    D7 & Best checkpoint & The maximum over validation evaluations, instead of the last checkpoint
    (\Cref{app:training_details}) \\
    D8 & Baseline present & A running-mean or learned value baseline is subtracted from the reward,
    instead of no baseline \\
    D9 & Different pretrained models & The compared models are finetuned from different pretrained models,
    instead of one common starting point, so part of a difference between them can come from these
    starting points rather than from the setting being compared \\
    D10 & Policy-transfer score & A common frozen receiving model's accuracy under externally supplied
    fixations, instead of the model's own metric \\
    D11 & Reduced budget & Fewer GPUs, a $24$~h wall clock or a truncated schedule, instead of the
    setup of \Cref{tab:hparams} \\
    D12 & Logged validation score & The validation score logged during training, with fixations drawn
    from the policy, instead of an evaluation with argmax fixations (\Cref{app:training_details}) \\
    D13 & Different learning-rate schedule & A constant learning rate, or one halved whenever the validation
    score stops improving, instead of the schedule of \Cref{tab:hparams} \\
    D14 & Entropy regularization & Entropy regularization $-\lambda_{\mathcal{H}} \bar{\mathcal{H}}$ with a fixed
    weight $\lambda_{\mathcal{H}} = 5\times10^{-3}$ is added to the loss, instead of none \\
    \bottomrule
  \end{tabular}
\end{table}

\clearpage
\section{The policy gradient on backbone and memory}
\label{app:detach_ablation}

Our default models keep the policy gradient out of backbone and memory with a stop-gradient on the memory features
(\Cref{eq:policy_loss}). \Cref{sec:gradient_conflicts} then ablates this stop-gradient operation. This appendix adds additional information to this ablation: it shows what goes
wrong when the policy gradient reaches backbone and memory, and which mitigations appear to help and which appear not to. \Cref{app:joint_cost} reports the drop in task performance without the
stop-gradient. \Cref{app:pg_cause} finds that worse fixations, gradient-norm clipping and the AdamW optimizer state are
unlikely to explain this drop, which suggests that the policy gradient on backbone and memory itself causes it.
\Cref{app:why_it_grows} follows how this gradient changes once the policy sharpens (meaning its logits grow so large that it
fixates almost deterministically, though still at image-dependent locations). \Cref{app:bf16} describes a separate
problem of the runs trained in bf16. \Cref{app:entropy_new} tests potential remedies: a smaller policy-loss weight, a
small policy head and an entropy target.
\Cref{app:alternatives} summarizes the findings and contextualizes our solution of using stop-gradient to prevent optimization issues. \Cref{app:pg_derivation} derives the bound used in \Cref{app:why_it_grows}.

The central comparison of this appendix (\Cref{tab:joint_ablation}) is between runs with and without the
stop-gradient that are otherwise trained alike. On C-MNIST, ADE20K and COCO-Stuff, the runs with stop-gradient are our default models. On ImageNet, the run with stop-gradient is our default ImageNet model after 75 of its 150 epochs, before its learning
rate is first halved, so it reaches a lower accuracy than in \Cref{tab:main_combined_cur}. To save compute, the
runs without stop-gradient are trained only for these 75 epochs. Later subsections also use other runs. How these differ from our default models is listed in the captions of the
figures and tables that show them, or in a footnote for runs reported only in the text (\Cref{tab:regimes}).

\subsection{Without the stop-gradient, task performance drops}
\label{app:joint_cost}
Without the stop-gradient (\emph{w/o stop-grad}), task performance drops on all four datasets
(\Cref{tab:joint_ablation}): accuracy by $36.1\%$ on ImageNet and $0.5\%$ on C-MNIST, and mIoU by $6.2$ on ADE20K
and $7.0$ on COCO-Stuff. In the models without the stop-gradient, the policy-loss weight
$\lambda$ is $0.1$ by default. The analyses of the following subsections cover ImageNet, ADE20K and COCO-Stuff,
where the drop is large.

\begin{table}[t]
  \centering
  \caption{\textbf{Task performance with the policy gradient on backbone and memory.} \emph{With stop-grad}
  (reference): the policy gradient stops at the memory features (our default models).
  \emph{w/o stop-grad}: no stop-gradient. \emph{w/o stop-grad + small head}, \emph{w/o stop-grad + entropy}: no
  stop-gradient, with one of the two variants of \Cref{app:entropy_new} (a $129$-parameter policy head or an entropy
  target of $3.5$ nats). On C-MNIST, ADE20K and COCO-Stuff, \emph{with stop-grad} are our default models. On ImageNet,
  \emph{with stop-grad} is our default ImageNet model after 75 of its 150 epochs, and the other rows are trained only for these 75 epochs. $\pm$: std over seeds where several exist.
  Within each column the runs share epochs, learning-rate schedule, batch size and initialization and differ only in the setting named in each row. \textbf{Deviations (\Cref{tab:regimes}):} gradient on backbone and memory (D4; all rows except \emph{with stop-grad}), reduced budget (D11; ImageNet), logged validation score (D12).}
  \label{tab:joint_ablation}
  \small
  \setlength{\tabcolsep}{5pt}
  \begin{tabular}{l cccc}
    \toprule
    & C-MNIST & ImageNet & ADE20K & COCO-Stuff \\
    & Acc (\%) & Acc (\%) & mIoU (\%) & mIoU (\%) \\
    \midrule
    \emph{with stop-grad} & $99.3\pm0.0$ & $61.6$ & $24.3\pm0.2$ & $24.9\pm0.2$ \\
    \midrule
    w/o stop-grad              & $98.8\pm0.1$ & $25.5$ & $18.1\pm0.8$ & $17.9\pm0.5$ \\
    w/o stop-grad + small head & $95.1$ & $61.4$ & $23.1$ & $24.1$ \\
    w/o stop-grad + entropy    & $99.3$ & $61.9$ & $23.8$ & $23.5$ \\
    \bottomrule
  \end{tabular}
\end{table}

\begin{table}[t]
  \centering
  \caption{\textbf{The policies of \cref{tab:joint_ablation}, transferred to one random-location model.} In \cref{tab:joint_ablation} each row changes both the policy \emph{and} backbone and memory, so a low cell there can come from either. Here the receiving model is frozen: in every row, the dataset's default random-location model is evaluated on its validation set, and only the fixations $f_2$--$f_4$ it is given change ($f_1$ is always the image center). Row~1 is that model under the uniform-random policy it was trained with, so each $\Delta$ is the value of the policy alone.   \textbf{Deviations (\Cref{tab:regimes}):} gradient on backbone and memory (D4; the policies without stop-gradient), policy-transfer score (D10), reduced budget (D11; the ImageNet policies).}
  \label{tab:xpol_jointabl}
  \small
  \setlength{\tabcolsep}{4pt}
  \begin{tabular}{l cc cc cc cc}
    \toprule
    & \multicolumn{2}{c}{C-MNIST} & \multicolumn{2}{c}{ImageNet} & \multicolumn{2}{c}{ADE20K} & \multicolumn{2}{c}{COCO-Stuff} \\
    \cmidrule(lr){2-3} \cmidrule(lr){4-5} \cmidrule(lr){6-7} \cmidrule(lr){8-9}
    & Acc (\%) & $\Delta$ & Acc (\%) & $\Delta$ & mIoU (\%) & $\Delta$ & mIoU (\%) & $\Delta$ \\
    \midrule
    \emph{random locations} & $90.5$ & -- & $63.6$ & -- & $22.2$ & -- & $23.0$ & -- \\
    \midrule
    with stop-grad & $98.3$ & $+7.8$ & $66.0$ & $+2.5$ & $23.0$ & $+0.8$ & $23.9$ & $+0.8$ \\
    w/o stop-grad & $98.3$ & $+7.8$ & $64.9$ & $+1.4$ & $23.2$ & $+1.0$ & $24.1$ & $+1.1$ \\
    w/o stop-grad + small head & $97.7$ & $+7.2$ & $66.3$ & $+2.8$ & $23.0$ & $+0.8$ & $24.3$ & $+1.3$ \\
    w/o stop-grad + entropy & $98.4$ & $+7.9$ & $66.2$ & $+2.7$ & $23.3$ & $+1.1$ & $24.3$ & $+1.3$ \\
    \bottomrule
  \end{tabular}
\end{table}

\subsection{Worse fixations, clipping and the optimizer state are unlikely to explain the drop in task performance}
\label{app:pg_cause}
Without the stop-gradient, the runs of \Cref{tab:joint_ablation} differ from those with it in one respect: the policy
gradient also reaches backbone and memory. The drop in task performance can therefore come from this gradient directly,
or from what changes it induces: the fixations the policy chooses, how often gradient-norm clipping acts, and the
AdamW optimizer state. Whether worse fixations explain the drop is checked with the accuracy after the first glimpse and
with the policy transfers of \Cref{tab:xpol_jointabl}; whether clipping or the optimizer state explains it, with
statistics logged during training.

\paragraph{Worse fixations do not play a major role.}
In all our models, which locations are fixated changes task performance by a few points: our default ImageNet model, for
example, loses $5.2\%$ when it is evaluated with random locations instead of its own policy
(\Cref{tab:policy_transfer_fused}). Even this drastic change of the fixations costs far less than the $36.1\%$ of accuracy that the ImageNet model loses without the stop-gradient, which suggests that the drop is not mainly the result of worse fixations. Also, after the first glimpse, which is always at the image center and so does not depend on the policy, the ImageNet model without stop-gradient of \Cref{tab:joint_ablation} reaches only $21.4\%$ accuracy against $52.1\%$ for the model with stop-gradient. Transferred to a frozen
random-location model, the policy trained without stop-gradient improves on random locations about as much as the policy
trained with stop-gradient on three datasets, and $1.1\%$ less on ImageNet ($1.4$ against $2.5\%$;
\Cref{tab:xpol_jointabl}). Most of the $36.1\%$ accuracy drop on ImageNet therefore cannot be the result of worse fixations.

\begin{figure}[t]
  \centering

  \includegraphics[width=\textwidth]{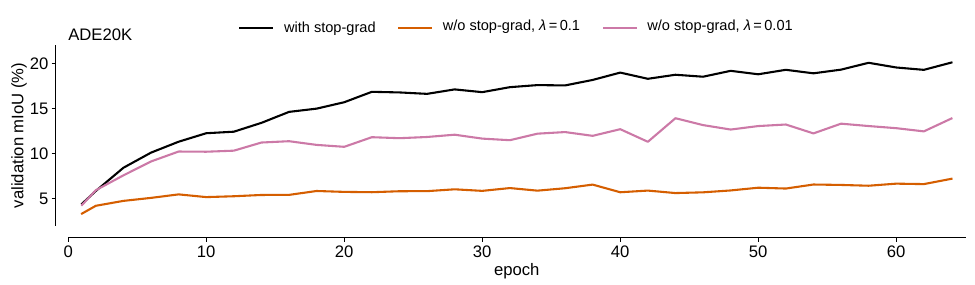}
  \caption{\textbf{Training variants on ADE20K.} Validation mIoU of the run with stop-gradient, a run without
  stop-gradient at policy-loss weight $\lambda = 0.1$ and one at $\lambda = 0.01$, otherwise identical. First $65$ epochs; single runs.
  \textbf{Deviations (\Cref{tab:regimes}):} gradient on backbone and memory (D4; models without stop-gradient), reduced budget (D11; the run at $\lambda = 0.01$ ends at epoch~$64$), logged validation score (D12).}
  \label{fig:ade_interventions}
\end{figure}

\paragraph{Clipping and optimizer state are also unlikely to affect task performance substantially.}
Large policy gradients could slow learning through two side effects, without having a negative effect on backbone and memory per se. (1)~Gradient-norm
clipping (\Cref{tab:hparams}) scales down the whole update, including the task part, whenever the gradient norm exceeds
its threshold. (2)~AdamW divides each step by the square root of a running average of the squared gradients. This average
forgets slowly, so a single large gradient keeps it high for many steps and makes the following steps smaller, including
those of the task gradient. This might influence optimization since we show that the policy gradient on backbone and memory arrives in rare, large
bursts without the stop-gradient (\Cref{app:why_it_grows}). However, we find that neither of these factors explains the drop in task performance. As for (1), clipping cannot have started the drop: on ImageNet, the run without stop-gradient of
\Cref{tab:joint_ablation} is already behind at epoch~4 in the validation score logged during training ($15.3\%$ against
$26.9\%$), before clipping appears in the logged minibatches (one per epoch) at epoch~5. On
ADE20K, a run without stop-gradient with a $10\times$ smaller policy-loss weight ($\lambda = 0.01$ instead of the default
$0.1$; \Cref{fig:ade_interventions}) falls behind
the run with stop-gradient after about eight epochs, although clipping acts on none of its steps up to epoch~9 and on at most $2\%$
afterwards. Switching clipping off does not help either: in an earlier ablation on ImageNet (single runs,
stopped after about $16$ epochs), two otherwise identical runs, both without stop-gradient, reached $23.1\%$ without
clipping and $24.4\%$ with it, against $49.0\%$ for the run with stop-gradient in the same set at the same epoch.\footnote{An earlier set of ImageNet runs with policy-loss weight $\lambda = 1$, trained for $24$~h on one GPU. \textbf{Deviations (\Cref{tab:regimes}):} level normalizer (D3), gradient on backbone and memory (D4), early epoch (D6), baseline present (D8), reduced budget (D11), logged validation score (D12), different learning-rate schedule (D13), entropy regularization (D14). In addition, these runs use an earlier version of the backbone and a policy head with a single $1\times1$ convolution.} As for (2), smaller AdamW steps would act like a lower learning rate, so a run that falls behind should show a clearly
larger square root of the running average, by which each step is divided, than runs that keep learning. On backbone and
memory this square root is within about $1.5\times$ of that of the run with stop-gradient in the $\lambda = 0.01$ run, which falls behind, and within about $2\times$ in the two variants of
\Cref{app:entropy_new}, which keep learning (compared at matched checkpoints within the first $37$ epochs). Smaller AdamW steps therefore do not seem to explain the drop in task performance either.

\subsection{The policy gradient arrives in rare, large bursts once the policy sharpens}
\label{app:why_it_grows}
\Cref{app:pg_cause} suggests that the policy gradient on backbone and memory causes the drop in task performance. This
subsection describes how this gradient develops during training without the stop-gradient. It first gives an upper bound
on the size of this gradient (derived in \Cref{app:pg_derivation}). Two factors of the bound can change during training:
the sharpness of the policy (how close it comes to fixating one location per image with certainty) and the norm of the
policy head's Jacobian. For a sharp policy, the bound predicts that most samples send almost no gradient and a few send a
large one; the Jacobian norm sets how large these few gradients can be. The subsection then follows both factors through training and shows how the gradient is distributed over the
samples.

\paragraph{What the bound predicts.}
For one fixation drawn from the policy (one image and one decision step), the policy gradient at the memory features $\mathbf{H}_t$ has
norm at most $\frac{\lambda}{B}\,|\tilde r_t|\,\|\mathbf{J}_t\|\,\sqrt{2}\,(1 - \pi_{f})$ (\Cref{eq:pg_bound_features},
derived in \Cref{app:pg_derivation}). Here $\lambda$ is the weight of the policy loss, $B$ the batch size, $\tilde r_t$ the
standardized reward, $\mathbf{J}_t$ the Jacobian of the policy logits with respect to $\mathbf{H}_t$, and $\pi_{f}$ the
probability of the drawn fixation $f_{t+1}$, one cell of the $64\times64$ grid. Since $|\tilde r_t| \le \sqrt{n-1}$ for
the $n$ samples standardized together and $1 - \pi_f \le 1$, only $\|\mathbf{J}_t\|$ can grow without limit. We call it the
Jacobian norm of the policy head: it sets how strongly the head amplifies a change of the memory features into a change of
the logits, and equally how strongly it amplifies the policy gradient on its way back into the memory features. For a sharp
policy, the most likely cell has a probability close to $1$: if it is drawn, $1 - \pi_f \approx 0$ and the gradient is almost zero, and only the rare samples of another cell give $1 - \pi_f \approx 1$ and a gradient whose size is set by the
Jacobian norm. Once the policy has sharpened, the bound thus predicts that the gradient into the memory arrives in rare,
large bursts.

\begin{figure}[t]
  \centering

  \includegraphics[width=\textwidth]{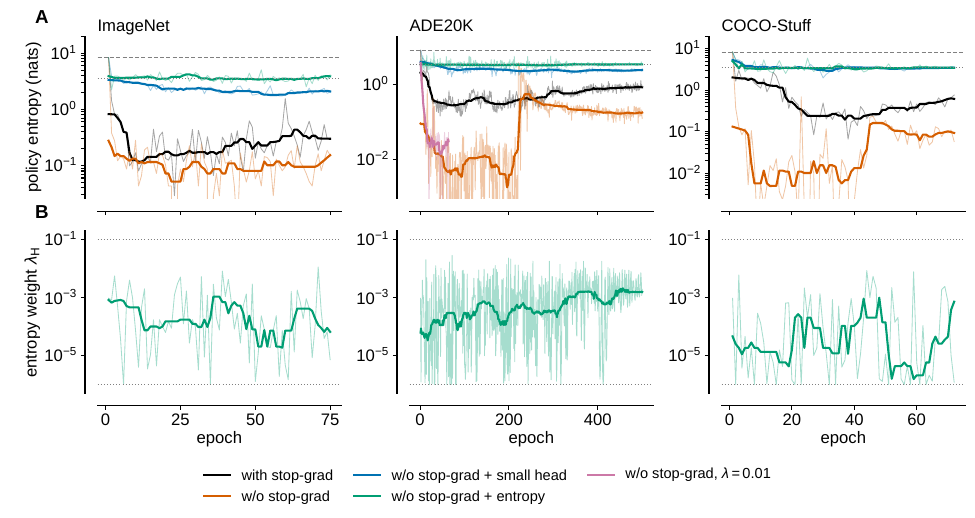}
  \caption{\textbf{Policy entropy and the entropy weight during training.} Models as in \Cref{tab:joint_ablation} on
  ImageNet, ADE20K and COCO-Stuff (one model each); on ADE20K also the run without stop-gradient with a $10\times$ smaller
  policy-loss weight ($\lambda = 0.01$) of \Cref{fig:ade_interventions}. \textbf{(A)}~policy entropy (dashed: uniform
  policy, $\log 4096 \approx 8.3$ nats; dotted: the entropy target of $3.5$ nats of the entropy-target variant).
  \textbf{(B)}~the entropy weight $\lambda_{\mathcal{H}}$ set by the update rule of the entropy-target variant
  (dotted: the bounds $10^{-6}$ and $10^{-1}$ within which it is kept); in the second half of training it is about $1.6\times10^{-4}$ on ImageNet, $1.0\times10^{-3}$ on
  ADE20K and $3.6\times10^{-5}$ on COCO-Stuff. One logged minibatch per epoch; bold lines are rolling medians (window a
  tenth of the run, at least nine epochs), faint lines the raw values.
  \textbf{Deviations (\Cref{tab:regimes}):} gradient on backbone and memory (D4; all models except the one with stop-gradient), reduced budget (D11; ImageNet and the run at $\lambda = 0.01$).}
  \label{fig:entropy_training}
\end{figure}

\paragraph{The policy sharpens and the Jacobian norm of its head grows.}
Without the stop-gradient, the policies sharpen within a few epochs: their entropy falls to $0.01$--$0.2$~nats, also in
the ADE20K run with a $10\times$ smaller policy-loss weight ($\lambda = 0.01$; \Cref{fig:entropy_training}A). The
range of their logits (largest minus smallest logit of a probability map) is much larger than with the stop-gradient and
by far the largest on ImageNet (\Cref{tab:logit_range}), whose runs, unlike those on ADE20K and COCO-Stuff, are trained
in bf16 (\Cref{app:bf16}). The Jacobian norm of the policy head also grows, both with and without the stop-gradient, but
far more without it: on ImageNet, it rises between the first and the last epoch from $137$ to
\num[group-digits=integer]{22982} without the stop-gradient (about $170$-fold) and from $50$ to \num[group-digits=integer]{1063} with it (about
$21$-fold; \Cref{tab:logit_range}).

\begin{figure}[t]
  \centering

  \includegraphics[width=\textwidth]{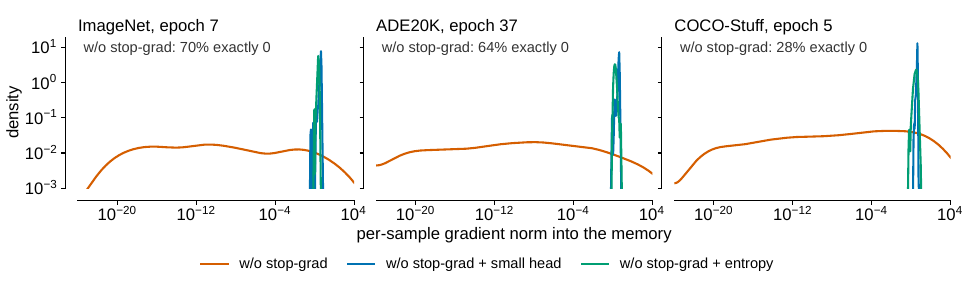}
  \caption{\textbf{Bursts of the policy gradient.} Distribution of the per-sample gradient norm
  $\|\partial \log\pi_\theta(f_{t+1} \mid \mathbf{H}_t)/\partial \mathbf{H}_t\|$ into the memory features, for the model
  without stop-gradient and the two variants of \Cref{app:entropy_new}, as a kernel density estimate over $\log_{10}$ of
  the norm. One sample is one fixation drawn from one map ($8$ per map; $64$ ImageNet or $32$ segmentation validation
  images), computed in fp32. Samples with a gradient of exactly zero cannot be placed on the log axis and are left out of
  the curves, so each curve integrates to the share of samples with a nonzero gradient. Each panel prints the share of
  zeros for the model without stop-gradient (e.g.\ $70\%$ on ImageNet). Epochs as in the panel
  titles; single runs. \textbf{Deviations (\Cref{tab:regimes}):} gradient on backbone and memory (D4), early epoch (D6; the epochs in the panel titles), reduced budget (D11; ImageNet).}
  \label{fig:policy_bursts}
\end{figure}

\paragraph{The gradient arrives in bursts.}
As the bound predicts, once the policies without stop-gradient have sharpened, the median per-sample gradient into the
memory is essentially zero: on most probability maps one cell holds almost all of the probability, so $1 - \pi_f \approx 0$
whichever cell is drawn. Almost all of the gradient comes from the few maps ($2$--$6\%$) on which the policy is still
undecided, mostly between two cells. There, the less likely cell is sometimes drawn despite its low probability, and
such a sample has a large $1 - \pi_f$ and therefore a large gradient. These samples are rare ($0.2$--$1.4\%$ of all
samples) but carry $34$--$55\%$ of the summed gradient (checkpoints of \Cref{fig:policy_bursts}).

Per minibatch, the policy gradient on backbone and memory of the models without stop-gradient is either negligible or
many times the task gradient there: the ratio of the two norms is below $0.1$ on $30$--$50\%$ of the minibatches and above $10$ on $2$--$18\%$. On the most extreme minibatch, the policy gradient is $140$ times the task gradient. The ratio is computed
as in \Cref{fig:joint_ratio_pdf}B but on $40$ training minibatches at the checkpoints of
\Cref{fig:policy_bursts}. Its direction is unrelated to that of
the task gradient: the cosine between the two has a median of $-0.064$ to $0.004$ over training minibatches on the final
checkpoints of \Cref{tab:joint_ablation}.

\begin{table}[t]
  \centering
  \caption{\textbf{Logit range and Jacobian norm of the policy head, without and with stop-gradient.} \emph{Logit range}:
  largest minus smallest policy logit of one probability map, median over $500$ validation images and the three decision
  steps, final checkpoints. \emph{Jacobian norm}: $\|\mathbf{J}_t\|$, the largest singular value of the Jacobian
  of the $64\times64$ policy logits with respect to the memory features ($128\times64\times64$), computed by power iteration
  without forming the matrix; median over $32$ validation images and the three decision steps, at the first and the last
  epoch; ImageNet only. Models as in
  \Cref{tab:joint_ablation}.
  \textbf{Deviations (\Cref{tab:regimes}):} gradient on backbone and memory (D4; models without stop-gradient), reduced budget (D11; ImageNet).}
  \label{tab:logit_range}
  \small
  \begin{tabular}{lrr}
    \toprule
    & w/o stop-grad & with stop-grad \\
    \midrule
    \multicolumn{3}{l}{\emph{Logit range}} \\
    ImageNet   & \num[group-digits=integer]{82946} & $360$ \\
    ADE20K     & \num[group-digits=integer]{1802}  & $240$ \\
    COCO-Stuff & \num[group-digits=integer]{3184}  & $616$ \\
    \midrule
    \multicolumn{3}{l}{\emph{Jacobian norm of the policy head, ImageNet}} \\
    first epoch & $137$ & $50$ \\
    last epoch  & \num[group-digits=integer]{22982} & \num[group-digits=integer]{1063} \\
    \bottomrule
  \end{tabular}
\end{table}

\subsection{A separate problem in bf16: rounding ties the largest logits}
\label{app:bf16}
The classification runs (ImageNet, C-MNIST) are trained in bf16. On ImageNet, where it was measured, this adds a second
problem to the bursts of \Cref{app:why_it_grows}, caused by rounding of the largest policy logits. bf16 stores a logit $z$ with $|z| \in [2^e, 2^{e+1})$,
for an integer exponent $e$, in steps of $2^{e-7}$ ($256$ at $|z| \approx 4\times10^4$). With logits as large as those of
the models without stop-gradient (\Cref{tab:logit_range}), the $k$ largest logits can round to the same value. Then
$1 - \pi_{f} \approx (k-1)/k$ on every sample from such a map, and the gradient of each of these samples, not only of the rare ones of
\Cref{app:why_it_grows}, grows with the Jacobian norm $\|\mathbf{J}_t\|$.

On the checkpoints of the ImageNet run without stop-gradient of \Cref{tab:joint_ablation}, we compute the gradients once
in fp32 and once in bf16. The ratio $q$ of the policy to the task gradient norm on backbone and memory agrees until
epoch~1 ($1.3$ vs.\ $1.2$) and then separates. In fp32 it falls below $0.01$ from epoch~19, as the policy sharpens. In
bf16 it rises to $61$, with about $10$--$25\%$ of the probability maps having a tied maximum. In training, which ran in bf16, gradient-norm clipping acted on every logged minibatch from epoch~5 on ($71$ of $75$; one minibatch is logged per epoch).

fp32 does not have this problem. On ADE20K and COCO-Stuff, which are trained in fp32, we compute the per-sample policy
gradient into the memory features once in fp32 and once in fp64, on $12$ and $11$ checkpoints spread over training of the
runs without stop-gradient of \Cref{tab:joint_ablation} ($16$ validation images, $8$ fixations drawn per probability map).
The two agree within $2\%$ on $16$ of the $23$ checkpoints and within about a factor of two on all, with no trend over
training. The runs without
stop-gradient on these datasets lose task performance all the same (\Cref{tab:joint_ablation}). bf16 thus enlarges the
policy gradient on ImageNet but does not alone cause the drop in task performance; computing the policy logits in fp32 avoids
the ties. 

\subsection{Potential remedies}
\label{app:entropy_new}
This subsection tests potential remedies for the drop in task performance without the stop-gradient. The first scales
down the policy gradient on backbone and memory with a $10\times$ smaller policy-loss weight $\lambda$. The other two
are variants that act on the two properties that shape this gradient (\Cref{app:why_it_grows}): the Jacobian norm
$\|\mathbf{J}_t\|$ of the policy head (the norm of the Jacobian of the policy logits with respect to the memory features),
which sets how large a burst can be, and the sharpness of the policy, which makes the gradient arrive in bursts. A small
policy head limits the Jacobian norm, and with fewer weights its policy may also sharpen less; an entropy target limits
the sharpening directly. Each of the three changes only this one thing in the runs
without stop-gradient of \Cref{tab:joint_ablation}. We compare their task performance, the value of their policies and how
their policy gradient reaches backbone and memory with those of the models with and without stop-gradient.

\paragraph{Smaller policy-loss weight.}
The run without stop-gradient is repeated on ADE20K with a $10\times$ smaller policy-loss weight, $\lambda = 0.01$
instead of $0.1$, and otherwise identically (single run). The policy head receives only the policy gradient, and AdamW
normalizes each parameter's step by its own gradient scale, so its steps stay about the same. Backbone and memory also
receive the task gradient, so for them the smaller $\lambda$ scales down the policy gradient against the task gradient:
it acts as a partial stop-gradient.

\paragraph{Small head.}
The policy head (\Cref{tab:architecture_heads}) is replaced by a single $1\times1$ convolution from the $128$-channel memory
to one logit per cell ($128$ weights $w$ and a bias, $129$ parameters). Its Jacobian norm $\|\mathbf{J}_t\|$ is then at most
the norm $\|w\|$ of these weights. It can then grow only with the norm of these $128$ weights, not with a product over
four layers as in the default head. The head was sized so that its
policy reaches about the entropy that the entropy target condition (below) enforces ($2.0$--$3.5$ nats, \Cref{fig:entropy_training}A), so the small head limits the sharpening as well.

\paragraph{Entropy target.}
The policy head is kept, and entropy regularization (\Cref{sec:gradient_conflicts}), $-\lambda_{\mathcal{H}} \bar{\mathcal{H}}$, is added to the loss, with $\bar{\mathcal{H}}$
the policy entropy averaged over the batch and the decision steps. To hold the entropy at a target
$\mathcal{H}^{\star} = 3.5$ nats, $\lambda_{\mathcal{H}}$ is multiplied after every training step by
$\exp\!\big(\eta\,(\mathcal{H}^{\star} - \bar{\mathcal{H}})\big)$, with step size $\eta = 10^{-2}$, and kept within
$[10^{-6}, 10^{-1}]$: it grows while the policy is sharper than the target and shrinks while it is flatter. We use a
target rather than a fixed weight because the weight that holds $3.5$ nats differs by about an order of magnitude between
datasets and drifts within a run (\Cref{fig:entropy_training}B). With the entropy held at the target, the
policy cannot become sharp: $1 - \pi_f$ is not, as in \Cref{app:why_it_grows}, almost always near $0$ and occasionally
near $1$.

\paragraph{Results.}
A smaller policy-loss weight only delays the drop in task performance. The policy sharpens at about the same speed as at
$\lambda = 0.1$ (\Cref{fig:entropy_training}A). On ADE20K, the run at $\lambda = 0.01$ stays within about
$1$~mIoU of the run with stop-gradient for eight epochs, while the run without stop-gradient at $\lambda = 0.1$
falls behind from the first epoch (\Cref{fig:ade_interventions}). After about eight epochs, the run at $\lambda = 0.01$ falls
behind the run with stop-gradient as well ($11.5$ against $17.4$~mIoU at epoch~32). Trivially, a much smaller
weight would eventually behave like the stop-gradient, but we did not investigate further configurations.

The small head and the entropy target keep the entropy at $2.0$--$3.5$ nats (\Cref{fig:entropy_training}A), and gradient-norm clipping
almost never acts on them. They recover the full task performance on ImageNet ($61.4\%$ and $61.9\%$ against $61.6\%$
with stop-gradient) and most of it on the segmentation datasets, $80$--$92\%$ of the drop (\Cref{tab:joint_ablation}). On C-MNIST, the entropy target matches the model
with stop-gradient ($99.3\%$), while the small head loses $4.2\%$ (down to $95.1\%$). Transferred to a frozen
random-location model, their policies improve on random locations about as much as the policy with stop-gradient or
slightly more (\Cref{tab:xpol_jointabl}).

Both variants also change how the policy gradient reaches backbone and memory: it is spread evenly over the samples
instead of arriving in bursts (\Cref{fig:policy_bursts}). Measured as in \Cref{app:why_it_grows}, on the
checkpoints of \Cref{fig:policy_bursts}, it stays within $0.2$--$6\times$ the task gradient on every minibatch. The variants thus receive a steady policy gradient on backbone and memory. In many minibatches it is larger than in
the runs without stop-gradient, where it is often negligible and large only in rare bursts
(\Cref{app:why_it_grows}). Yet the variants lose much less task performance. Policy gradient on
backbone and memory may therefore not necessarily be harmful in itself. The severe drop goes together with its arrival in rare, large
bursts.

\subsection{Summary and the stop-gradient in our default models}
\label{app:alternatives}
\paragraph{What goes wrong.}
Without the stop-gradient, task performance drops on all four datasets (\Cref{app:joint_cost}). Worse fixations,
clipping and the optimizer state are unlikely to explain this drop (\Cref{app:pg_cause}), which suggests that the
policy gradient on backbone and memory itself causes it. During training, the policy
sharpens and the Jacobian norm of its head grows, so backbone and memory receive rare, large bursts of policy gradient in
directions unrelated to the task gradient (\Cref{app:why_it_grows}). On ImageNet, which is trained in bf16, rounding ties the largest policy logits, so that many samples send a large gradient (\Cref{app:bf16}).

\paragraph{What prevents it.}
The stop-gradient prevents the drop in task performance. Without it, a small policy head or an entropy target recovers
most of the task performance (\Cref{app:entropy_new}). Both keep the policy from sharpening.

\paragraph{What does not.}
A $10\times$ smaller policy-loss weight $\lambda$ only delays the drop in task performance (\Cref{app:entropy_new}), and
gradient-norm clipping does not prevent it (\Cref{app:pg_cause}). The runs trained in fp32 (ADE20K,
COCO-Stuff), in which the largest policy logits do not round to the same value as they do in bf16, lose task
performance as well (\Cref{app:bf16}).

\paragraph{What remains open.}
Which property of the bursts (their size, their direction, their frequency, effect on optimizer) harms backbone and memory is not identified,
and neither is the precise reason why the policy becomes so sharp. The run with a smaller policy-loss weight
(\Cref{app:entropy_new}) is a single run on ADE20K. The other factors tested (worse fixations, clipping, optimizer
state) are unlikely to explain the drop in task performance, but further contributing factors are not excluded.

\paragraph{The choice of the stop-gradient for our default models.}
The stop-gradient is one option among several and may not necessarily be the best one in general. Arguably, it is a sound baseline and needs no per-dataset tuning. Also, we found that it performs at least as well as the small head and the entropy target (they match
it on ImageNet and lose $0.5$--$1.4$~mIoU on segmentation; single runs). The other variants warrant further attention in future research.

\subsection{Derivation of the bound}
\label{app:pg_derivation}
For completeness, this subsection derives the bound used in \Cref{app:why_it_grows}.

\paragraph{Setting.}
$\operatorname{sg}[\cdot]$ is the stop-gradient: it passes its argument unchanged in the forward pass and lets no
gradient through in the backward pass. The model without stop-gradient drops the $\operatorname{sg}[\cdot]$ around
$\mathbf{H}_t$ in \Cref{eq:policy_loss}, weights its policy loss by $\lambda = 0.1$ and averages it over the batch of
size $B$:
\begin{equation*}
  \lambda \mathcal{L}_\mathrm{policy} = -\frac{\lambda}{B} \sum_{b=1}^{B} \sum_{t=1}^{T-1}
    \operatorname{sg}[\tilde r_t]\, \log \pi_\theta(f_{t+1} \mid \mathbf{H}_t),
\end{equation*}
where $\tilde r_t$, $f_{t+1}$ and $\mathbf{H}_t$ belong to sample $b$ (index dropped); $f_{t+1}$ is the drawn fixation,
one cell of the $64\times64$ grid. The policy is the softmax over the cells of the policy head's output
(\Cref{eq:policy}); $\psi_\theta$ is the convolutional projection of the policy head (\Cref{sec:sampling_mechanism}),
which maps the memory features to one value per cell $k$. Written out,
\begin{equation*}
  \log \pi_\theta(f_{t+1} \mid \mathbf{H}_t)
  = \big[\psi_\theta(\mathbf{H}_t)\big]_{f_{t+1}} - \log \sum_{j} \exp\big(\big[\psi_\theta(\mathbf{H}_t)\big]_j\big).
\end{equation*}
$\mathbf{H}_t$ thus enters only through $\psi_\theta(\mathbf{H}_t)$. These values are the policy logits,
$\mathbf{z}_t = \psi_\theta(\mathbf{H}_t)$ with entries $z_k$, so that $\boldsymbol{\pi}_t =
\operatorname{softmax}(\mathbf{z}_t)$ is the policy written as a vector, with entries $\pi_k$, and
$\mathbf{e}_{f_{t+1}}$ is the one-hot vector of the drawn cell. To shorten notation, we write $\mathbf{z}$,
$\boldsymbol{\pi}$ and $f$ for $\mathbf{z}_t$, $\boldsymbol{\pi}_t$ and $f_{t+1}$.

\paragraph{The chain rule.}
Since the term of one sample and step depends on $\mathbf{H}_t$ only through the logits $\mathbf{z}_t$, its gradient is a
product of three factors: the derivative of the loss term with respect to $\log \pi_f$, of $\log \pi_f$ with respect to
the logits, and of the logits with respect to the memory features,
\begin{equation*}
  \frac{\partial \lambda \mathcal{L}_\mathrm{policy}}{\partial \mathbf{H}_t}
  = \underbrace{\frac{\partial \lambda \mathcal{L}_\mathrm{policy}}{\partial \log \pi_f}}_{\text{step 1}}\,
    \Big(\underbrace{\frac{\partial \log \pi_f}{\partial \mathbf{z}_t}}_{\text{step 2}}\,
    \underbrace{\frac{\partial \mathbf{z}_t}{\partial \mathbf{H}_t}}_{\text{step 3}}\Big)^{\!\top}.
\end{equation*}

\paragraph{Step 1: loss term.}
$\tilde r_t$ is held constant by $\operatorname{sg}[\cdot]$, so $\partial \lambda \mathcal{L}_\mathrm{policy} /
\partial \log \pi_f = -\frac{\lambda}{B}\,\tilde r_t$.

\paragraph{Step 2: log-softmax.}
Writing out the log-softmax and differentiating with respect to one logit $z_k$,
\begin{equation}
  \begin{aligned}
    \log \pi_{f} &= z_{f} - \log \sum_j e^{z_j},\\[4pt]
    \frac{\partial \log \pi_{f}}{\partial z_k} &= \mathbf{1}[k = f] - \frac{e^{z_k}}{\sum_j e^{z_j}}
      = \mathbf{1}[k = f] - \pi_k,\\[4pt]
    \frac{\partial \log \pi_{f}}{\partial \mathbf{z}} &= \mathbf{e}_{f} - \boldsymbol{\pi}.
  \end{aligned}
  \label{eq:pg_logits}
\end{equation}
Raising the drawn cell's logit raises $\log \pi_{f}$ by $1 - \pi_{f}$; raising any other logit lowers it by that cell's
probability, through the softmax denominator.

\paragraph{Step 3: policy head.}
The last factor is the Jacobian $\mathbf{J}_t = \partial \mathbf{z}_t / \partial \mathbf{H}_t$ of the policy head. Together,
\begin{equation*}
  \frac{\partial \lambda \mathcal{L}_\mathrm{policy}}{\partial \mathbf{H}_t}
    = -\frac{\lambda}{B}\, \tilde r_t\, \mathbf{J}_t^\top (\mathbf{e}_{f_{t+1}} - \boldsymbol{\pi}_t).
\end{equation*}
This gradient is added to that of the task loss at $\mathbf{H}_t$ and backpropagated through memory and backbone; with the
stop-gradient, $\mathbf{H}_t$ enters the policy loss as a constant and the term is zero. Through the memory recurrence,
the policy terms of later steps also reach $\mathbf{H}_t$; the bound below applies to each term.

\paragraph{The bound.}
The operator norm $\|\mathbf{J}_t\|$ (the Jacobian norm of the policy head) bounds how much $\mathbf{J}_t^\top$ can
stretch a vector: $\|\mathbf{J}_t^\top \mathbf{v}\| \le \|\mathbf{J}_t\|\,\|\mathbf{v}\|$ for every
vector $\mathbf{v}$. For $\mathbf{v} = \mathbf{e}_{f} - \boldsymbol{\pi}$,
\begin{equation*}
  \|\mathbf{e}_{f} - \boldsymbol{\pi}\|^2 = (1 - \pi_{f})^2 + \sum_{k \neq f} \pi_k^2 \le 2(1 - \pi_{f})^2,
\end{equation*}
because the probabilities of the other cells sum to $1 - \pi_f$, and the sum of their squares is at most the square
of that sum, $(1 - \pi_f)^2$. Together,
\begin{equation}
  \Big\|\frac{\partial \lambda \mathcal{L}_\mathrm{policy}}{\partial \mathbf{H}_t}\Big\|
    \le \frac{\lambda}{B}\, |\tilde r_t|\; \|\mathbf{J}_t\|\; \sqrt{2}\,(1 - \pi_{f_{t+1}}).
  \label{eq:pg_bound_features}
\end{equation}
Of the three factors, $|\tilde r_t| \le \sqrt{n-1}$ for the $n$ samples standardized together (\Cref{eq:reward_norm}),
and $1 - \pi_{f} \le 1$; only the Jacobian norm $\|\mathbf{J}_t\|$ of the policy head is unbounded.

\clearpage
\section{Per-step and end-of-episode signals in the loss}
\label{app:loss_choices}

Two choices of the default training recipe give a learning signal at every step rather than only at the end of an episode: the
task loss supervises the prediction of each step (\Cref{app:task_loss_choices}), and the reward credits each fixation
with its own improvement (\Cref{app:policy_loss_choices}). This appendix compares each choice with its end-of-episode
counterpart, at the trained budget and with fewer or more glimpses at inference. The runs come from earlier stages of
model development, and the captions in this appendix list how they differ from our default models (\Cref{tab:regimes}).

\subsection{Per-step supervision in the task loss}
\label{app:task_loss_choices}
The task loss supervises the prediction of every step (\Cref{eq:task_loss}). The alternative examined here supervises
only the last step: it replaces \Cref{eq:task_loss} by the cross-entropy of the final prediction,
$\mathcal{L}_{\text{CE}}(\hat{\mathbf{y}}_T, \mathbf{y})$, as in RAM~\citep{mnihRecurrentModelsVisual2014}. Other
weightings of the steps are possible but are not examined here. Both pairs of models in \Cref{tab:supervision} were trained before we introduced
the stop-gradient (deviation~D4, \Cref{tab:regimes}). Since the choice acts on the task loss, which trains backbone and
memory either way, we expect the comparison to carry over to our default setting.

\begin{table}[t]
  \centering
  \caption{\textbf{Supervision at every step against supervision at the last step only.} ImageNet accuracy (\%) of models trained with
  $T = 4$ glimpses and unrolled to $T_\text{eval}$ glimpses, each evaluated with the fixations it was trained with (random
  locations or its native policy). $\Delta$ is the accuracy in column \emph{Per-step} minus that in column \emph{Last only}. The models use an earlier version of the backbone, so their
  accuracies are not comparable with those of the other tables. \textbf{Deviations (\Cref{tab:regimes}):} terminal reward (D1), terminal supervision (D2; the last-only
  variants), level normalizer (D3), gradient on backbone and memory (D4), best checkpoint (D7; random-location pair),
  baseline present (D8), different pretrained models (D9), different learning-rate schedule (D13), entropy
  regularization (D14). In addition, the learned last-only model is evaluated with fixations drawn from its policy instead of argmax fixations.}
  \label{tab:supervision}
  \small
  \begin{tabular}{crrrrrr}
    \toprule
    & \multicolumn{3}{c}{Random locations} & \multicolumn{3}{c}{Learned policy} \\
    \cmidrule(lr){2-4} \cmidrule(lr){5-7}
    $T_\text{eval}$ & Per-step & Last only & $\Delta$ & Per-step & Last only & $\Delta$ \\
    \midrule
    1  & $58.71$ & $52.77$ & $+5.94$ & $55.25$ & $46.23$ & $+9.02$ \\
    \rowcolor{gray!15} 4 (trained) & $62.92$ & $63.81$ & $-0.89$ & $64.66$ & $64.34$ & $+0.32$ \\
    10 & $65.49$ & $64.64$ & $+0.85$ & $65.21$ & $64.14$ & $+1.07$ \\
    20 & $65.73$ & $63.18$ & $+2.55$ & $64.13$ & $62.19$ & $+1.94$ \\
    \bottomrule
  \end{tabular}
\end{table}

At the trained budget, the choice makes no consistent difference. With fewer or more glimpses, per-step supervision is
clearly ahead. With a single glimpse, it leads by $5.94\%$ with random locations and by $9.02\%$ with a learned
policy, since only the per-step models had their early predictions supervised. With $20$ glimpses, it leads by $2.55\%$
with random locations and by $1.94\%$ with a learned policy, and both models supervised at the last step only have
fallen below their accuracy at the trained budget. Per-step task supervision thus appears to promote the model's ability to perform well
when unrolled for fewer or more glimpses than it was trained with.

\subsection{Per-step reward in the policy loss}
\label{app:policy_loss_choices}
The policy is trained by policy gradient (\Cref{eq:policy_loss}). By default, the reward of the fixation selected at step
$t$ is the improvement it produced (\Cref{eq:reward}),
\[
    r_t \;=\; \mathcal{L}_{\text{CE}}(\hat{\mathbf{y}}_t,\, \mathbf{y})
             - \mathcal{L}_{\text{CE}}(\hat{\mathbf{y}}_{t+1},\, \mathbf{y}),
    \qquad t = 1, \dots, T-1 .
\]
The alternative is the terminal reward, which gives every fixation the negative task loss at the last step as reward,
\[
    r_t^{\text{term}} \;=\; -\,\mathcal{L}_{\text{CE}}(\hat{\mathbf{y}}_T,\, \mathbf{y}),
    \qquad t = 1, \dots, T-1 .
\]

\Cref{tab:reward_horizon_new} compares the two rules on ImageNet below, at and past the trained budget. We train only the policy head, on a frozen ImageNet model that was trained with random locations, and evaluate both
policies with this same model. Differences in performance between the two rules therefore come purely from the
fixations they produce. For each rule, two policy heads with one instead of three
$3\times3$ convolutions are trained with different random seeds for $24$ epochs of $2000$ minibatches (learning rate $10^{-4}$, constant for eight
epochs, then halved every four).

\begin{table}[t]
  \centering
  \caption{\textbf{Reward rule and horizon on ImageNet.} Accuracy (\%) with policy heads trained on a \emph{frozen} random-location
  model and evaluated with it, unrolled to $T_\text{eval}$ glimpses. The column \emph{Random locations} shows that model
  with its own, randomly drawn locations. \emph{Terminal} and \emph{Per-step} are means over two policy heads trained
  with different random seeds; $\pm$ is half the difference between the two. $\Delta$ is the accuracy in column
  \emph{Per-step} minus that in column \emph{Terminal}, computed per seed and then averaged. \textbf{Deviations (\Cref{tab:regimes}):} terminal reward (D1; the terminal column), level normalizer (D3), frozen representation (D5), baseline present (D8),
  policy-transfer score (D10), reduced budget (D11; one GPU, $24$ epochs of $2000$ steps). In addition, the policy head has one instead of three $3\times3$ convolutions.}
  \label{tab:reward_horizon_new}
  \small
  \begin{tabular}{crrrr}
    \toprule
    $T_\text{eval}$ & Random locations & Terminal & Per-step & $\Delta$ \\
    \midrule
    1  & $58.89$ & $58.89$ & $58.89$ & $0.00$ \\
    2  & $61.17$ & $63.12 \pm 0.06$ & $63.47 \pm 0.02$ & $+0.35 \pm 0.08$ \\
    \rowcolor{gray!15} 4 (trained) & $63.54$ & $65.27 \pm 0.47$ & $\mathbf{65.89 \pm 0.07}$ & $\mathbf{+0.62 \pm 0.40}$ \\
    6  & $64.65$ & $64.97 \pm 0.17$ & $66.50 \pm 0.06$ & $+1.53 \pm 0.23$ \\
    8  & $65.26$ & $64.56 \pm 0.16$ & $66.57 \pm 0.01$ & $+2.01 \pm 0.15$ \\
    10 & $65.80$ & $64.21 \pm 0.20$ & $\mathbf{66.58 \pm 0.04}$ & $\mathbf{+2.36 \pm 0.15}$ \\
    \bottomrule
  \end{tabular}
\end{table}

At the trained budget, the reward rule makes a small difference: the policy trained with the per-step reward is ahead by
$0.62\%$. For longer rollouts, the difference gets bigger. Up to $T_\text{eval} = 10$, the policy trained with the terminal reward falls from
$65.27$ to $64.21\%$ while random locations climb from $63.54$ to $65.80\%$, so its margin over them turns from $+1.73$
to $-1.59\%$. The policy trained with the per-step reward keeps rising, to $66.58\%$, and stays above random locations
($+0.78\%$). Both seeds show this pattern: with the terminal reward, each falls below random locations from
$T_\text{eval} = 8$ on, and with the per-step reward, each stays above them. Like per-step task supervision (\Cref{app:task_loss_choices}), the per-step reward thus appears to promote
the model's ability to perform well when unrolled for more glimpses than it was trained with. The reason for the advantage of the per-step forms, and how per-step supervision compares to terminal supervision when
using other training algorithms and optimization settings, remain to be investigated.

\end{document}